\documentclass[journal]{IEEEtran}
\ifCLASSINFOpdf
\else
\fi
\usepackage{multirow}
\usepackage{graphicx}
\usepackage{amsmath}
\usepackage{amssymb}
\usepackage{booktabs}
\usepackage{cite}
\usepackage{subfigure}
\usepackage{colortbl}
\usepackage{xcolor}
\usepackage{tikz}
\usepackage{array}
\usepackage{subcaption}
\usepackage{transparent}
\definecolor{tabblue}{HTML}{1f77b4}
\definecolor{taborange}{HTML}{ff7f0e}
\definecolor{tabgreen}{HTML}{2ca02c}
\definecolor{tabred}{HTML}{d62728}
\definecolor{tabpurple}{HTML}{9467bd}
\definecolor{tabbrown}{HTML}{8c564b}
\definecolor{tabpink}{HTML}{e377c2}
\definecolor{tabgray}{HTML}{7f7f7f}
\definecolor{tabyellowgreen}{HTML}{bcbd22}
\definecolor{tabcyan}{HTML}{17becf}
\begin{document}
%
\title{Classifier Enhancement Using Extended Context and Domain Experts for Semantic Segmentation}
%
%
%

\author{Huadong Tang,
        Youpeng Zhao, Min Xu, \textit{Senior Member, IEEE}, Jun Wang, \textit{Senior Member, IEEE}, Qiang Wu$^*$, \textit{Senior Member, IEEE}
\thanks{Huadong Tang, Min Xu and Qiang Wu are with the School of Electrical and Data Engineering, University of Technology Sydney, Sydney, Australia. (e-mail: huadong.tang@student.uts.edu.au; min.xu@uts.edu.au, qiang.wu@uts.edu.au.)}
\thanks{Youpeng Zhao and Jun Wang is with the Department of Computer Science, University of Central Florida, Florida, America. (e-mail: youpeng.zhao@ucf.edu; jun.wang@ucf.edu.) }
\thanks{*Co-corresponding author.}}

%
%

\markboth{IEEE TRANSACTIONS ON Multimedia}%
{Shell \MakeLowercase{\textit{et al.}}: Classifier Enhancement Using Extended Context and Domain Experts for Semantic Segmentation}
%



\maketitle

\begin{abstract}
Prevalent semantic segmentation methods generally adopt a vanilla classifier to categorize each pixel into specific classes. 
  Although such a classifier learns global information from the training data, this information is represented by a set of fixed parameters (weights and biases).
  However, each image has a different class distribution, which prevents the classifier from addressing the unique characteristics of individual images.
  At the dataset level, class imbalance leads to segmentation results being biased towards majority classes, limiting the model's effectiveness in identifying and segmenting minority class regions.
  In this paper, we propose an Extended Context-Aware Classifier (ECAC) that dynamically adjusts the classifier using global (dataset-level) and local (image-level) contextual information.
  Specifically, we leverage a memory bank to learn dataset-level contextual information of each class, incorporating the class-specific contextual information from the current image to improve the classifier for precise pixel labeling.
  Additionally, a teacher-student network paradigm is adopted, where the domain expert (teacher network) dynamically adjusts contextual information with ground truth and transfers knowledge to the student network.  
  Comprehensive experiments illustrate that the proposed ECAC can achieve state-of-the-art performance across several datasets, including ADE20K, COCO-Stuff10K, and Pascal-Context. 
\end{abstract}

\begin{IEEEkeywords}
 Semantic segmentation, Teacher-student network, Class-specific contextual information
\end{IEEEkeywords}

%
\IEEEpeerreviewmaketitle

\section{Introduction}
\label{sec:intro}

Semantic segmentation plays a vital role in modern computer vision applications. 
It aims to classify each pixel into a specific category or class. 
In recent years, most of the state-of-the-art segmentation methods~\cite{tang2023class,exploring,rethinking,10474206,SHANG2022124,tang2024caa,10462518,10109193,shen2024learning,ma2024active} have been driven by the cornerstone fully convolutional networks (FCNs)~\cite{fcn}. 
Its encoder-decoder structure extracts features and then uses up-sampling to restore the spatial details, resulting in significant enhancements in semantic segmentation. 

In the decoding stage, the model restores the spatial information and applies the class label on each pixel to get the segmentation results.
An essential component in the decoder is the classifier that ultimately assigns a label to each pixel. 
Existing methods generally utilize the classic vanilla classifier to learn a set of fixed parameters from the training data. 
This results in an inherent challenge when facing highly diverse data (\emph{e.g.} different objects, scenes, or conditions) during training. 
Additionally, the vanilla classifier is particularly sensitive to class imbalance, where the model tends to prioritize majority classes while neglecting minority classes.
To solve this problem, several methods~\cite{yu2022distribution,blv,aucseg} propose to address class imbalance by modifying the loss function, such as using weighted cross-entropy to increase the importance of minority classes or exploring the potential of AUC optimization in pixel-level long-tail problems.
However, these approaches often rely on indirect adjustments, requiring careful hyperparameter tuning and potentially failing to address underlying feature representation or data distribution issues.
Other methods~\cite{gmmseg,protoseg,mcibi,shi2024cognition} leverage prototypes or memory banks to mitigate the imbalance issue by maintaining robust feature representations for all classes. 
Though effective, these methods still suffer from feature inconsistency when test images exhibit varying data distributions, a challenge that can be exacerbated by class imbalance during training, where minority class features may be underrepresented.
To this end, the open question is: \emph{Can we leverage information from the memory bank to enhance the classifier, thereby addressing class imbalance issues and improving prediction accuracy?}

\begin{figure}[]
    \centering
    \subfigure[Parameters]{
        \begin{minipage}{0.5\linewidth} 
            \centering
            \includegraphics[width=1\linewidth, keepaspectratio]{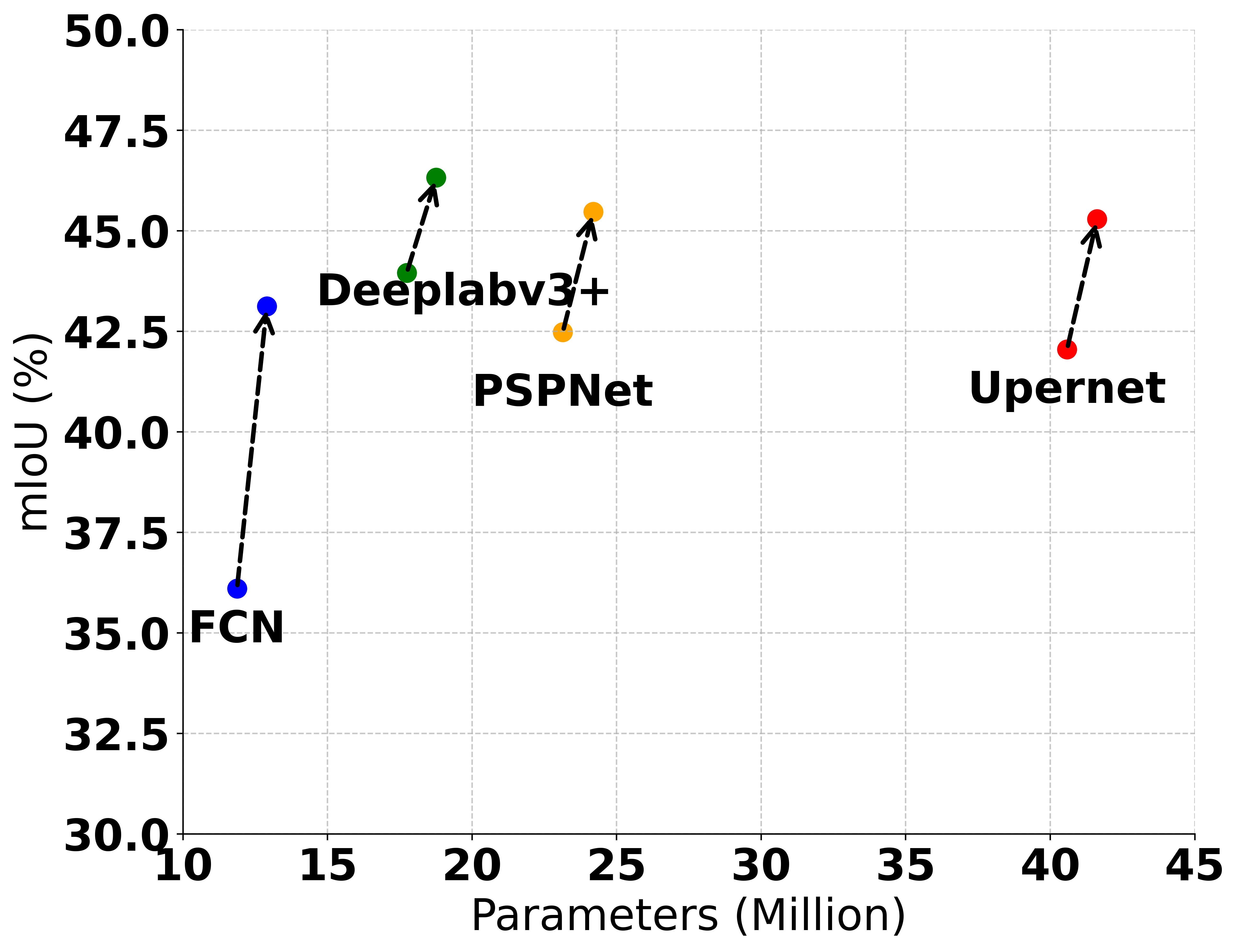}
            \label{fig:sub1}
        \end{minipage}
    }
    \hfill
    
    \subfigure[FLOPs]{
        \begin{minipage}{0.5\linewidth} 
            \centering
            \includegraphics[width=1\linewidth, keepaspectratio]{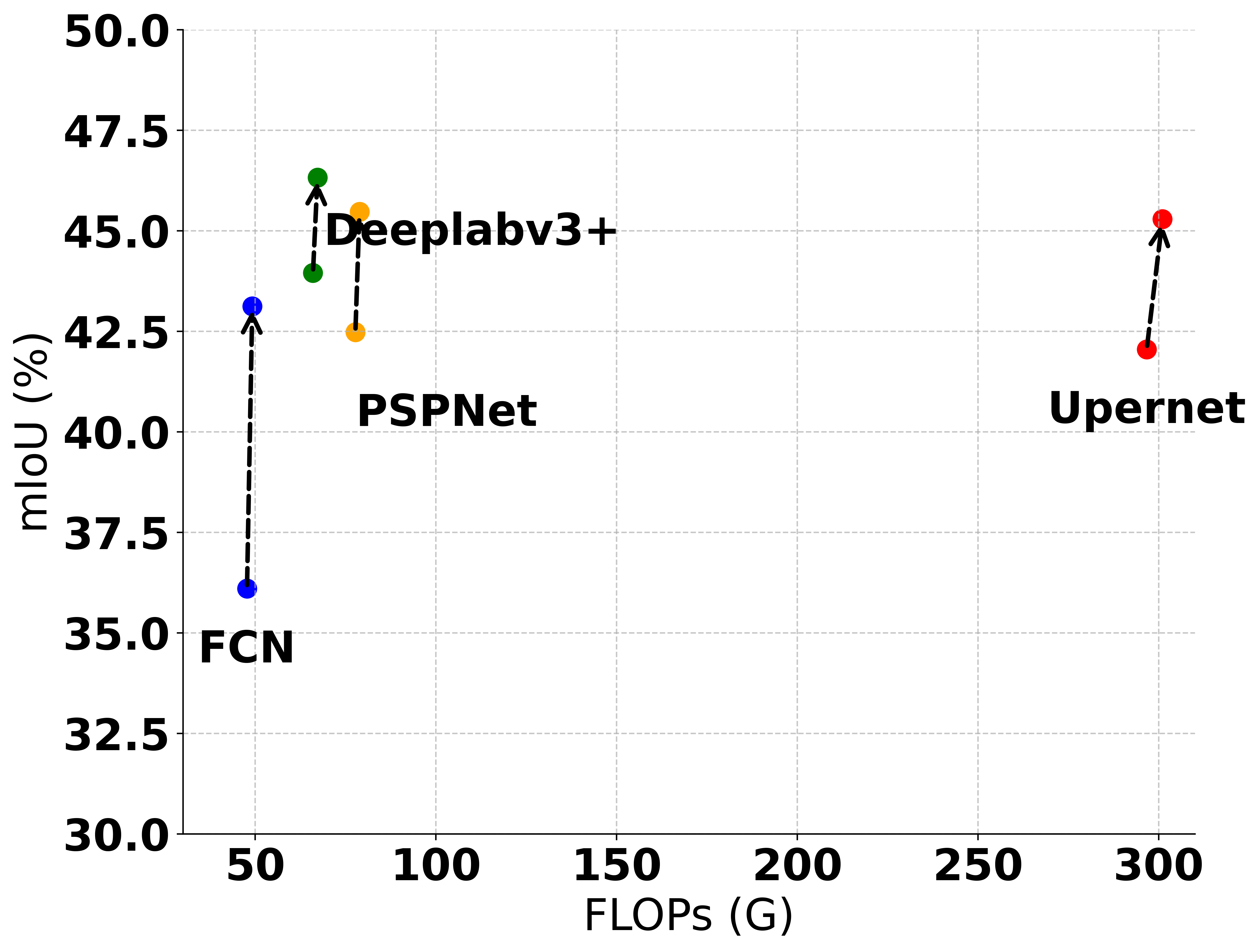}
            \label{fig:sub2}
        \end{minipage}
    }
    \vspace{-2mm}
    \caption{Analysis of inference complexity and accuracy for ADE20K. The arrows in the figure represent the improvement achieved by our ECAC method. The lower ends of the arrows represent the performance of the original methods, while the upper ends show the improved performance after incorporating ECAC. Our ECAC significantly improves the segmentation performance while bringing a little computational complexity.}
    \label{complex}
\end{figure}

To tackle the aforementioned problems, this paper proposes an Extended Context-Aware classifier (ECAC) that embeds global (dataset-level) and local (image-level) contextual information and a calibration stage to mitigate the class imbalance issue.
\underline{First}, we establish a dynamically updated memory bank to explicitly store dataset-level category information, serving as a dataset-level class center for each category. 
The memory bank independently updates each class, ensuring that even minority classes with limited samples retain sufficient category information. However, despite this design, the memory bank alone cannot fully overcome the biases induced by class imbalance.
To address this, we adopt a teacher-student network framework inspired by~\cite{cac}, implementing knowledge distillation where the teacher network, informed by GT labels, generates a GT-based class center, and the student network leverages the image-level class center derived from individual images to mimic it. 
These representations are then combined with the memory bank via a concatenation operation to form an enhanced classifier. 
Since GT labels are unavailable during inference, this approach enables the classifier to better approximate GT-level performance, improving classification accuracy under imbalanced conditions. The learning process is supervised by the Kullback-Leibler (KL) divergence, ensuring effective knowledge transfer from the teacher to the student.
Additionally, we propose an intra-variance loss to transfer intra-class feature variance from the teacher output to the student output, further enhancing the student model’s segmentation performance.
\underline{Second}, to further improve robustness, we additionally introduce a calibration stage to adjust classification scores by applying a learnable linear transformation, aligning the classifier output with the true class distribution. 
This ensures a more balanced and robust prediction by dynamically fine-tuning the logits through learnable parameters, ultimately improving the fairness and accuracy of the model.
Moreover, our approach improves the performance while only slightly increasing the model complexity, as demonstrated in Figure \ref{complex}.

In a nutshell, our main contributions to this work are:
\begin{itemize}
    \item 
    We propose an Extended Context-Aware Classifier (ECAC) that effectively tackles the issues of class imbalance and varying class distributions in semantic segmentation. By embedding both global (dataset-level) and local (image-level) contextual information, ECAC enhances pixel-wise classification precision across diverse datasets. 
    \item 
    We develop a novel framework integrating a dynamically updated memory bank with a teacher-student network paradigm. The memory bank preserves dataset-level class representations, while the teacher-student mechanism, augmented by a calibration stage, refines contextual understanding and mitigates biases, achieving robust performance even for underrepresented classes.
    \item We establish ECAC as a lightweight, plug-and-play module compatible with existing segmentation architectures. Comprehensive experiments on benchmarks such as ADE20K, COCO-Stuff10K, and Pascal-Context demonstrate its superior performance and adaptability, significantly advancing state-of-the-art segmentation capabilities with minimal computational overhead.
\end{itemize}

\section{Related Work}
\subsection{Semantic Segmentation based on Contextual Information}
In recent years, the field of semantic segmentation has witnessed remarkable advancements. 
Such progress is largely attributed to the rapid development of deep learning-based backbone architectures, which have evolved from CNN-based structures like ResNet\cite{resnet} to Transformer-like models such as ViT\cite{vit} and Swin\cite{swin}. 
Contextual information plays a key role in achieving accurate semantic segmentation, and aggregating this information is a widely used technique to improve feature representation in models.
 Methods for capturing contextual information generally encompass individual-image context aggregation (\emph{i.e}, multi-scale, attention mechanisms and relational context) and cross-image context aggregation (\emph{i.e}, contrastive learning and memory-based). 
 For instance, CPNet~\cite{cpnet} aims to capture long-range spatial (pixel-to-pixel) information for enhanced feature representation. 
 In contrast, approaches~\cite{ocrnet, isnet} based on pixel-to-class relations introduce the idea of a class center to capture the holistic context of each class.

 Recent works propose to utilize the concept of memory banks to study cross-image contextual information for image semantic segmentation, such as weakly supervised learning~\cite{sun2020mining}, semi-supervised learning~\cite{alonso2021semi,9861688} and fully supervised learning~\cite{exploring, rethinking, mcibi, mcibi++}.
CIPC~\cite{exploring} proposes a supervised, pixel-wise contrastive learning approach for semantic segmentation, lifting the traditional image-level training strategy to an inter-image, pixel-to-pixel paradigm. 
Protoseg~\cite{rethinking} introduces a non-parametric framework to learn dataset-level prototypes. 
MCIBI and MCIBI++~\cite{mcibi, mcibi++} mine the homogeneous contextual dependencies between images. 
However, despite this design, the memory bank alone cannot fully overcome the biases induced by class imbalance.

\subsection{Class Imbalance in Semantic Segmentation.}
Class imbalance is a well-recognized challenge in semantic segmentation, where certain classes (~\emph{e.g.}, background or common objects) dominate the dataset, while others (~\emph{e.g.}, rare objects or fine details) are underrepresented. 
Several approaches have been proposed to address this issue. 
For instance, DAMC~\cite{yu2022distribution} tackles class imbalance by introducing a margin calibration strategy that adjusts decision boundaries based on the label distribution.
BLV~\cite{blv} approaches the problem by injecting category-wise variation into logits during training.
Besides, AUCSeg~\cite{aucseg} explores AUC optimization methods in the context of pixel-level long-tail semantic segmentation.
However, these methods address class imbalance implicitly by adjusting the loss function or learning dynamics, without explicitly modeling class distributions. In contrast, prototype-based methods~\cite{gmmseg,protoseg,cac,ssa} learn representative embeddings for each class, directly enhancing feature discrimination, especially for tail classes. For example, SSA~\cite{ssa} proposes a novel semantic and spatial adaptive classifier to mitigate the class imbalance issue. 
Nevertheless, these methods still rely on the fixed parameters of the vanilla classifier, which often exacerbate class imbalance and degrade performance on minor categories.

To address the existing challenge, we propose a simple yet effective approach. It leverages a memory bank to learn dataset-level category features, capturing global distributions to alleviate class imbalance, and uses a teacher-student network to correct biases. A calibration stage further mitigates class imbalance, enhancing segmentation accuracy across all categories.

\begin{figure*}[t]
\begin{center}
\includegraphics[width=1\textwidth, keepaspectratio]{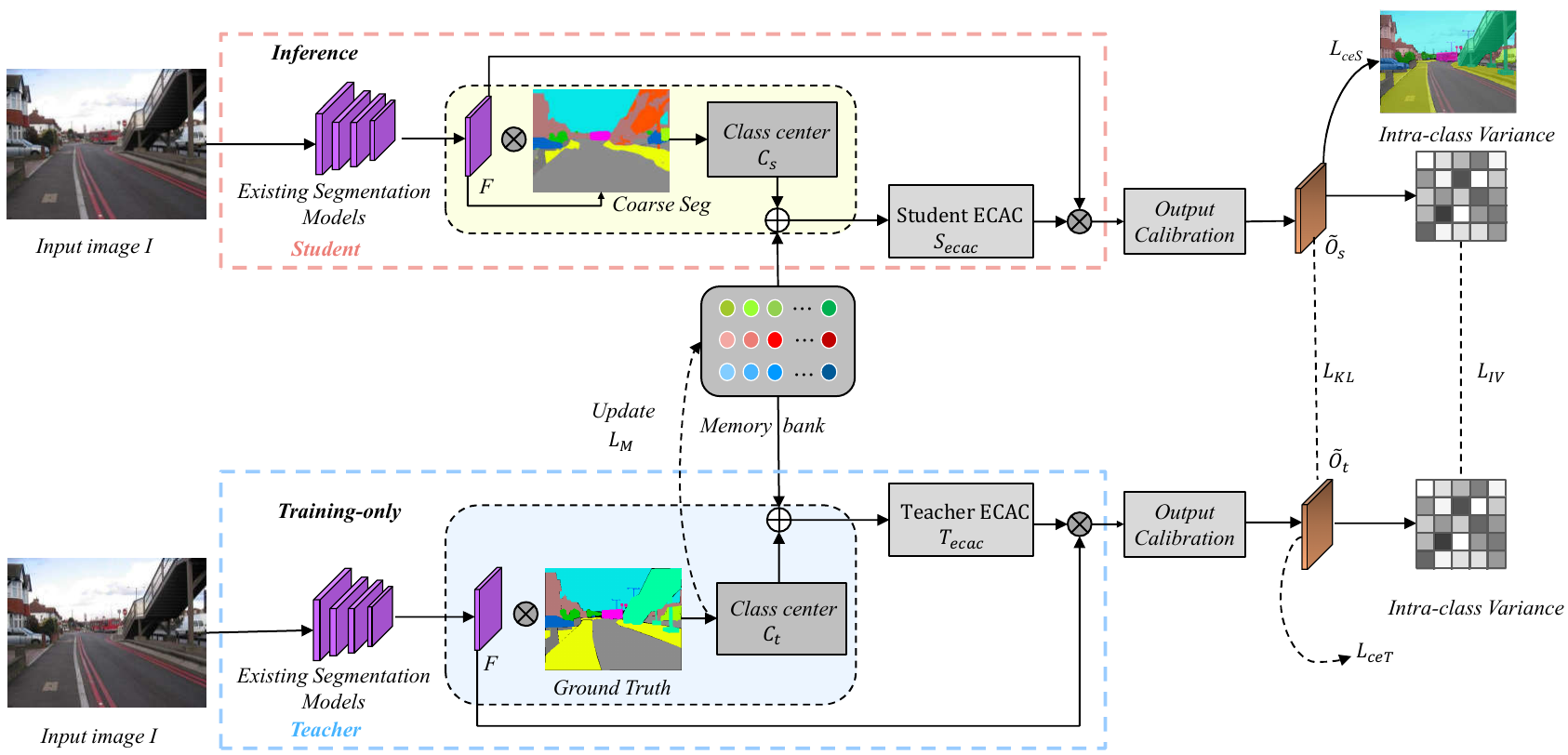}
\end{center}
\vspace{-2mm}
\caption{\textbf{The overview of the proposed ECAC.} The memory bank $\mathcal{M}$ stores the dataset-level category information. Combined with the within-image class center, we obtain an extended context-aware classifier. A calibration is adopted to mitigate the imbalanced issue. A teacher-student network is adopted to transfer comprehensive contextual information extracted by the ground-truth label to further enhance the classifier. ({\color{green}$\bullet$} {\color{red}$\bullet$} {\color{blue}$\bullet$}) represents the mean features of different classes.}
\label{fig:arc}
\end{figure*}

\section{Methodology}
\subsection{Overview}
Our proposed ECAC framework is illustrated in Figure \ref{fig:arc}.
We first set up a dynamically updated memory bank to store global (dataset-level) category information. 
This ensures that even minority classes with limited samples maintain sufficient class-specific representations, unlike a vanilla classifier that might struggle to adequately represent minority classes due to its dependence on balanced training across the dataset.
However, it alone cannot fully address biases from class imbalance, and image-level features lack the rich context of Ground-Truth (GT) labels. To tackle this, we use a teacher-student framework inspired by~\cite{cac} with knowledge distillation. 
The teacher, guided by GT labels, creates a GT-based class center, while the student mimics it using image-level class centers from the memory bank. 
These are concatenated to form an enhanced classifier, approximating GT-level performance during inference when GT labels are unavailable, with learning supervised by KL divergence. 
We also add an intra-variance loss to transfer intra-class feature variance from teacher to student, boosting segmentation performance under imbalanced conditions.
At last, we add a calibration stage that adjusts classification scores using a learnable linear transformation. 
This aligns the classifier output with the true class distribution, dynamically refining logits to improve fairness and accuracy.
\begin{figure}[t]
\begin{center}
\includegraphics[width=1\linewidth, keepaspectratio]{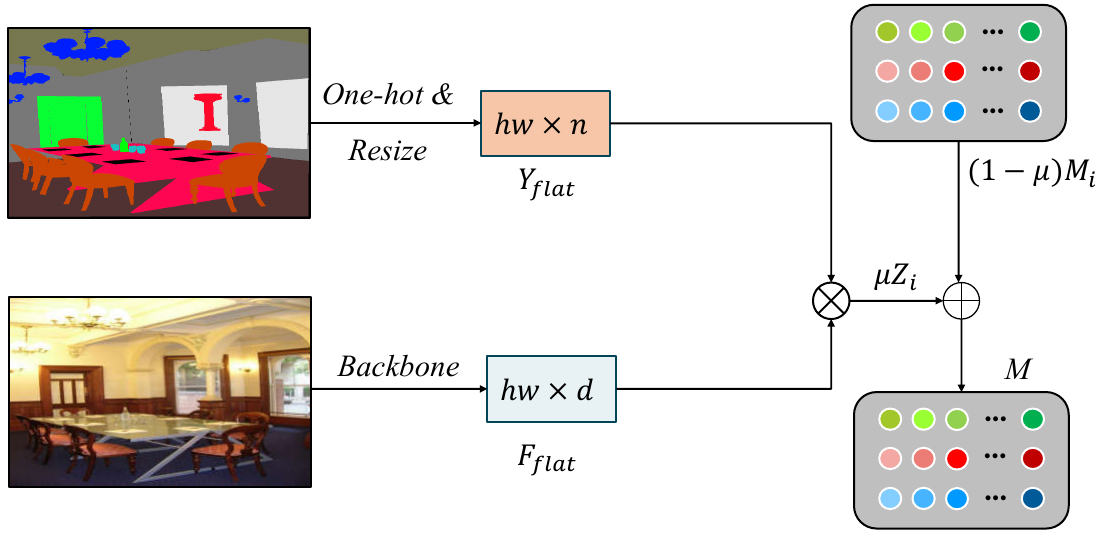}
\end{center}
\vspace{-2mm}
\caption{\textbf{The process of memory bank updating.} The memory bank $\mathcal{M}$ is initialized as an empty structure of size $n\times d$.
We update the memory bank using a momentum-based approach (Eq.\ref{eq:memory_update}), where $i-th$ class representation $\mathcal{M}_{i}$ is refined by blending the previous memory bank state with the newly computed class-specific features $Z_i$.}   
\label{fig:mem}
\end{figure}
\subsection{Memory Bank} 
\label{3.2}
Given an input image $I\in \mathbb{R}^{3\times H\times W}$, an encoder network (\emph{i.e.}, FCN~\cite{fcn} and PSPNet~\cite{pspnet}) first maps pixels into a non-linear embedding space as $F\in \mathbb{R}^{d\times h\times w}$, where $d$, $h$, and $w$ represent the feature dimension, height and width, respectively. 
Then, we create an empty memory bank $\mathcal{M}$ of size $n\times d$ and initialize it by using the values in the masked feature map $Z$ as shown in Figure \ref{fig:mem}, where $n$ is the number of classes.
To update the $i$-th item in the memory bank, we conduct average pooling across the designated region, guided by the segmentation mask corresponding to the $i$-th class, as follows:
\begin{equation}
Z = \frac{Y_{flat}F_{flat}^{T}}{K_i}
\end{equation}
where $K_i$ is the number of pixels belonging to the $i$-th class, $Y_{flat}$ is a resized one-hot segmentation ground truth with the size of $n\times hw$ and $Z \in \mathbb{R}^{n\times d}$ is a masked feature map.

During the training phase, we compute the average feature vector for every class to set up the memory bank $\mathcal{M}$. 
The memory bank is updated after each training iteration using the following equation:
\begin{equation}
\mathcal{M} = (1-\mu) \mathcal{M}_{i} + \mu{Z_{i}} \label{eq:memory_update}
\end{equation}
where $\mu\in\mathbb[0,1]$ is the momentum parameter. $\mathcal{M}_{i}$ denotes the $i$-th class representation and $Z_i$ is corresponding masked class center.
\subsection{Knowledge Distillation}
The memory bank stores dataset-level contextual information, acting as a global classifier, but it lacks the precision of ground truth labels. 
To address this, we adopt a teacher-student framework inspired by~\cite{cac,ssa} for knowledge distillation, designing a Teacher Extended Context-Aware Classifier (Teacher-ECAC) and a Student-ECAC. During training, Teacher-ECAC leverages individual image ground truth segmentation to refine the classifier with precise class information. 
However, since ground truth is unavailable during inference, we use a coarse segmentation from a pre-trained model for Student-ECAC to mimic the teacher’s output. 
This architecture enables effective knowledge transfer from the teacher model, enhancing the student's performance in real-world scenarios.

\subsubsection{Teacher-ECAC ($T_{ecac}$)}
\label{3.3}
To form ECAC, we propose to combine the memory bank $\mathcal{M}$ with the local (image-level) class center. 
Despite the superior performance of recently proposed methods based on the class center, they cannot be compared with ground truth labels. 
Therefore, we propose to use the teacher-student network, which leverages additional ground truth to calculate the class center during training for transferring comprehensive contextual information to the student network.

We first apply one-hot encoding on the ground-truth label $Y$ and resize it as $Y_{flat}\in \mathbb{R}^{n\times hw}$. Then, we calculate the teacher's class center as follows:
\begin{equation}
C_t = \frac{Y_{flat}\times F_{flat}^{T}}{\sum_{i=1}^{hw} Y| (y_i=c_i)}
\end{equation}
where $Y_{flat}$ of size $hw$ stores the ground truth category labels. Then, we concatenate $C_t$ and $\mathcal{M}$ to get the teacher extended context-aware classifier:
\begin{equation}
T_{ecac} = \xi(C_t\oplus \mathcal{M})
\end{equation}
where $\xi$ is a projector head consisting of two linear layers to reduce the channel dimension. $T_{ecac}$ is of size $n\times d$. The final output of the teacher network is given by:
\begin{equation}
O_{t} = T_{ecac}\cdot F
\label{eq6}
\end{equation}
where $O_{t}$ provides more precise information to transfer knowledge to the student network. 

Although the memory bank can learn the mean features of different classes, the imbalanced distribution issue of different datasets still exists.
To tackle this issue, we adopt a unified distribution alignment strategy~\cite{dalign} to calibrate the classifier's output by aligning it with a reference distribution of classes, promoting balanced predictions. 

\subsubsection{Student-ECAC ($S_{ecac}$)}
\label{3.4}
For Student-ECAC, $F$ is first divided into $n$ class regions (coarse segmentation result) as $F_{coarse}\in \mathbb{R}^{n\times h\times w}$. 
Next, we get the flattened pixel representations $F_{flat}\in \mathbb{R}^{d\times hw}$ and flattened coarse segmentation result $F'_{coarse}\in \mathbb{R}^{n\times hw}$. 
We calculate the class center as follows:
\begin{equation}
C_s = \mathrm{softmax}(F'_{coarse})\times F_{flat}^{T}
\end{equation}
Here, $C_s\in\mathbb{R}^{n\times d}$ is the average features of all pixel features to a category. Then, we concatenate $C_s$ and $\mathcal{M}$ to obtain student extended context-aware classifier:
\begin{equation}
S_{ecac} = \xi(C_s\oplus \mathcal{M})
\end{equation}
where $\xi$ is also a projector head a projector head consisting of two linear layers to reduce the channel dimension. $S_{ecac}$ is of size $n\times d$.
$\mathcal{M}$ is the memory bank of size $n\times d$ storing the dataset's mean features of different classes. $n$ is the number of classes. Note that $S_{ecac}$ serves as a more comprehensive classifier, yielding improved prediction results during inference, as shown in Equation \ref{eq5}.
\begin{equation}
O_{s} = S_{ecac}\cdot F
\label{eq5}
\end{equation}
where $Z_{s}$ is the final output of the student network. 

\subsubsection{Output Calibration for Enhancing Segmentation}
We apply a calibration step to refine the final outputs of the teacher and student networks, addressing class imbalance and improving segmentation accuracy. For the teacher network, this calibration is defined as follows:
\begin{equation}
\tilde{O}_{t,i} = \gamma_i \cdot {O}_{t,i} + \delta_i, \quad \forall i \in \mathcal{C}
\end{equation}
where $\tilde{O}_{t,i}$ represents the logits produced by the teacher classifier $T_{ecac}$ (Equation~\ref{eq6}). $\gamma_i$ is a learnable class-specific scaling factor that adjusts the magnitude of the logits for each class $i$, and $\delta_i$ is a learnable class-specific offset that shifts the decision boundary, both of which will be learned from data. 
Similarly, for the student network, the output is calibrated as shown:
\begin{equation}
\tilde{O}_{s,j} = \gamma_j \cdot {O}_{s,j} + \delta_j, \quad \forall j \in \mathcal{C}
\end{equation}
Here, $\tilde{O}_{s,j}$ denotes the logits from the student classifier $S{ecac}$ (Equation~\ref{eq5}). 
The learnable parameters $\gamma_{j}$ and $\delta_{j}$ are specifically tailored for the student model and differ from those used in the teacher model.
This output calibration serves as a critical refinement step following the teacher-student distillation process. 
By applying class-specific adjustments to the logits with parameters unique to the student, it better mitigates the bias toward majority classes that often arises due to imbalanced class distributions in semantic segmentation datasets.

\subsection{Loss function}
\subsubsection{Knowledge Distillation}
Knowledge distillation enhances the performance of small models by leveraging knowledge transferred from large models~\cite{hinton}. 
Using knowledge distillation, the student model attains remarkable accuracy without compromising efficiency.
Inspired by this, we utilize KL divergence to force the student classifier $T_{ecac}$ to mimic the teacher classifier $S_{ecac}$, where the loss is defined as:
\begin{equation}
L_{KL} = -\frac{1}{N}\sum\limits_{i=1}^{N}\sum\limits_{c=1}^{C}{\tilde{O}_{t}}^i\cdot log({\tilde{O}_{s}}^i)
\label{eq9}
\end{equation}
where $N$ denotes the total number of pixels in the image and $C$ represents the number of classes.
Following previous work~\cite{cac}, we incorporate entropy $H$ into the KL loss to capture the distributional information of each pixel:

\begin{equation}
L_{KL} = -\frac{1}{C}\sum\limits_{k=1}^{C}\frac{\sum\limits_{i=1}^{N}\sum\limits_{c=1}^{C}{B^i_kH^i{\tilde{O}_{t}}}^i\cdot log({\tilde{O}_{s}}^i)}{\sum\limits_{i=1}^{N}B^i_kH^i}
\end{equation}
where $B_k$ is a binary mask representing the presence of the $k$-th class.
\subsubsection{Re-weighting Cross-Entropy Loss}
To tackle the issue of imbalanced distribution, we apply a re-weighting strategy~\cite{cui2019class} to the cross-entropy loss.
Specifically, we propose a calibration strategy that aligns the model's predictions with a reference class distribution designed to encourage balanced predictions. Then, the re-weighting loss can be written as:

\begin{equation}
L_{rce} = w_c \cdot L_{ce}
\end{equation}

where $w_c$ is the class weight and $L_{ce}$ is cross-entropy loss. We formulate the reference weights as a distribution derived from the empirical class frequencies $\mathbf{f} = [f_1, \dots, f_K]$, calculated using the training set.

\begin{equation}
w_c = \frac{(1 / f_c)^\rho}{\sum_{k=1}^K (1 / f_k)^\rho}, \quad \forall c \in \mathcal{C}
\end{equation}
where $\rho$ serves as a scaling hyper-parameter, offering greater flexibility in representing the class prior.
\subsubsection{Intra-Class Variance}
Increasing intra-class compactness can lead to a wider margin between classes, thereby yielding features that are easier to distinguish. 
To reduce intra-class variance, we use the map of output ($O_s, O_t$) similarity between pixels and their class centers. 
First, we compute the class center for each class by taking the mean of the output ($O_s, O_t$) for all pixels in the class. 
Then, we calculate the cosine similarity between each pixel and its corresponding class center. 
The intra-class variance of student and teacher is defined as:
\begin{equation}
D_{intraS} = \cos{({\tilde{O}_{s}}, CenterSeg_s)}
\end{equation}
\begin{equation}
D_{intraT} = \cos{({\tilde{O}_{t}}, CenterSeg_t)}
\end{equation}
where $Seg_s$ and $Seg_t$ denote the outputs of the student and teacher classifier, and $cos$ denotes the cosine similarity function. 
$CenterSeg_s$ and $CenterSeg_t$ stand for the class center of output as follows:
\begin{equation}
CenterSeg_s =\frac{1}{Nc}\sum\limits_{i=1}^{N_c}{\tilde{O}_{s,i}}
\end{equation}
\begin{equation}
CenterSeg_t =\frac{1}{Nc}\sum\limits_{i=1}^{N_c}{\tilde{O}_{t,i}}
\end{equation}
where $N_c$ represents the number of pixels sharing the same label. To address large discrepancies between the intra-class variance of the student and teacher outputs, we employ a straightforward yet effective Mean Squared Error loss. The intra-class variance loss is formulated as:
\begin{equation}
L_{IV} = f_{mse}(D_{intraS}, D_{intraT})
\end{equation}
 Apart from $L_{KL}$ and $L_{iv}$, cross-entropy loss is applied to $Seg_s$ and $Seg_t$, defined as $L_{ceS}$ and $L_{ceT}$. 
 \subsubsection{Memory Bank Updating}
 We propose a memory loss for updating the memory bank:
 \begin{equation}
L_\mathcal{M} = \frac{1}{hw}\sum\limits_{k=1}^{hw}-Y_m[k]\cdot log(\mathrm{softmax}(C_m))
\end{equation}
 where $Y_m$ is the ground-truth map.  
 To ensure robust initialization of the memory bank, especially for rare or initially absent classes, we delay the computation of \( L_\mathcal{M} \) until the first \textbf{1,000 training iterations} are completed. This design choice is motivated by two key factors:

\begin{itemize}
    \item \textbf{Feature Stability:} In early training stages, model parameters and feature representations are rapidly evolving. Delaying \( L_\mathcal{M} \) allows the network to learn coarse yet stable features before optimizing the memory bank, avoiding premature updates based on noisy representations.
    
    \item \textbf{Class Coverage:} For long-tailed datasets, some classes may appear infrequently or not at all in initial batches. By postponing \( L_\mathcal{M} \), the memory bank accumulates sufficient samples for all classes through multiple updates (Eq.~\eqref{eq:memory_update}), ensuring balanced feature initialization.
\end{itemize}
 
\noindent Hence, the overall loss is defined as:
\begin{equation}
L =  L_{rceS} +  L_{rceT} + L_\mathcal{M} + \alpha L_{KL} + \beta L_{iv}
\end{equation}
where $\alpha$ is set to 1 and $\beta$ is set to 50. \( L_{\mathcal{M}} \) is activated only after 1,000 iterations. 

\subsection{Integrating with other methods}
As shown in Figure~\ref{fig:arc}, ECAC seamlessly integrates with Existing Segmentation Models, including FCN \cite{fcn}, PSPNet \cite{pspnet}, DeeplabV3+ \cite{deeplabv3+}, and UperNet \cite{upernet}, achieving significant performance improvements. While these methods emphasize image-level context aggregation, ECAC introduces dataset-level contextual information, further enhancing segmentation performance.

To integrate our module into existing methods, we need to compute class centers based on the feature outputs of the current image. These centers are then combined with the memory bank to construct a dataset-level classifier. Simultaneously, ground truth is used to create a teacher-ECAC for knowledge distillation.

\begin{table*}[!t]
\renewcommand\arraystretch{1}
\setlength{\tabcolsep}{4mm}
\centering
{\footnotesize
\begin{tabular}{p{1mm}|c|c|cccccc} \hline
\toprule
\multirow{2}{*}{} & \multirow{2}{*}{Method} & \multirow{2}{*}{Backbone} & \multicolumn{6}{c}{mIoU (\%)} \\ \cline{4-9}
 &  &  & \multicolumn{2}{c}{ADE20K} & \multicolumn{2}{c}{COCO-Stuff10k} & \multicolumn{2}{c}{Pascal-Context} \\ \cline{4-9}
 &  &  & s.s. & m.s. & s.s. & m.s. & s.s. & m.s. \\ \hline
\multirow{8}{*}{\rotatebox{90}{CNN backbones}} & PSPNet~\cite{pspnet} & ResNet-101 & \transparent{0.7}44.39 & 45.35 & \transparent{0.7}37.76 & 38.86 & \transparent{0.7}52.47 & 53.50 \\ 
& CPNet~\cite{context} & ResNet-101 & \transparent{0.7}--& 46.27 & \transparent{0.7}-- & -- & \transparent{0.7}-- & 53.90\\ 
& DeeplabV3plus~\cite{deeplabv3+} & ResNet-101 & \transparent{0.7}45.47 & 46.35 & \transparent{0.7}-- & 39.00 & \transparent{0.7}53.2 & 54.67\\
& CSFLM~\cite{csflm} & ResNet-101 & \transparent{0.7}-- & 46.65 & \transparent{0.7}-- & 41.30 & \transparent{0.7}-- & 55.90\\
& PCAA~\cite{partial} & ResNet-101 & \transparent{0.7}-- & 46.74 & \transparent{0.7}-- & -- & \transparent{0.7}-- & 55.60\\  
& ISNet~\cite{isnet} & ResNet-101 & \transparent{0.7}45.92 & 47.31 & \transparent{0.7}40.53 & 41.60 & \transparent{0.7}--& --\\ 
& CPT-M~\cite{CPT} & ResNet-101 & \transparent{0.7}45.6 & 47.00 & \transparent{0.7}-- & -- & \transparent{0.7}-- & 56.30\\ 
& CAA~\cite{tang2024caa} & ResNet-101 & \transparent{0.7}46.1 & 47.45 & \transparent{0.7}40.0 & 41.00 & 54.32 & 55.57\\ \hline
& \multirow{2}{*}{\textbf{DeeplabV3plus + ECAC (\emph{ours})}} & ResNet-50 & \transparent{0.7}46.31 & 47.67 & \transparent{0.7}40.19 & 40.93 & \transparent{0.7}53.47 & 54.99 \\ 
& & ResNet-101 & \transparent{0.7}47.11 & \textbf{48.93} & \transparent{0.7}40.84 & \textbf{42.53} & \transparent{0.7}55.06 & \textbf{57.44} \\ \hline
\midrule 
\multirow{13}{*}{\rotatebox{90}{Transformer backbones}} & StructToken~\cite{StructToken} & ViT-Base & \transparent{0.7}50.72 & 51.85 & \transparent{0.7}-- & -- & \transparent{0.7}-- &--\\
& SETR-MLA~\cite{setr} & ViT-Large & \transparent{0.7}-- & {50.28} & \transparent{0.7}-- & -- & \transparent{0.7}-- & 55.83\\ 
& MCIBI~\cite{mcibi} & ViT-Large & \transparent{0.7} 49.73 & {50.80} &\transparent{0.7} 44.27 & {44.89} & \transparent{0.7}-- & --\\ 
& SegNeXt~\cite{segnext} & MSCAN-L & \transparent{0.7}50.22 & 51.28 & \transparent{0.7}43.7 & 44.77 & \transparent{0.7}57.99 &  59.49\\ 
& UperNet~\cite{internimage} & InternImage-B & \transparent{0.7}50.8 & 51.30 & \transparent{0.7}44.82 & 45.30&\transparent{0.7}60.12 & 61.34\\ \cmidrule{2-9}
& \textbf{SegNeXt + ECAC (\emph{ours})} & MSCAN-L~\cite{segnext} & \transparent{0.7}50.54 & 51.74 &\transparent{0.7}43.92& 45.72 &\transparent{0.7}58.79 & 60.19\\ 
& \textbf{UperNet + ECAC (\emph{ours})} & InternImage-B~\cite{internimage} & \transparent{0.7}51.24 & \textbf{52.21} & \transparent{0.7}45.33 & \textbf{46.32} & \transparent{0.7}60.92 & \textbf{61.75}\\ 
\cmidrule{2-9}
& SemCL~\cite{semcl} & Swin-Base & \transparent{0.7}-- & 48.68 & \transparent{0.7}-- & -- & \transparent{0.7}-- & --\\
& UperNet+CAC~\cite{cac} & Swin-Base & \transparent{0.7}52.00 & 53.52 & \transparent{0.7}-- & -- & \transparent{0.7}60.78 & 62.16 \\
& GSS-FT-W~\cite{gss} & Swin-Large &\transparent{0.7}-- & 48.54 & \transparent{0.7}-- & -- & \transparent{0.7}-- & --\\
& CoT~\cite{cot} & Swin-Large & \transparent{0.7}53.01 & 54.16 & \transparent{0.7}-- & -- & \transparent{0.7}55.73 & 57.21\\ 
& TSG~\cite{tsg} & Swin-Large & \transparent{0.7}-- & 54.20 &\transparent{0.7}--& -- & \transparent{0.7}-- & 63.30\\
& UperNet + SSA~\cite{ssa} & Swin-Large & \transparent{0.7}52.69 & 54.54 & \transparent{0.7}48.94 & 50.25 & \transparent{0.7}61.83 &64.25\\ \midrule
& \multirow{2}{*}{\textbf{UperNet + ECAC (\emph{ours})}} & Swin-Base & \transparent{0.7}52.12 & 53.69 &\transparent{0.7}46.82 & 48.19 &\transparent{0.7}61.34 & 63.64\\ 
&  & Swin-Large &\transparent{0.7}52.96 & \textbf{54.70} &\transparent{0.7}49.0 & \textbf{50.49} & \transparent{0.7}61.92 &\textbf{64.50}\\ 
\bottomrule
\end{tabular}
}
\vspace{-1mm}
\caption{Comparisons with state-of-the-art methods are conducted using single-scale (s.s.) and multi-scale (m.s.) and on the ADE20K val set, COCO-Stuff10K test set, and Pascal-Context test set, with mIoU as the evaluation metric.} \label{tab}
\end{table*}

\section{Experiments}
This section begins by introducing the benchmark datasets and outlining the implementation details of our method. Subsequently, we present the ECAC results across various semantic segmentation datasets. Finally, we conduct ablation studies to evaluate the influence of key components in ECAC.
\subsection{Datasets}

\noindent \textbf{ADE20K.} ADE20K~\cite{ade20k} is a challenging segmentation dataset, which contains more than 20,000 images. The dataset comprises  150 semantic categories, \emph{i.e.}, water, buildings, vegetation, and discrete objects such as animals, bicycles, and furniture. It is divided into 20k/2k/3k images for training, validation, and testing.

\noindent \textbf{COCO-Stuff10k.} The COCO-Stuff10k~\cite{coco} dataset includes 171 semantic classes, with 81 thing classes and 91 stuff classes. It contains 9k training images and 1k testing images.
\noindent \textbf{PASCAL-Context.} The PASCAL-Context~\cite{pascalcontext} is a widely used semantic segmentation dataset with 10,100 images, including 4,996 for training and 5,104 for testing, labeled across 59 categories.
\subsection{Implementation Details}
\noindent\textbf{Model Settings.} We conduct all the experiments using MMsegmentation~\cite{mmseg2020} and follow the standard training configurations. 
We employ ImageNet pre-trained ResNet and Swin-Transformer as backbones for our method. 
Each model is trained on dual NVIDIA L40 GPUs with 48GB of memory.

\noindent\textbf{Training Settings.} We optimize the network using stochastic gradient descent (SGD) with a momentum of 0.9, a polynomial learning rate scheduler defined as $(1-\frac{iter}{iter_{max}})^{0.9}$, and synchronized batch normalization (SyncBN).
During testing, we employ multi-scale ratios from 0.5 to 1.75 and incorporate data augmentation techniques such as flipping and random cropping.
Detailed settings are provided for each benchmark. 
\begin{itemize}
\item ADE20K: The initial learning rate is set to 0.01, with a crop size of $512\times512$ and a weight decay of 0.0005. By default, training runs for 130 epochs with a batch size of 16.
\item COCO-Stuff10K: The initial learning rate is 0.001, crop size is $512\times512$, and weight decay is 0.0001. Unless specified, we train for 110 epochs with a batch size of 16.
\item PASCAL-Context: We use an initial learning rate of 0.001, crop size of $480\times480$, and a weight decay of 0.0001. If not specified, training consists of 110 epochs with a batch size of 16.
\end{itemize}
\begin{figure*}[]
    \centering
    \begin{minipage}[b]{0.245\textwidth}
        \centering
        \includegraphics[width=4.45cm,height=1.9cm]{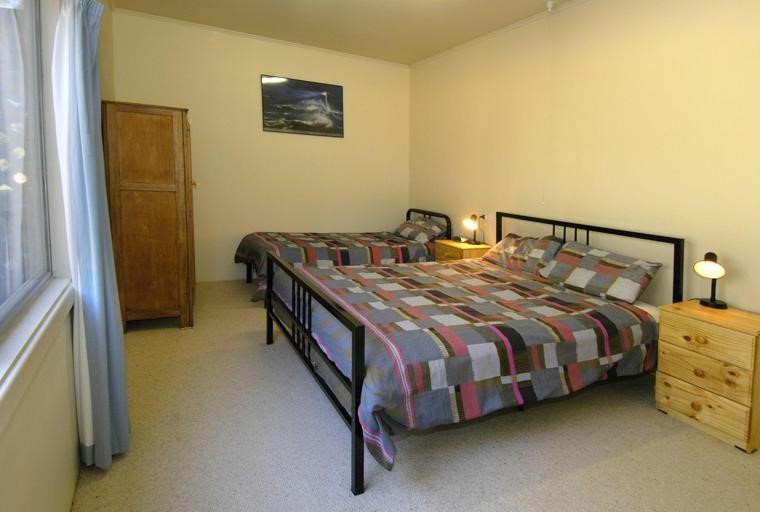}\vspace{1pt}
        \includegraphics[width=4.45cm,height=1.9cm]{vis/img2.jpg}\vspace{1pt}
        \includegraphics[width=4.45cm,height=1.9cm]{vis/img3.jpg}\vspace{1pt}
        \includegraphics[width=4.45cm,height=1.9cm]{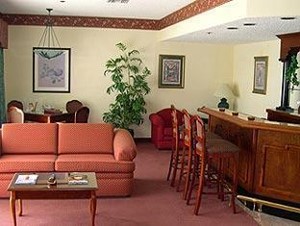}\vspace{1pt}
        \includegraphics[width=4.45cm,height=1.9cm]{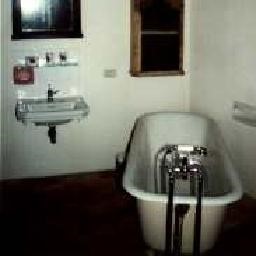}\vspace{1pt}
        \includegraphics[width=4.45cm,height=1.9cm]{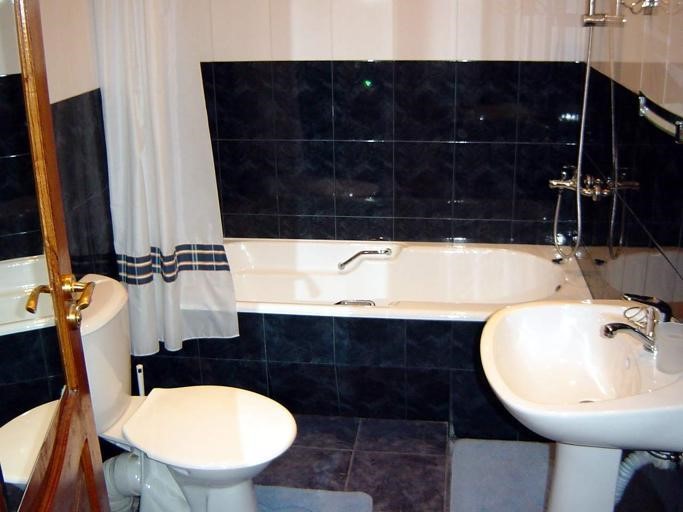}
        \caption*{(a) Image}
    \end{minipage}
    \begin{minipage}[b]{0.245\textwidth}
        \centering
        \includegraphics[width=4.45cm,height=1.9cm]{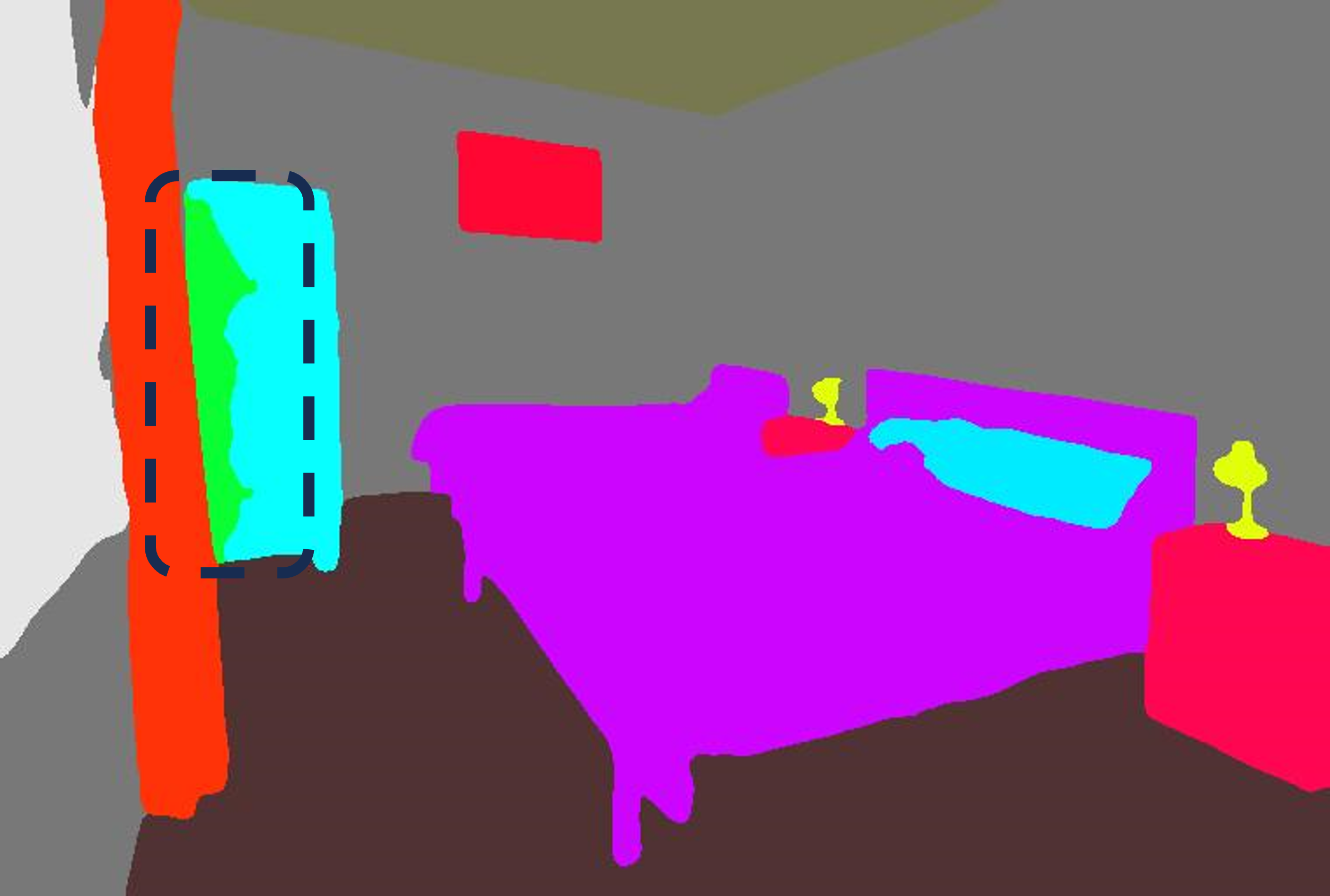}\vspace{1pt}
        \includegraphics[width=4.45cm,height=1.9cm]{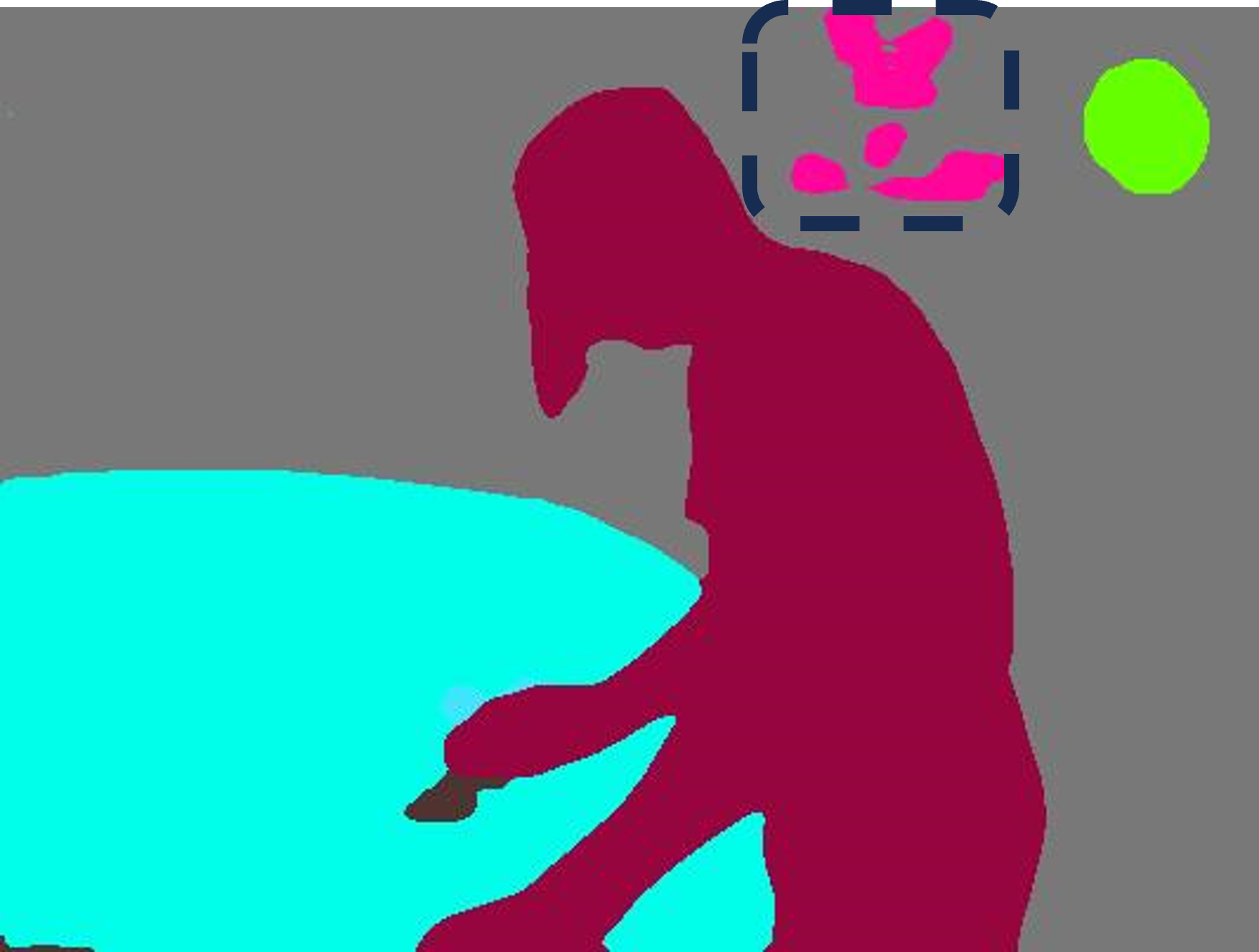}\vspace{1pt}
        \includegraphics[width=4.45cm,height=1.9cm]{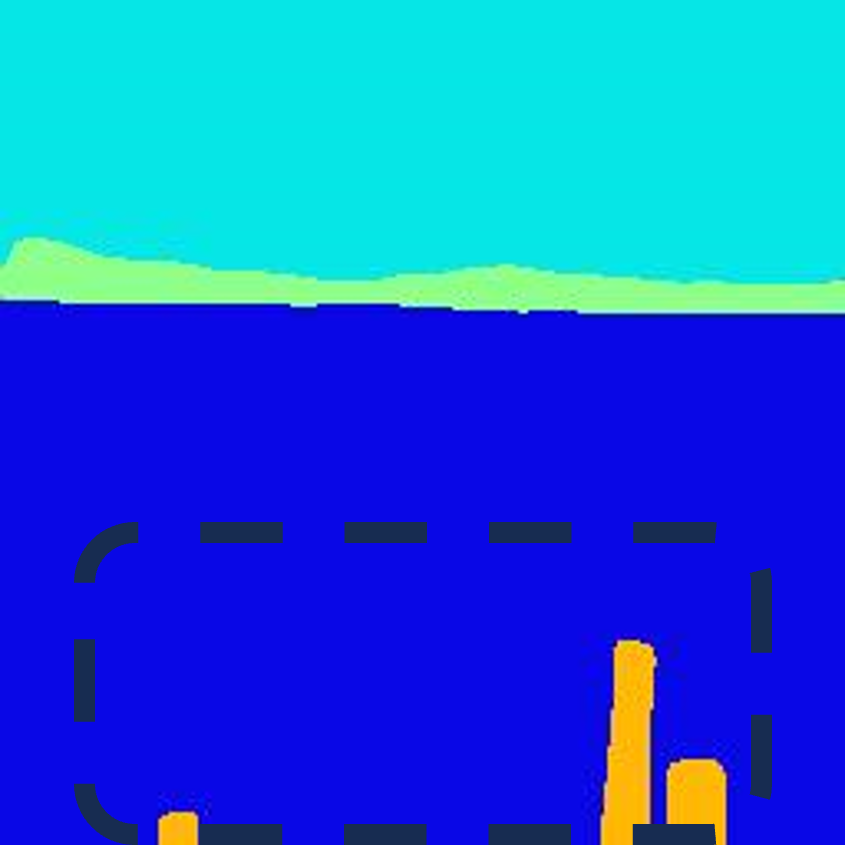}\vspace{1pt}
        \includegraphics[width=4.45cm,height=1.9cm]{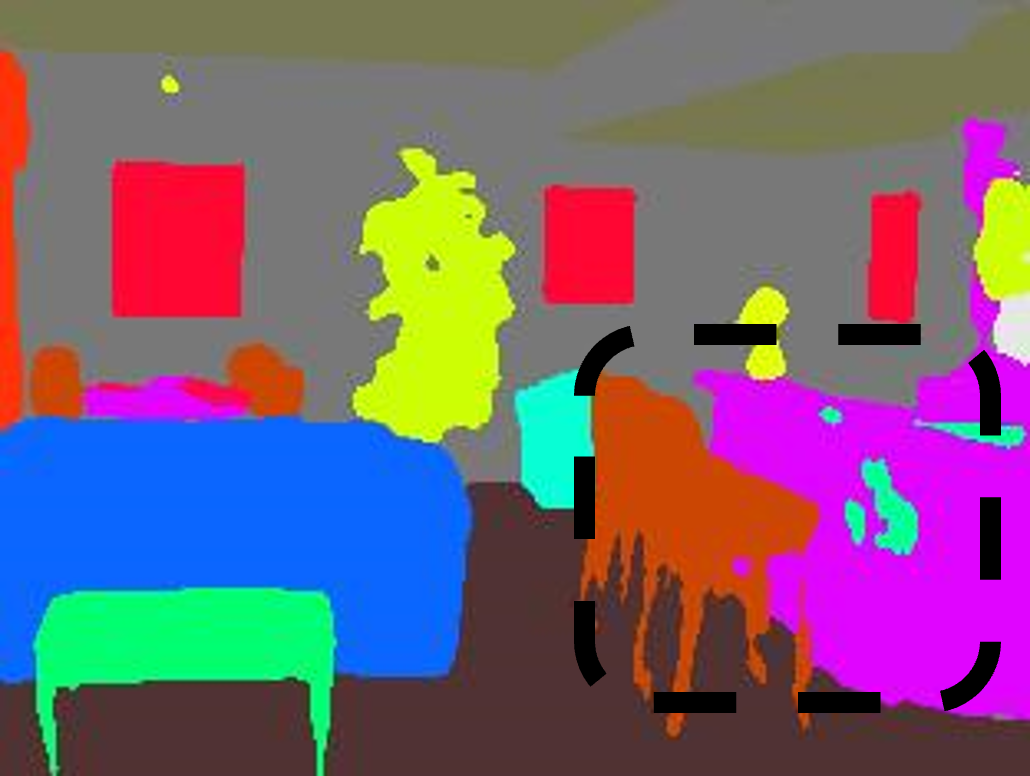}\vspace{1pt}
        \includegraphics[width=4.45cm,height=1.9cm]{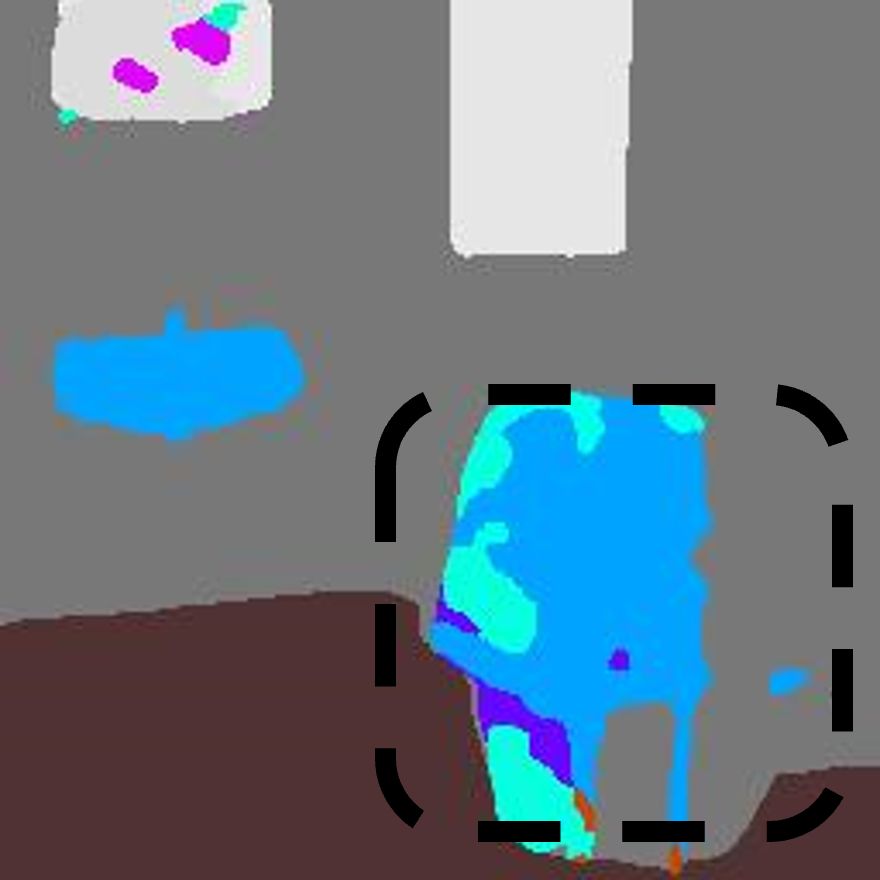}\vspace{1pt}
        \includegraphics[width=4.45cm,height=1.9cm]{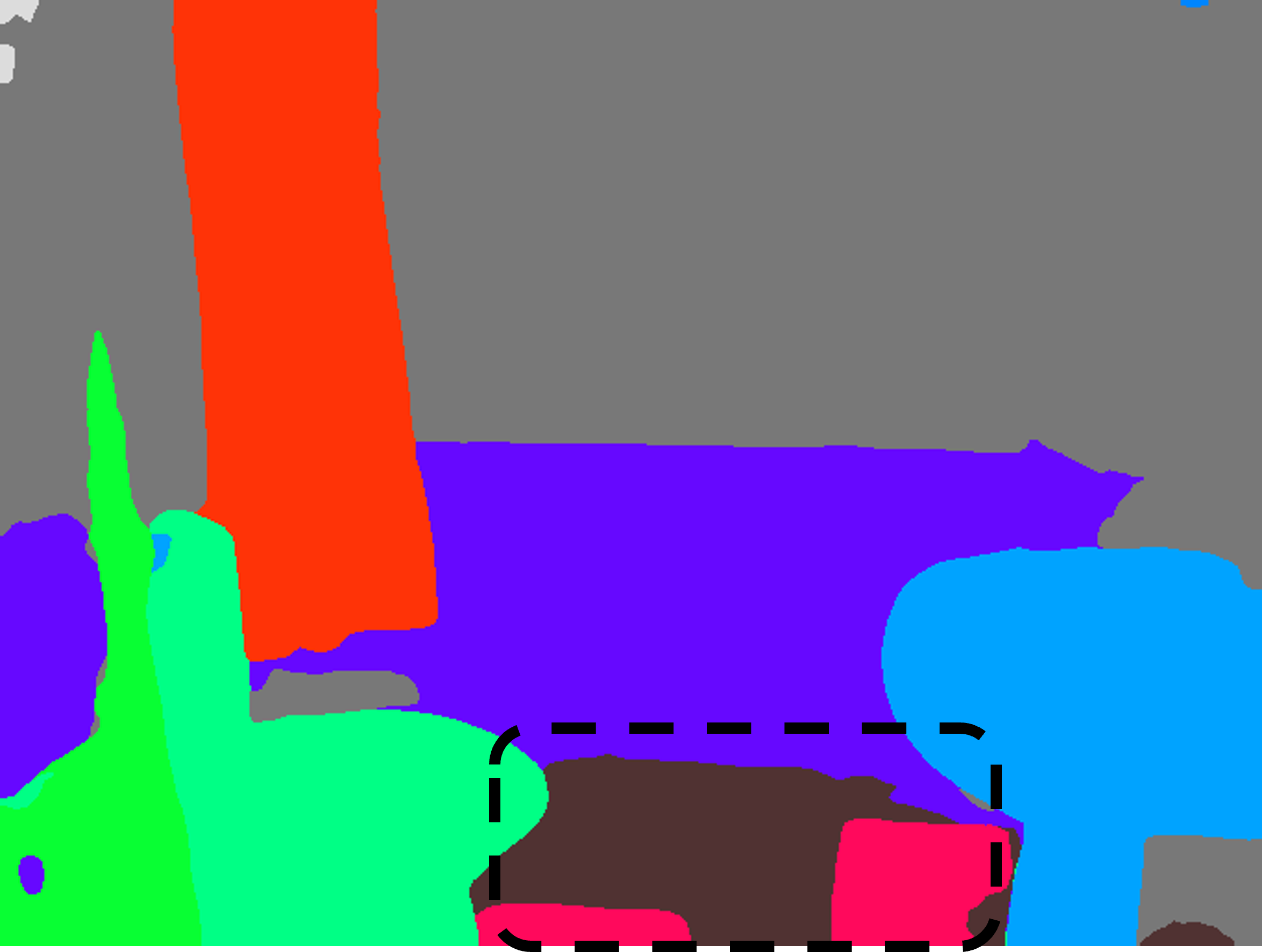}
        \caption*{(b) DLV3plus}
    \end{minipage}
    \begin{minipage}[b]{0.245\textwidth}
        \centering
        \includegraphics[width=4.45cm,height=1.9cm]{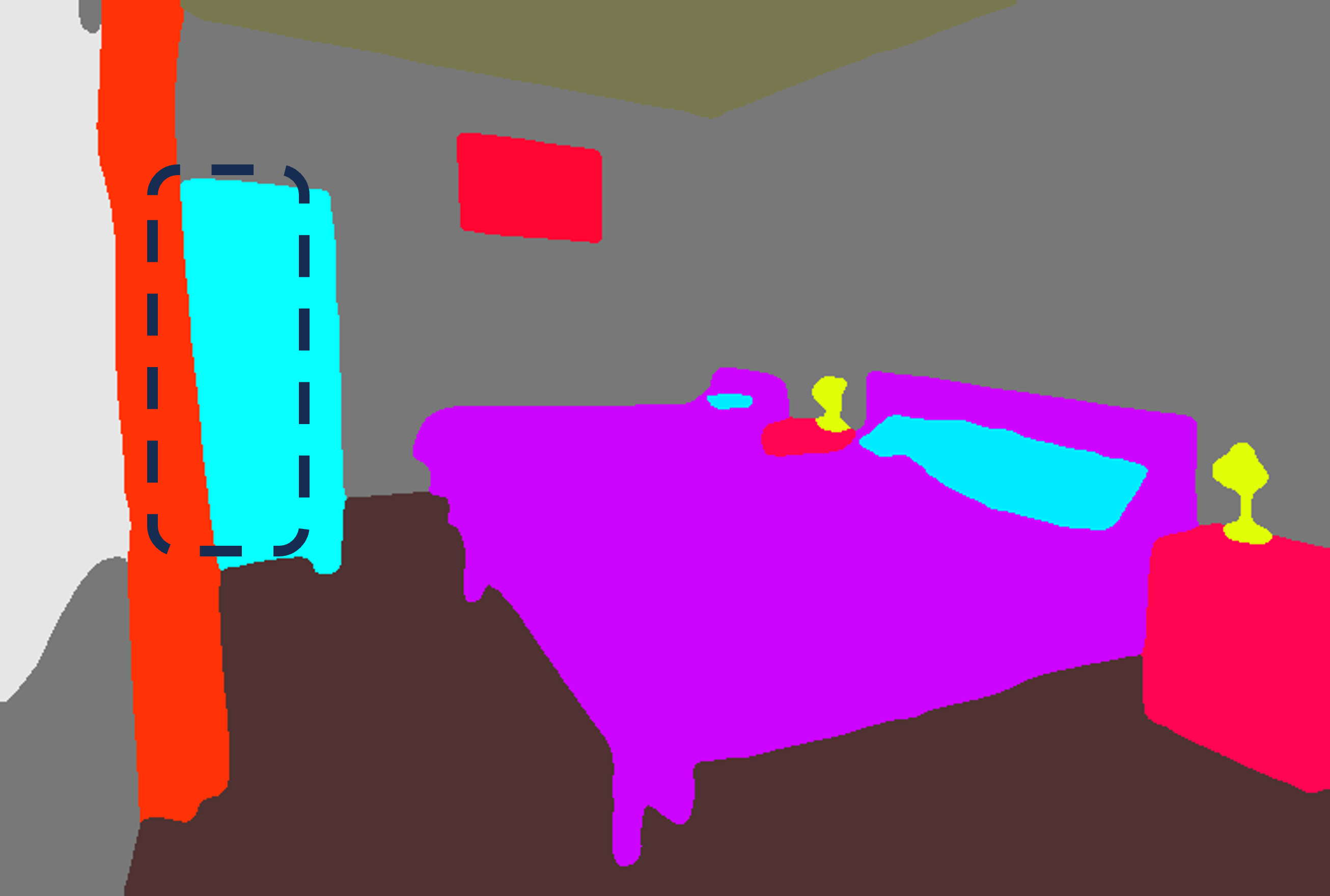}\vspace{1pt}
        \includegraphics[width=4.45cm,height=1.9cm]{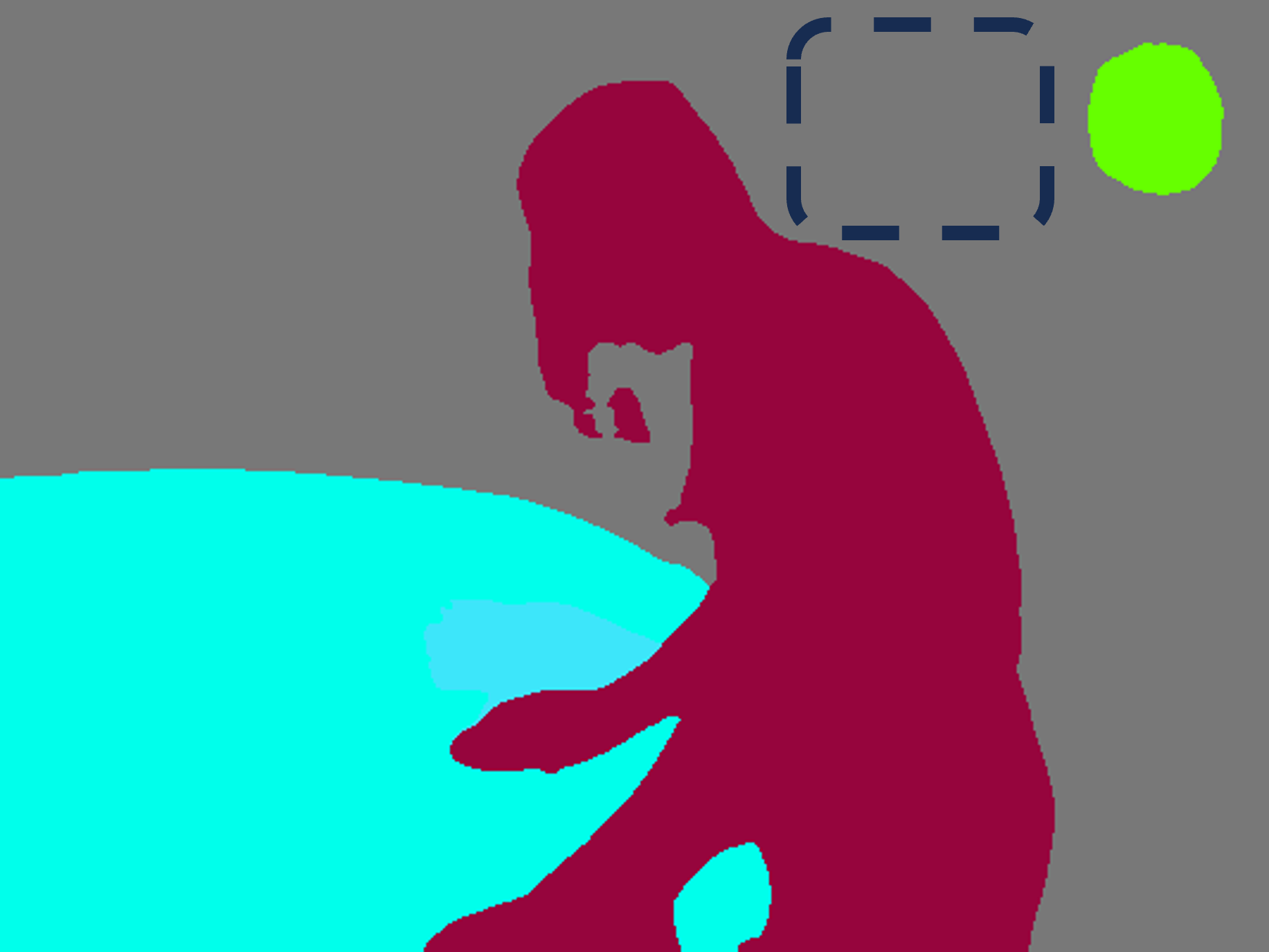}\vspace{1pt}
        \includegraphics[width=4.45cm,height=1.9cm]{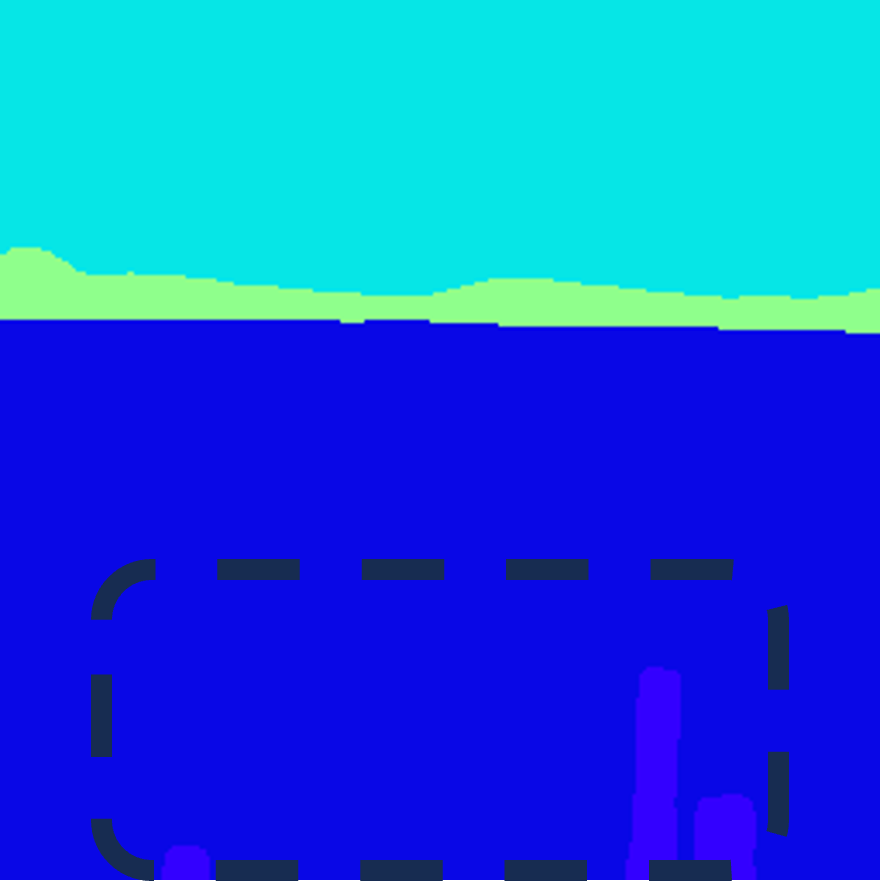}\vspace{1pt}
        \includegraphics[width=4.45cm,height=1.9cm]{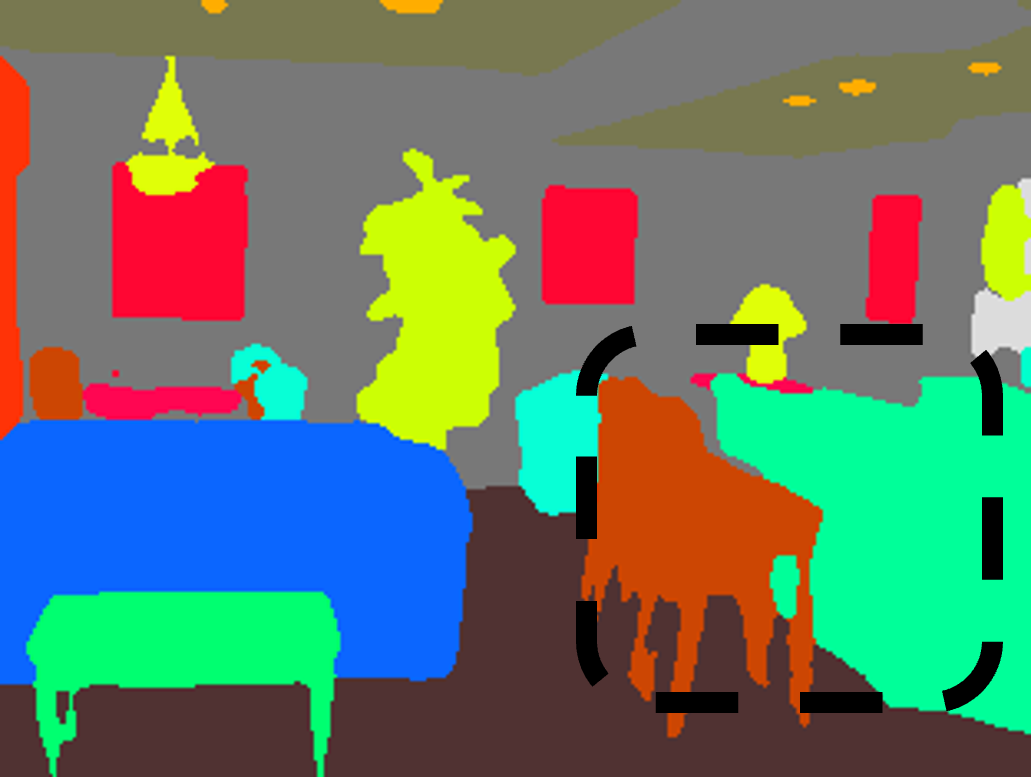}\vspace{1pt}
        \includegraphics[width=4.45cm,height=1.9cm]{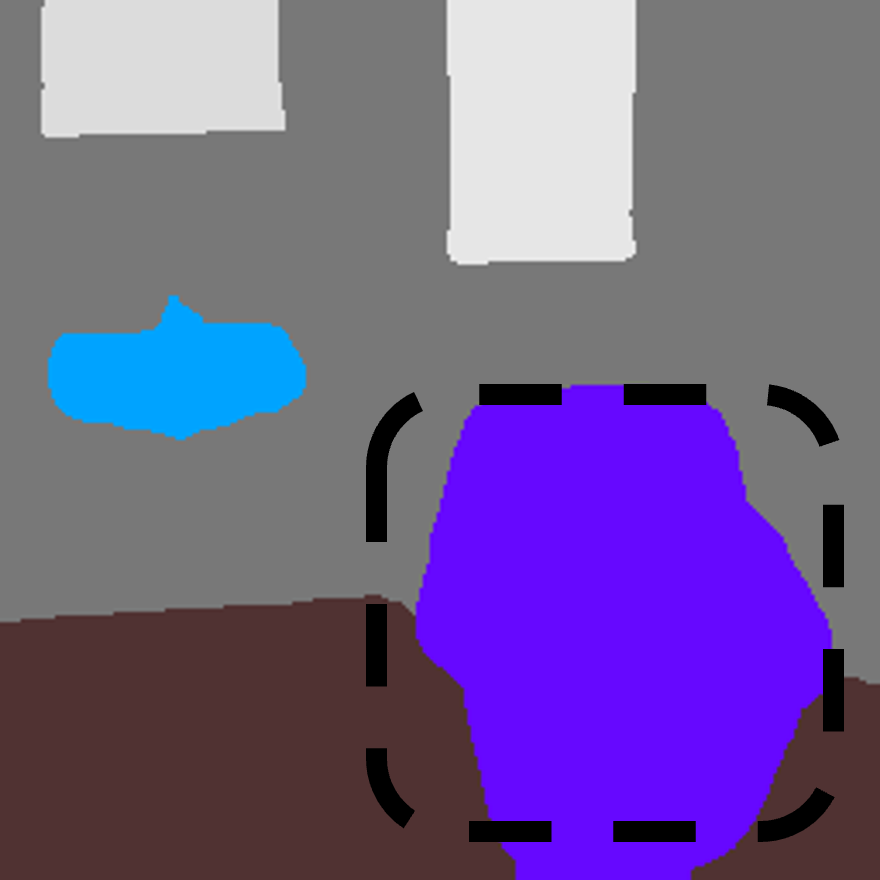}\vspace{1pt}
        \includegraphics[width=4.45cm,height=1.9cm]{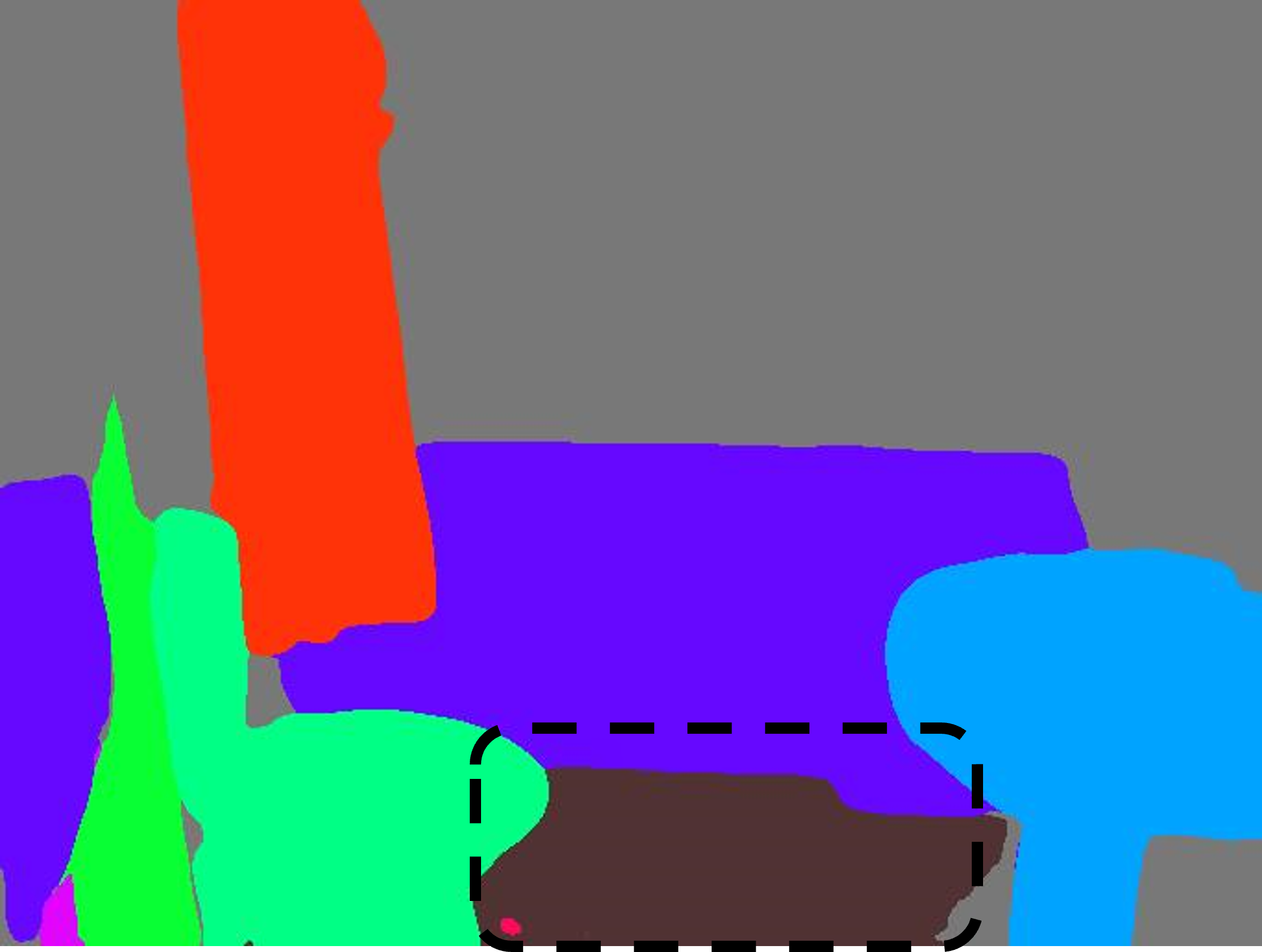}
        \caption*{(c) DLV3plus+ECAC(ours)}
    \end{minipage}
    \begin{minipage}[b]{0.245\textwidth}
        \centering
        \includegraphics[width=4.45cm,height=1.9cm]{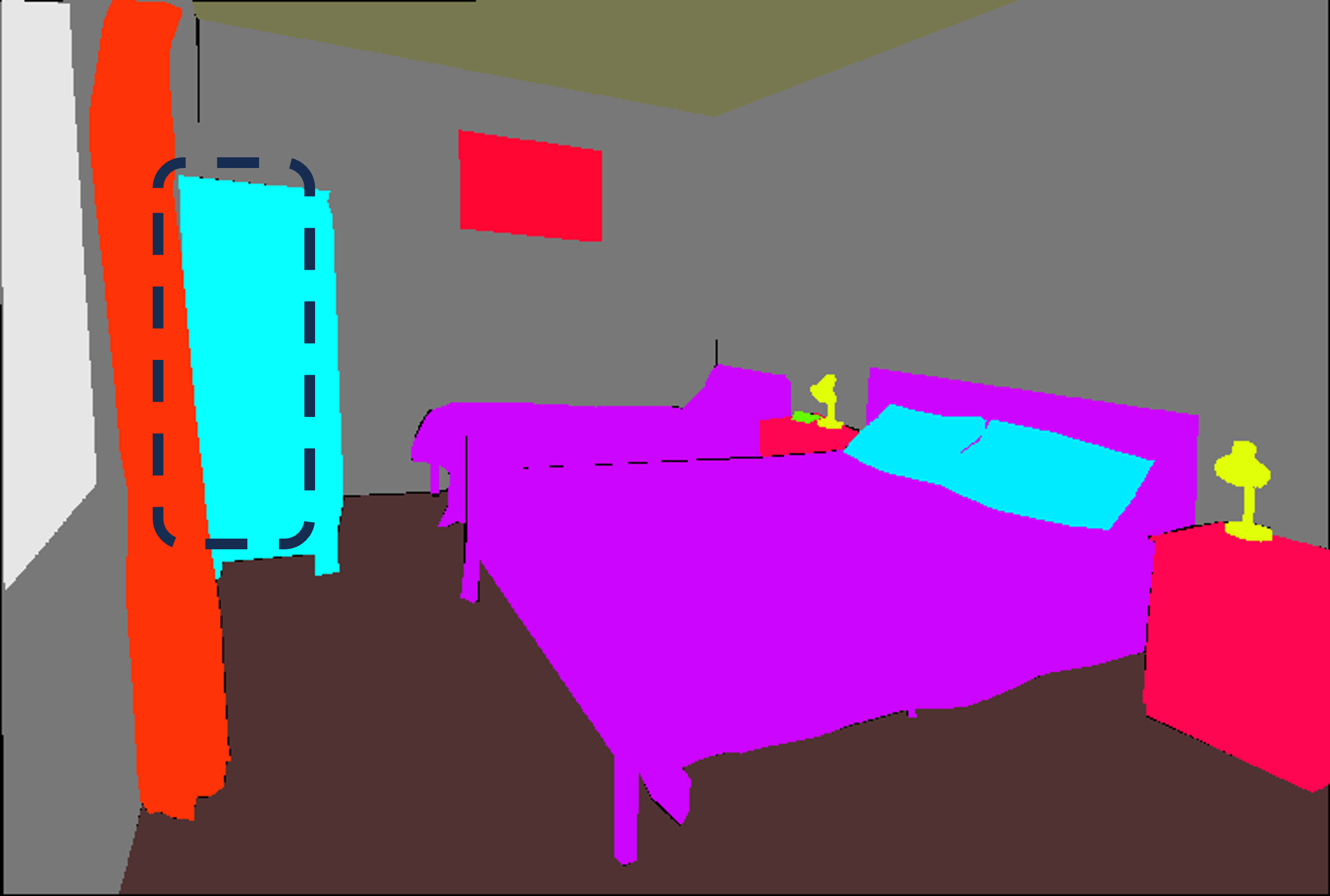}\vspace{1pt}
        \includegraphics[width=4.45cm,height=1.9cm]{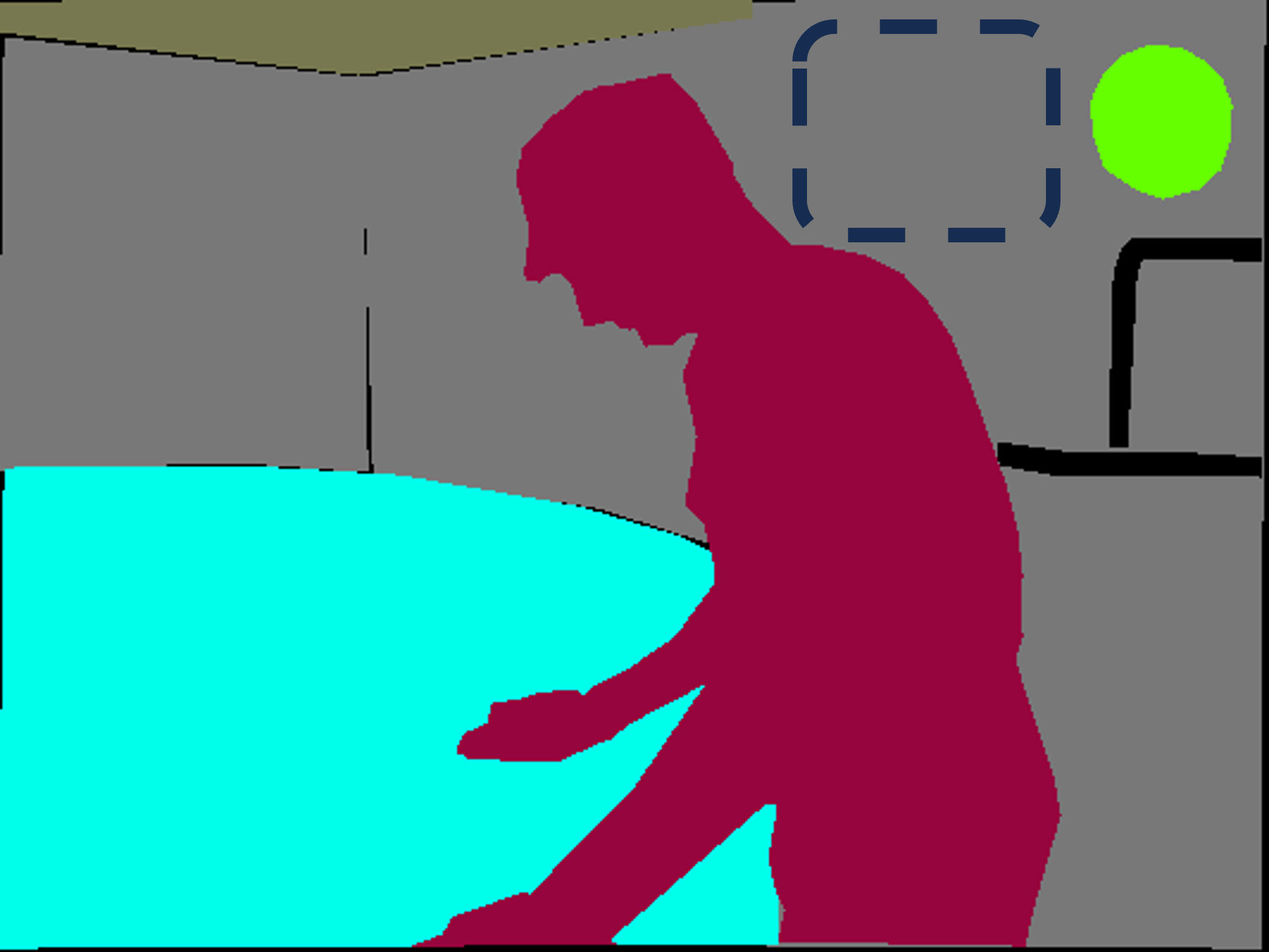}\vspace{1pt}
        \includegraphics[width=4.45cm,height=1.9cm]{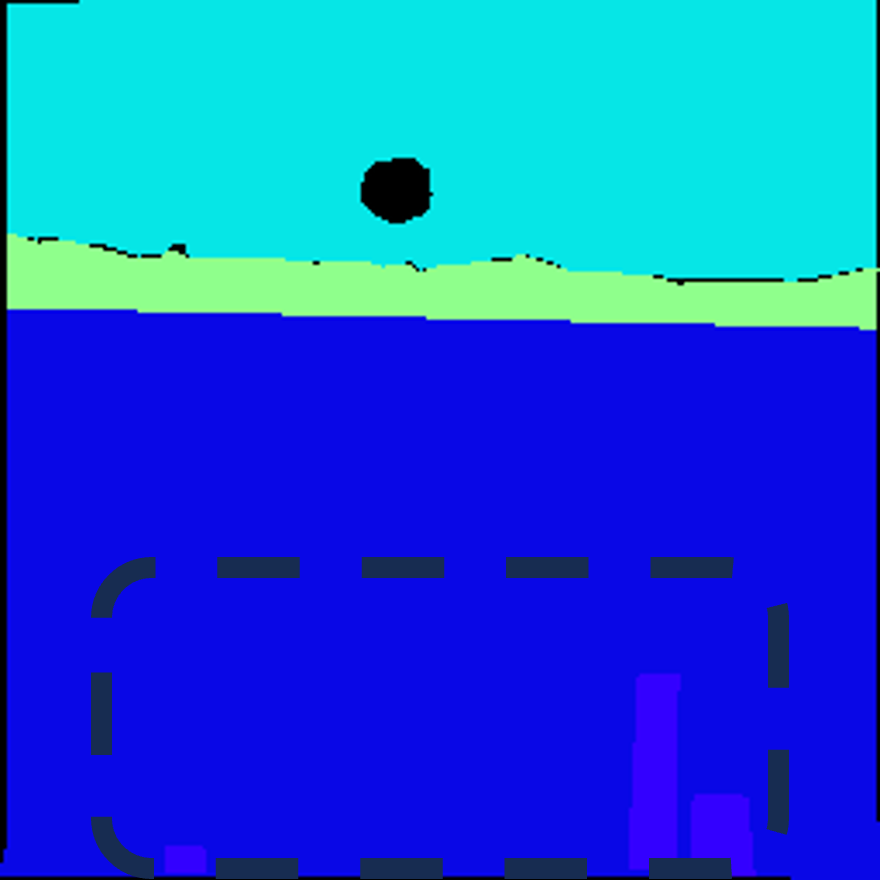}\vspace{1pt}
        \includegraphics[width=4.45cm,height=1.9cm]{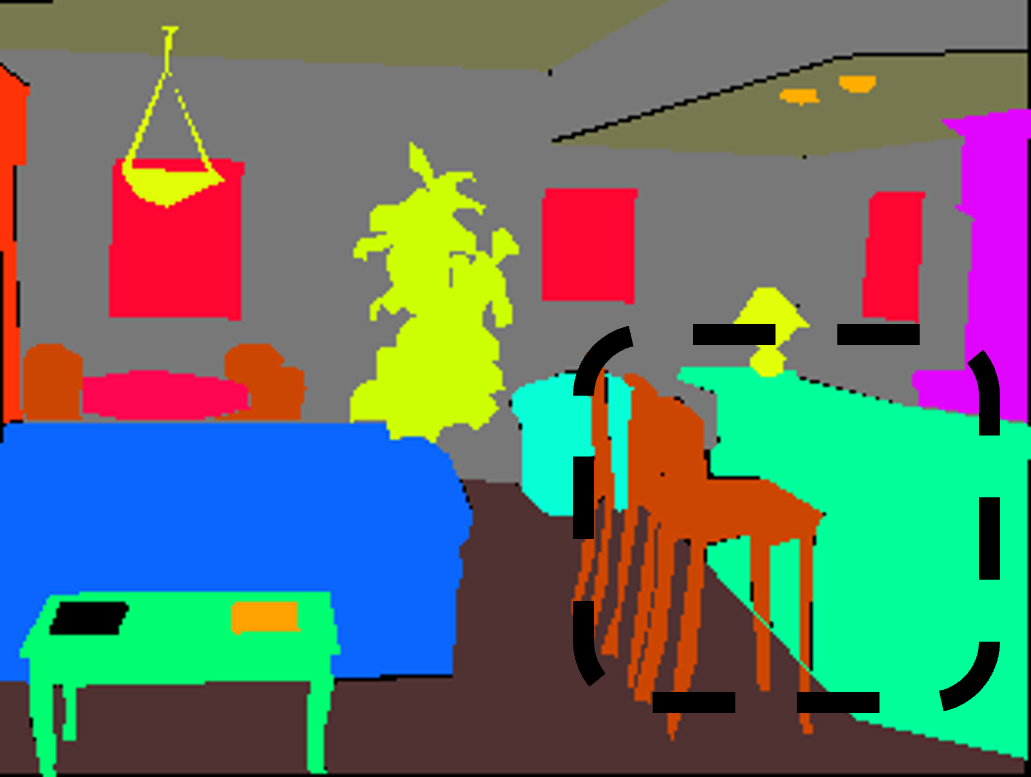}\vspace{1pt}
        \includegraphics[width=4.45cm,height=1.9cm]{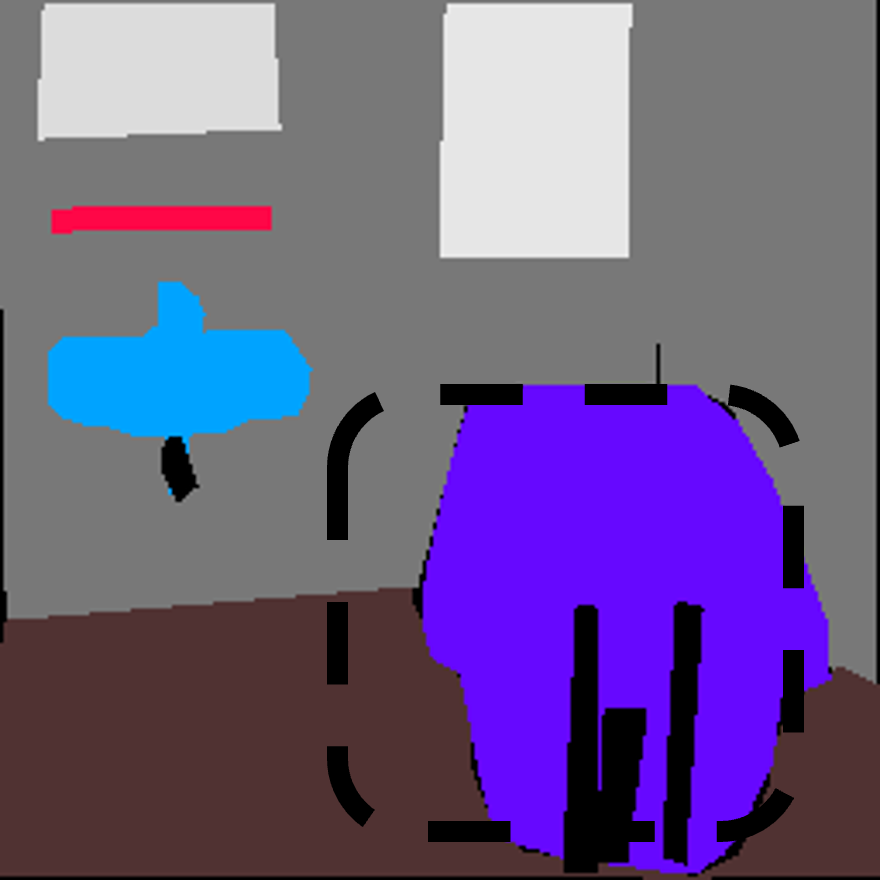}\vspace{1pt}
        \includegraphics[width=4.45cm,height=1.9cm]{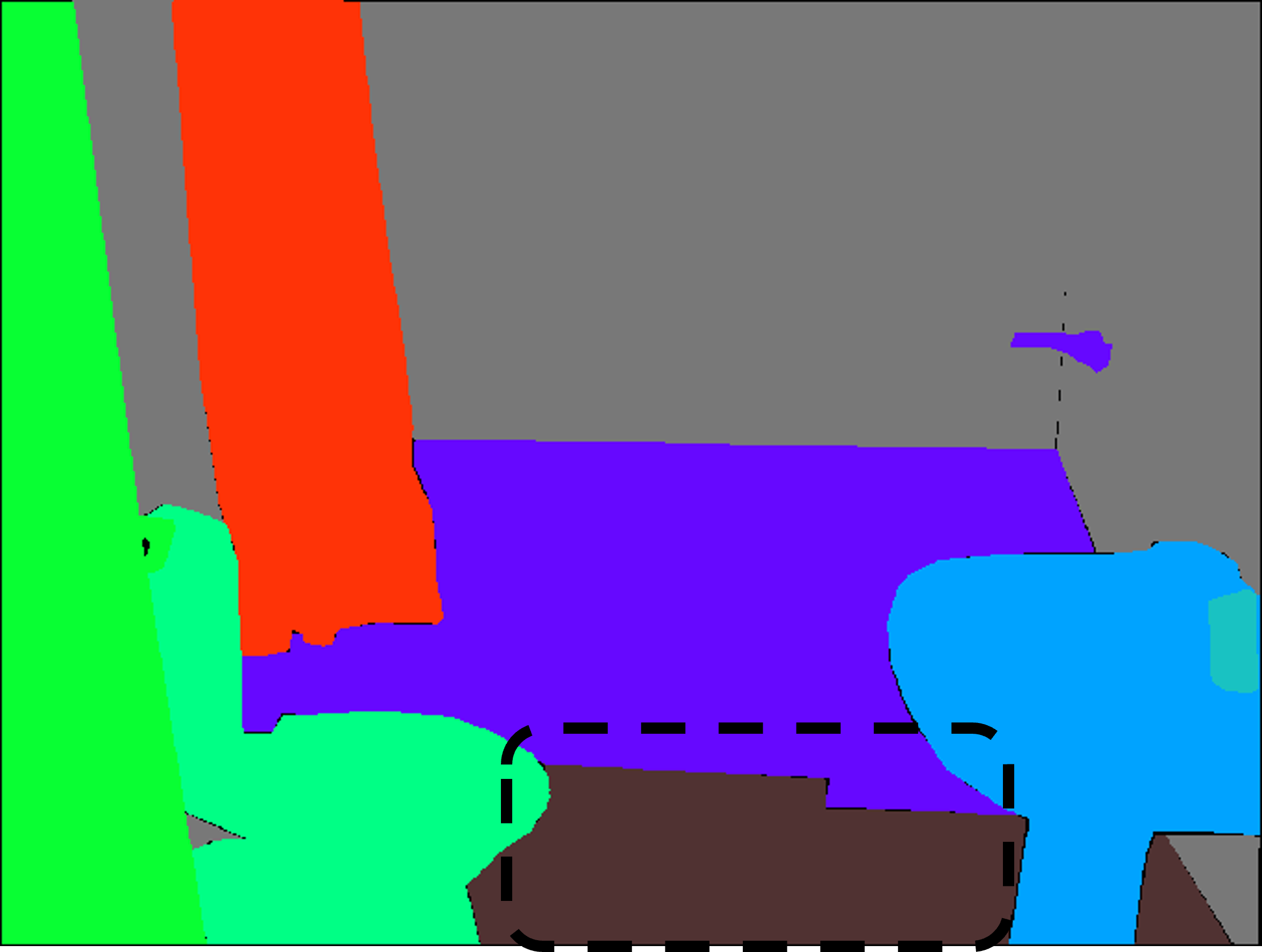}
        \caption*{(d) Ground truth}
    \end{minipage}
    \vspace{-5mm}
             \caption{Qualitative comparisons on ADE20K val~\cite{ade20k}. The dotted boxes emphasize areas with significant improvements achieved through the application of the proposed ECAC. DLV3 is the abbreviation for DeeplabV3plus~\cite{deeplabv3+}.}
    \label{fig:ade}
\end{figure*}
\begin{figure*}[]
    \centering
    \begin{minipage}[b]{0.245\textwidth}
        \centering
        \includegraphics[width=4.45cm,height=1.9cm]{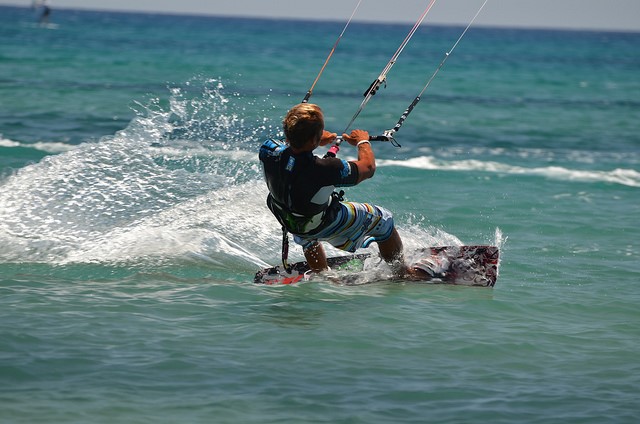}\vspace{1pt}
        \includegraphics[width=4.45cm,height=1.9cm]{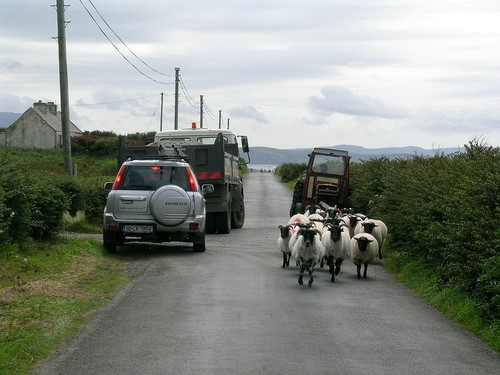}\vspace{1pt}
        \includegraphics[width=4.45cm,height=1.9cm]{vis/img6.jpg}\vspace{1pt}
        \includegraphics[width=4.45cm,height=1.9cm]{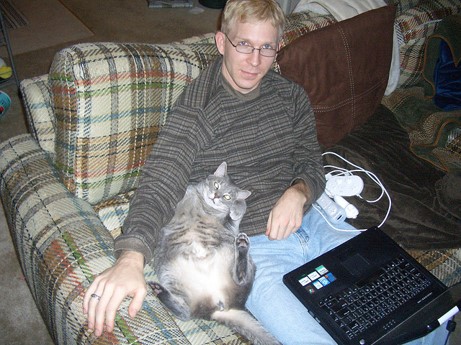}\vspace{1pt}
        \includegraphics[width=4.45cm,height=1.9cm]{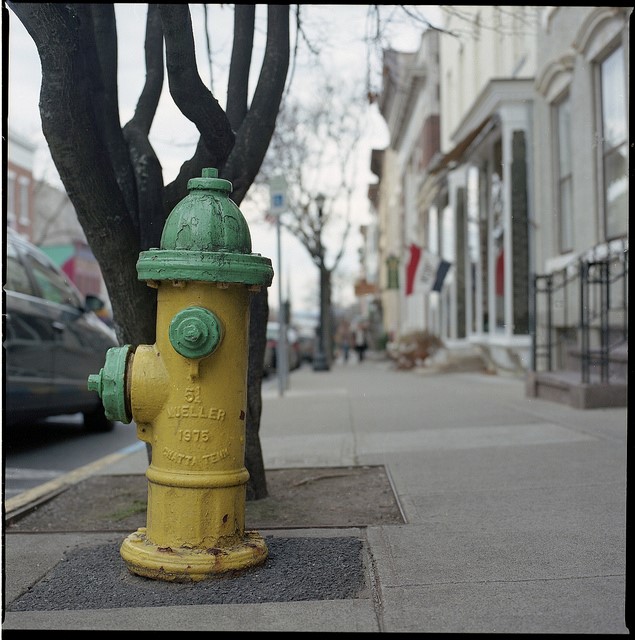}\vspace{1pt}
        \includegraphics[width=4.45cm,height=1.9cm]{vis/img8_.jpg}
        \caption*{(a) Image}
    \end{minipage}
    \begin{minipage}[b]{0.245\textwidth}
        \centering
        \includegraphics[width=4.45cm,height=1.9cm]{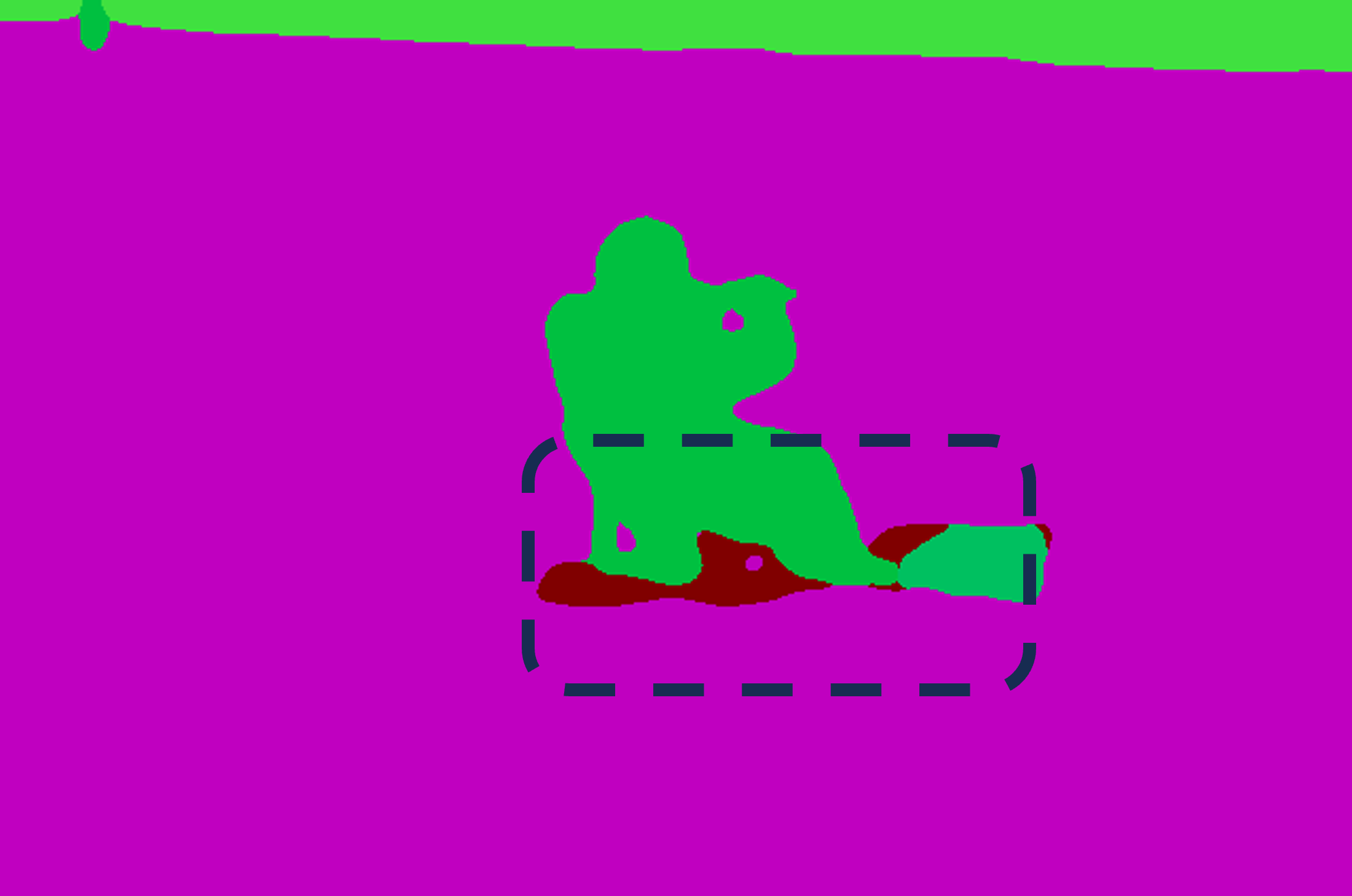}\vspace{1pt}
        \includegraphics[width=4.45cm,height=1.9cm]{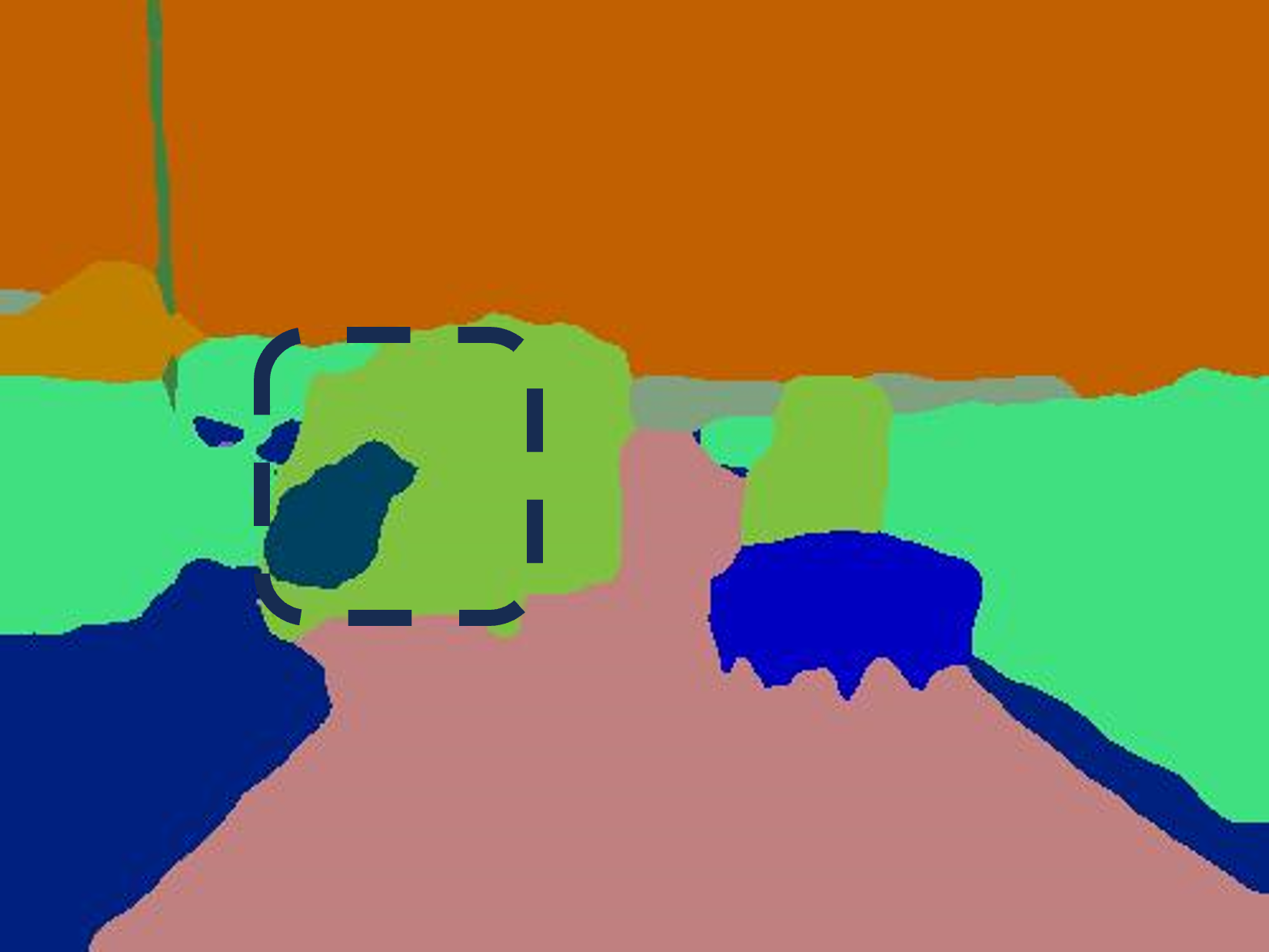}\vspace{1pt}
        \includegraphics[width=4.45cm,height=1.9cm]{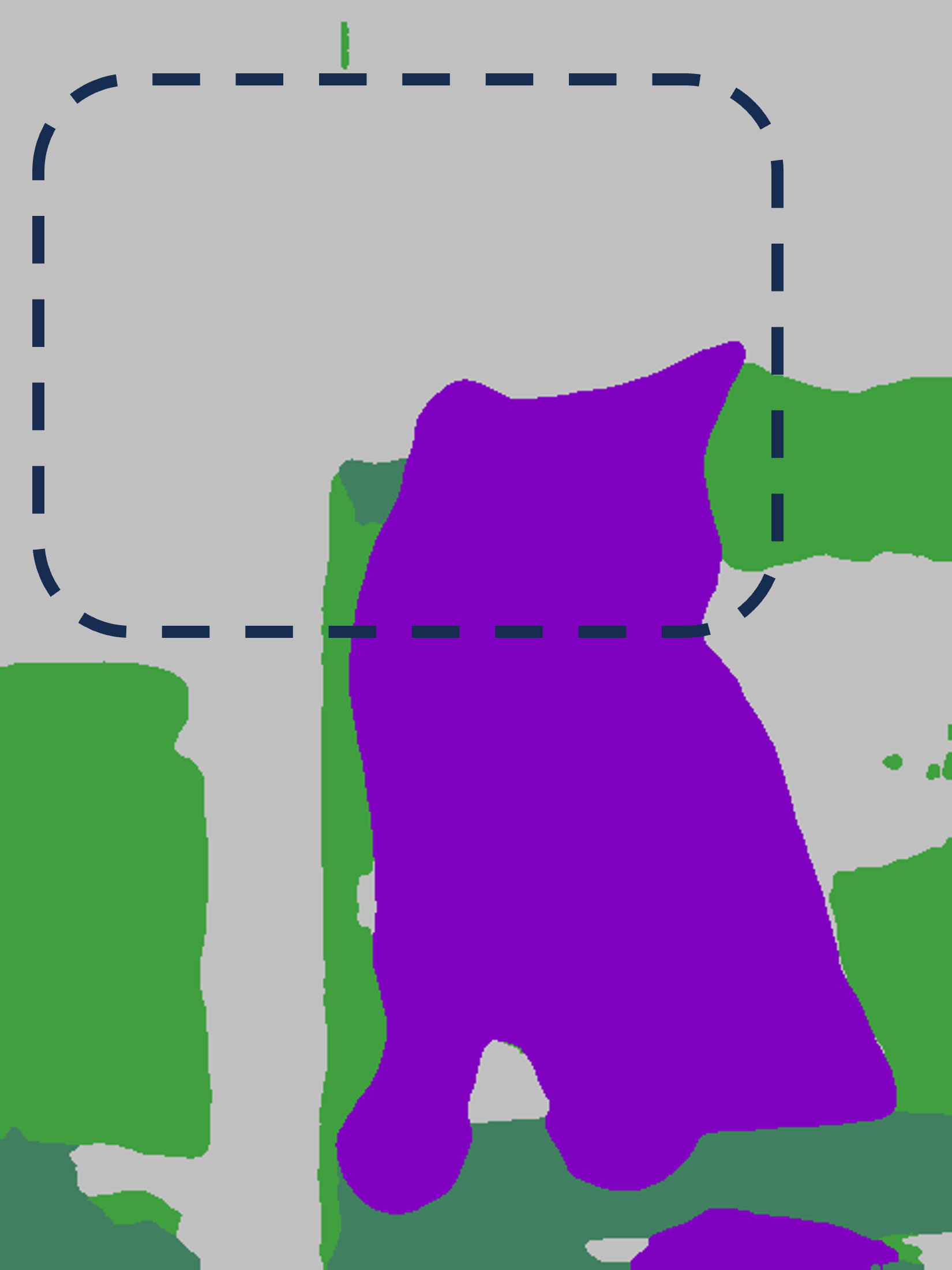}\vspace{1pt}
        \includegraphics[width=4.45cm,height=1.9cm]{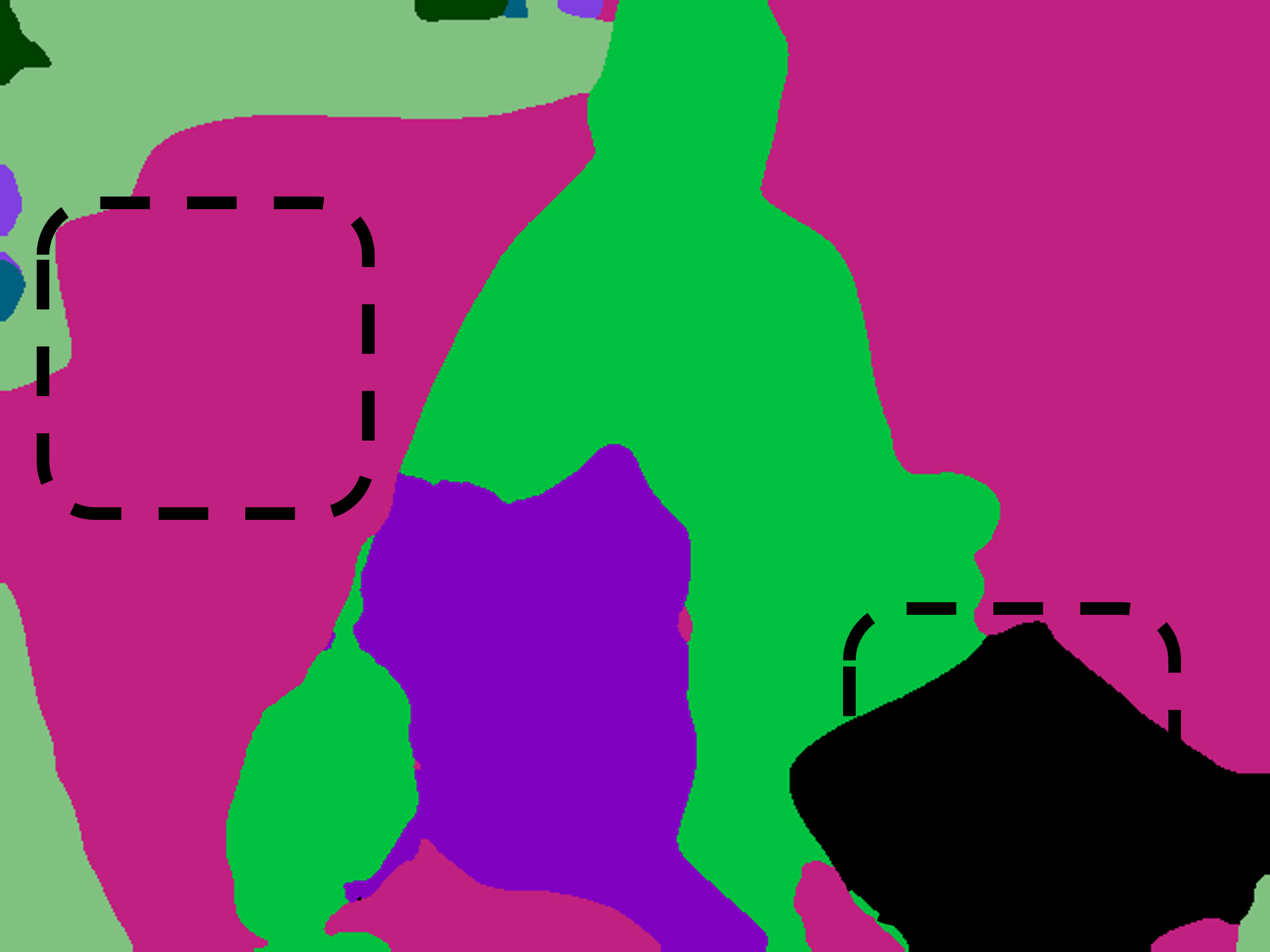}\vspace{1pt}
        \includegraphics[width=4.45cm,height=1.9cm]{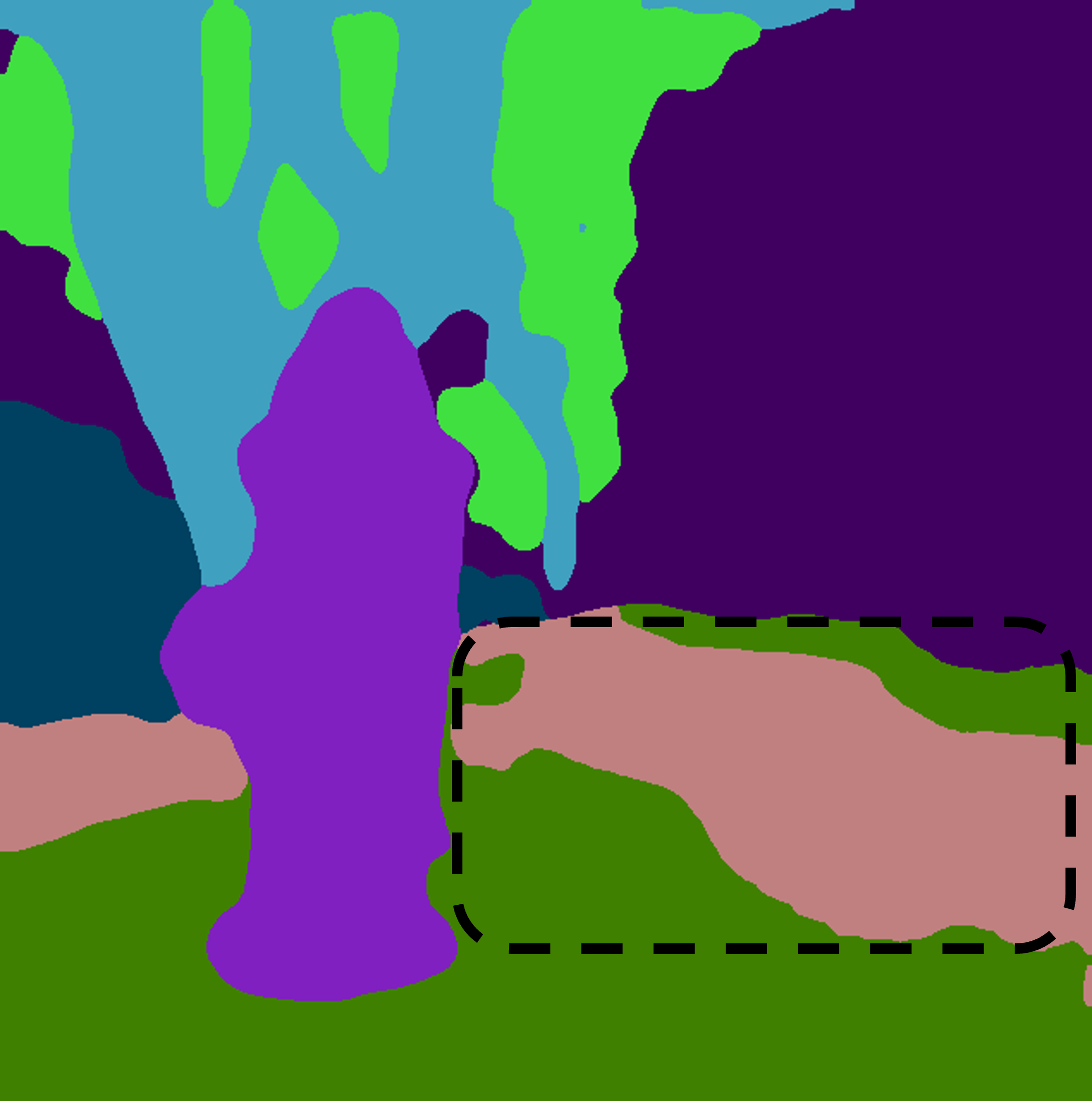}\vspace{1pt}
        \includegraphics[width=4.45cm,height=1.9cm]{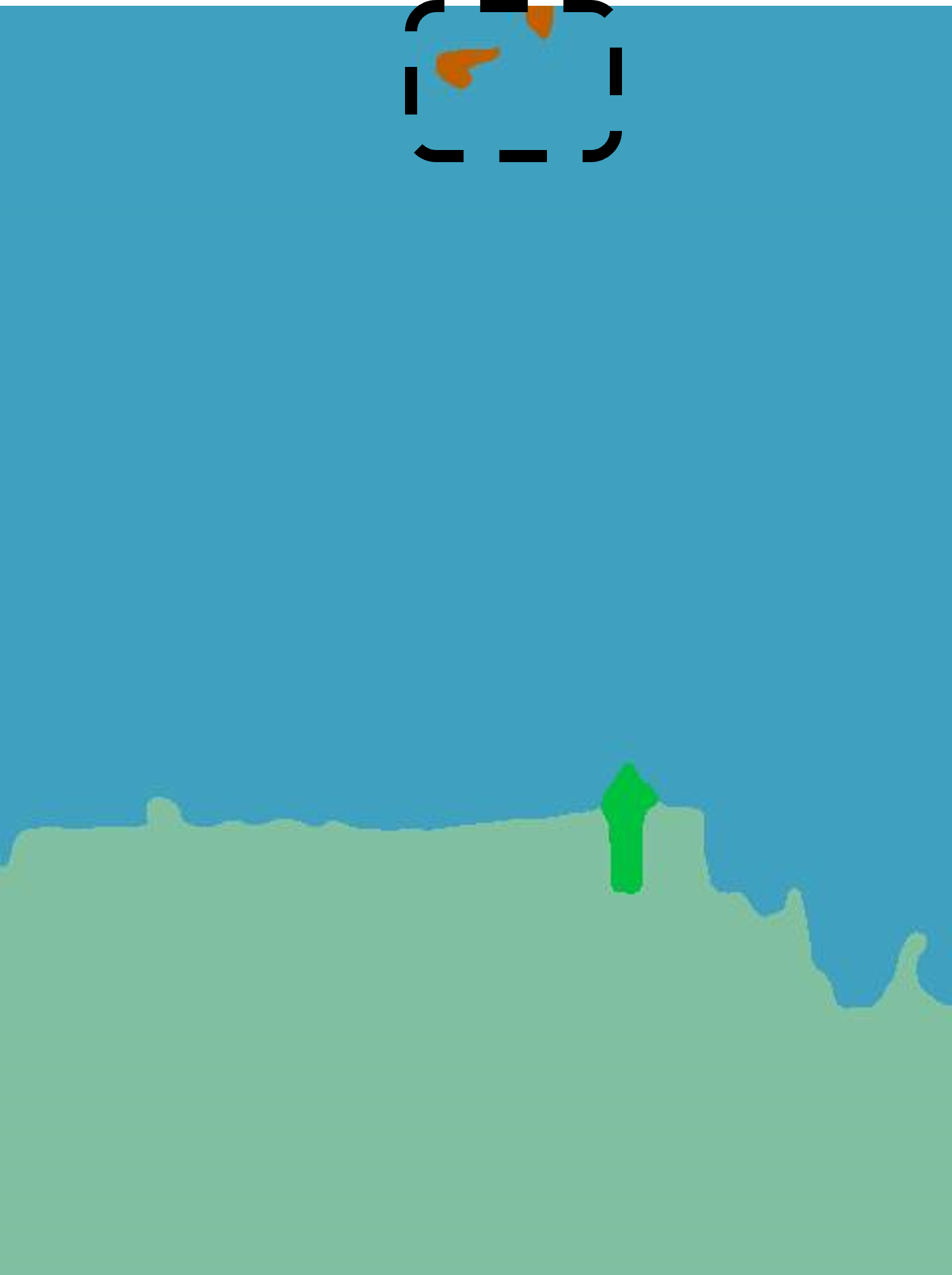}
        \caption*{(b) DLV3plus}
    \end{minipage}
    \begin{minipage}[b]{0.245\textwidth}
        \centering
        \includegraphics[width=4.45cm,height=1.9cm]{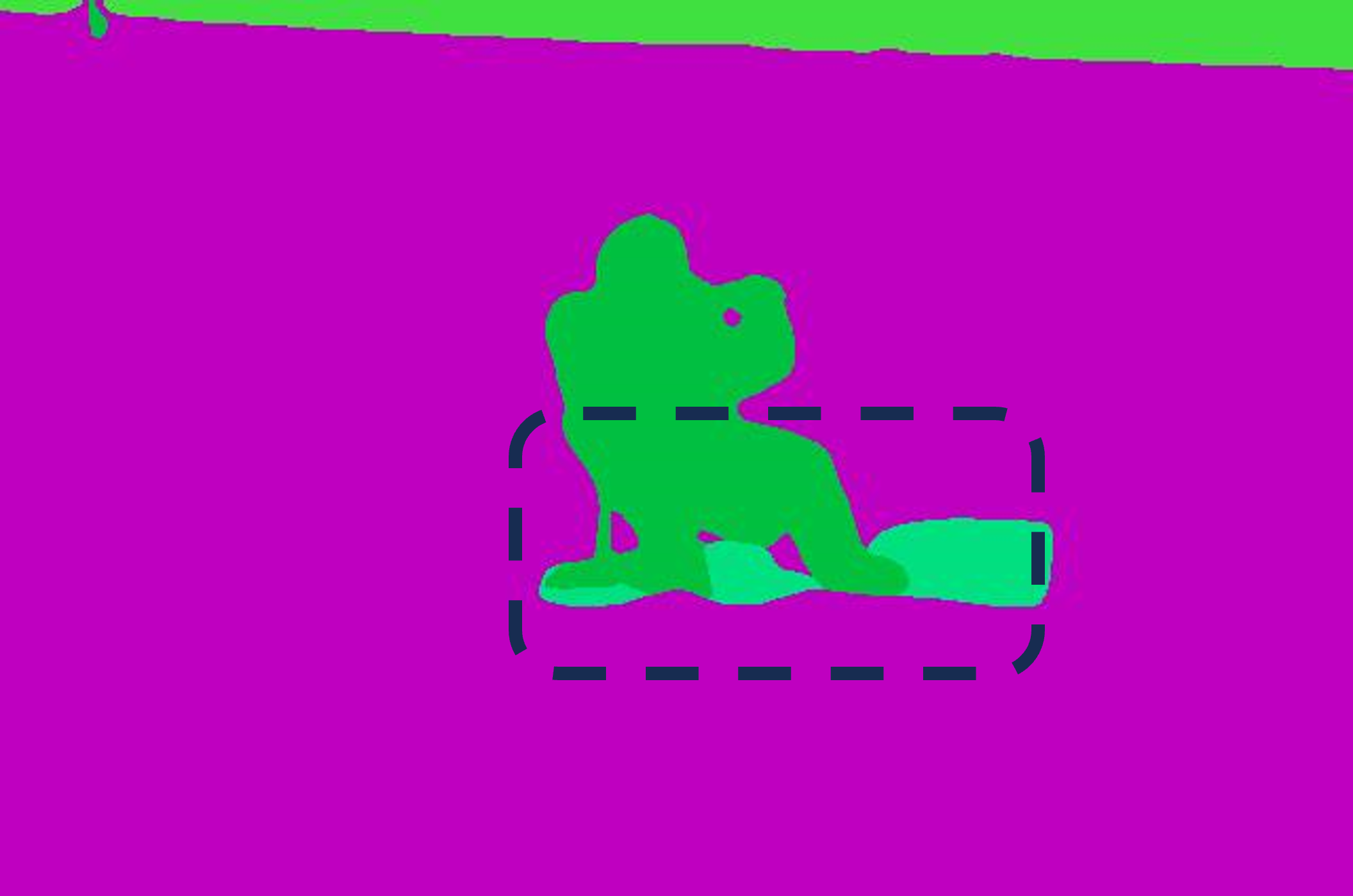}\vspace{1pt}
        \includegraphics[width=4.45cm,height=1.9cm]{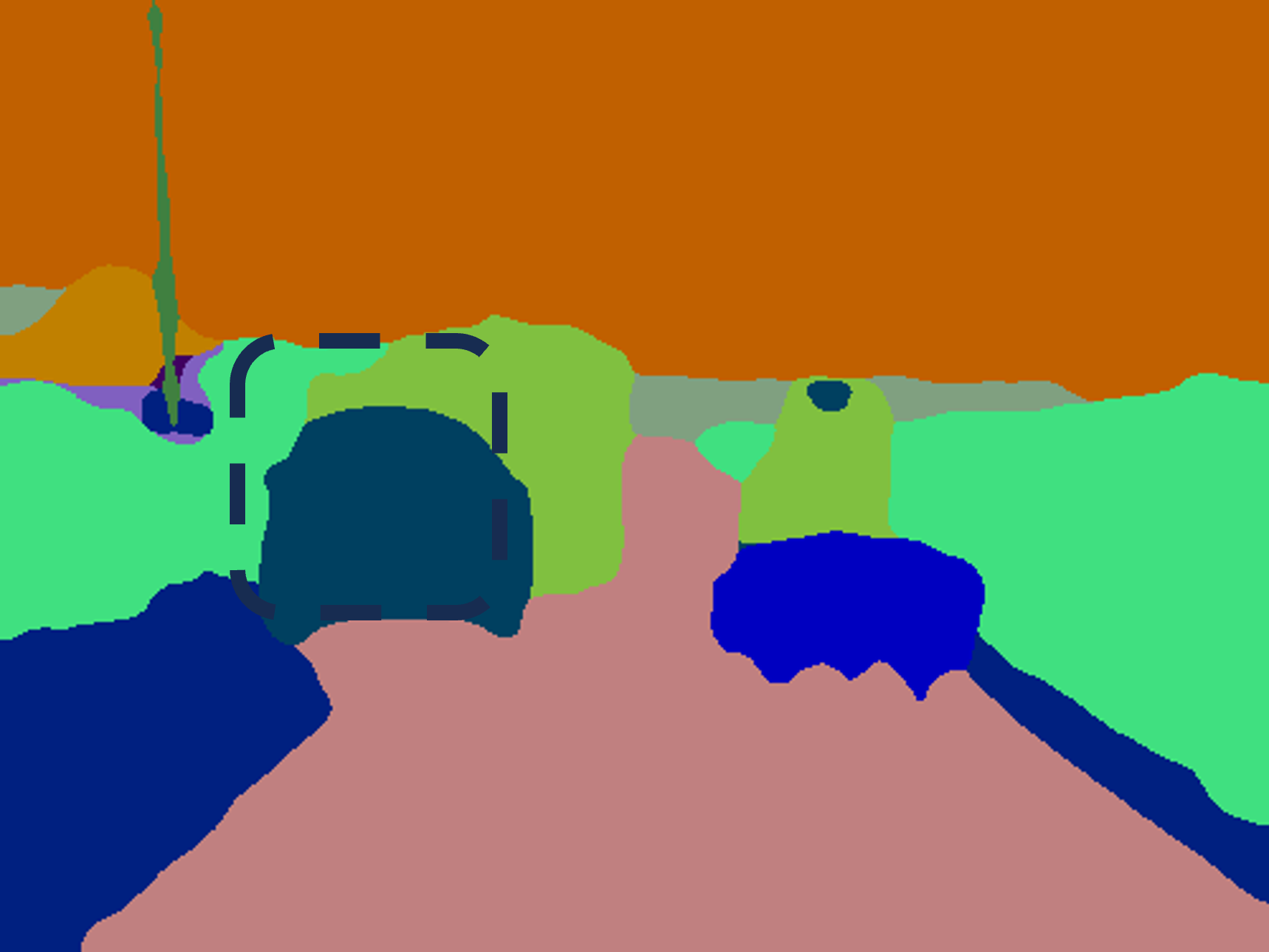}\vspace{1pt}
        \includegraphics[width=4.45cm,height=1.9cm]{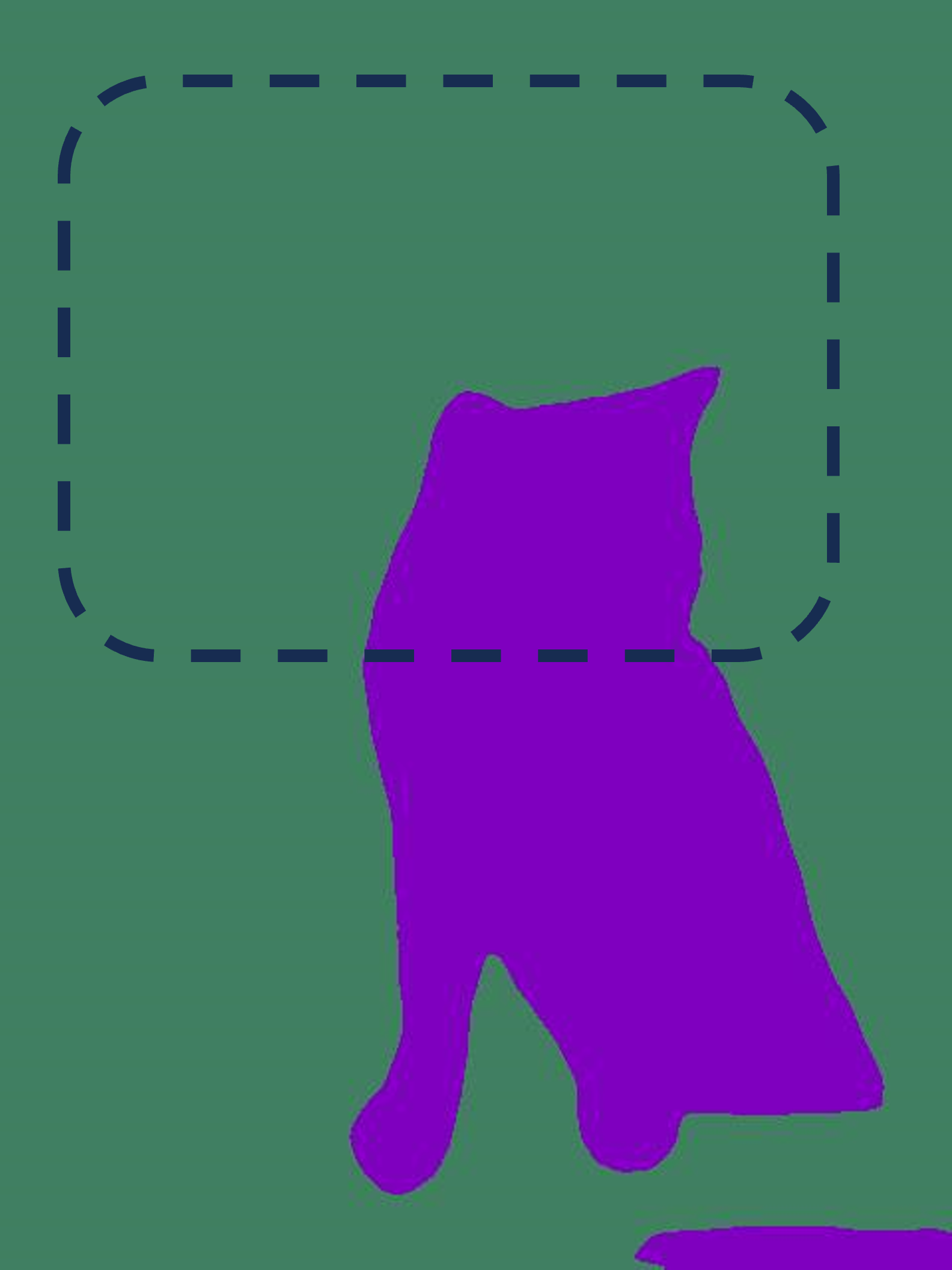}\vspace{1pt}
        \includegraphics[width=4.45cm,height=1.9cm]{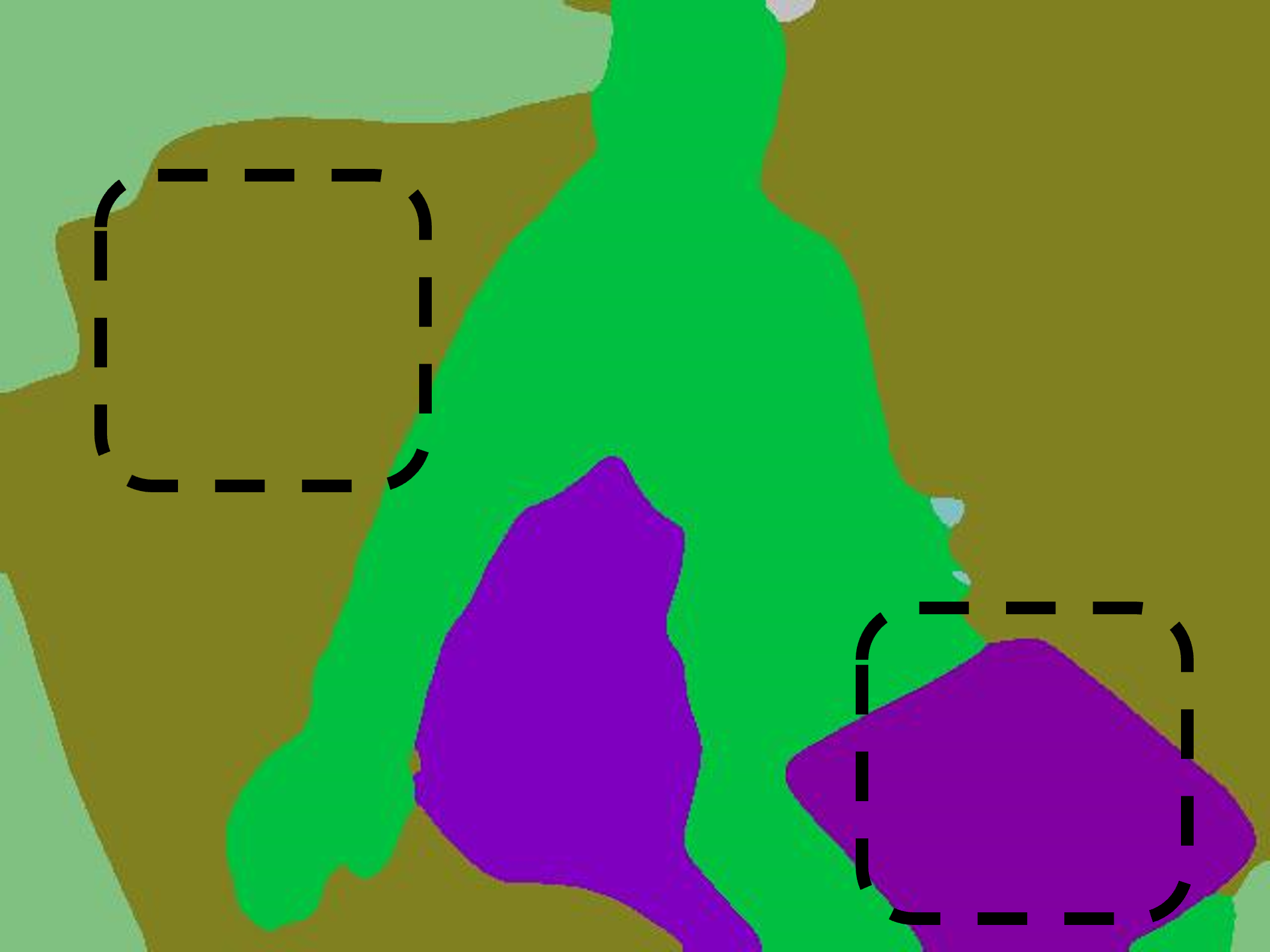}\vspace{1pt}
        \includegraphics[width=4.45cm,height=1.9cm]{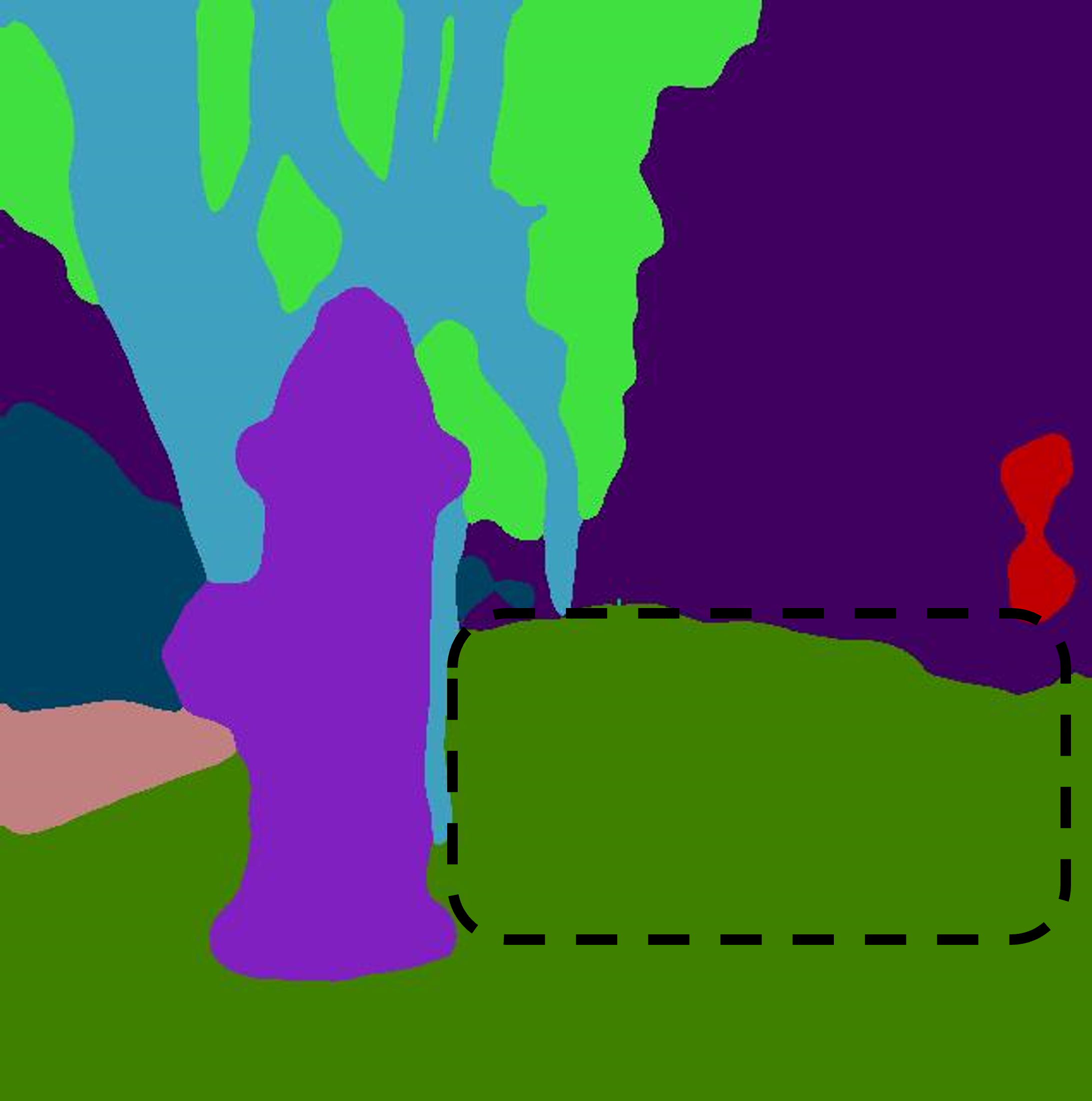}\vspace{1pt}
        \includegraphics[width=4.45cm,height=1.9cm]{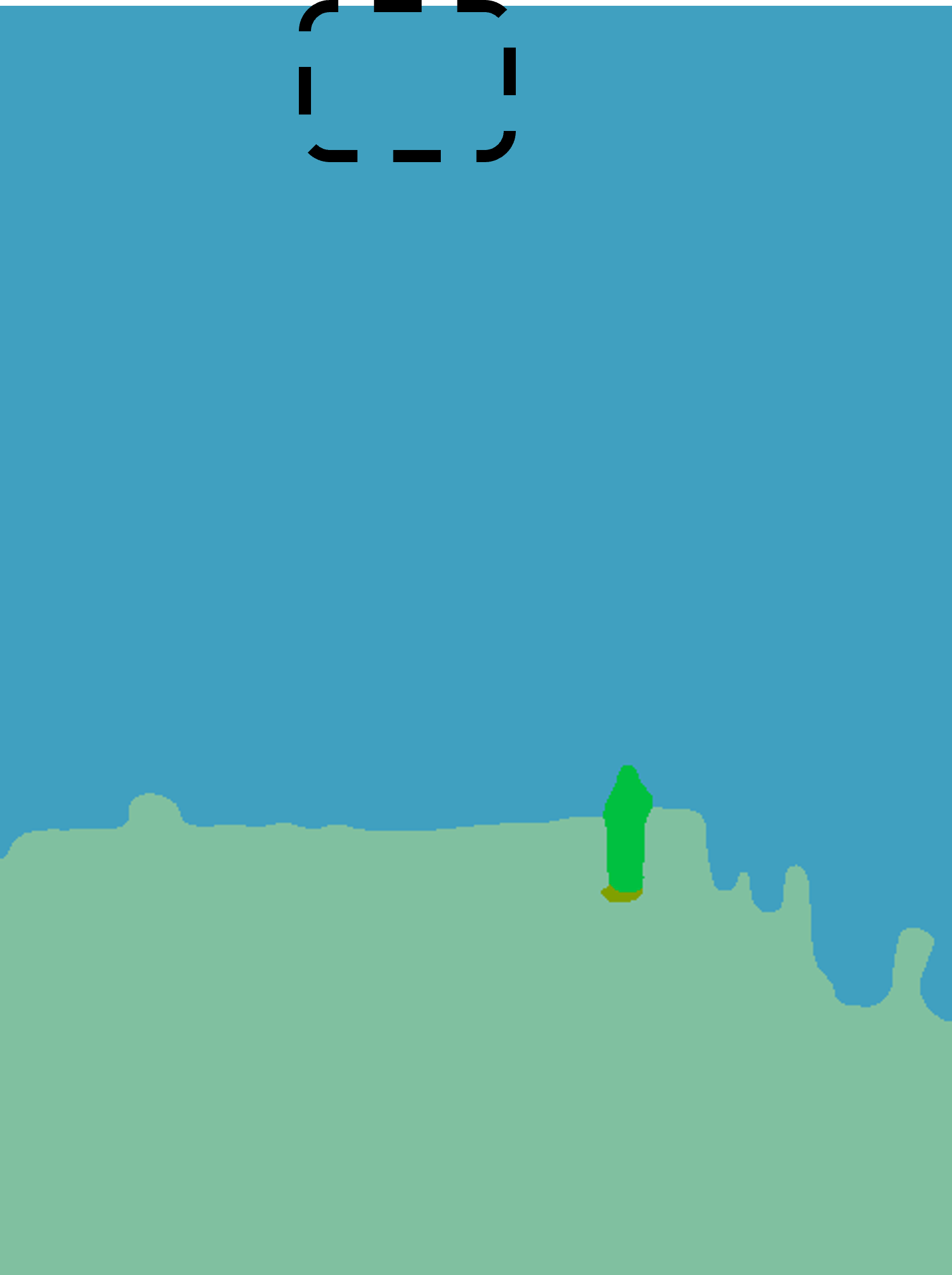}
        \caption*{(c) DLV3plus+ECAC(ours)}
    \end{minipage}
    \begin{minipage}[b]{0.245\textwidth}
        \centering
        \includegraphics[width=4.45cm,height=1.9cm]{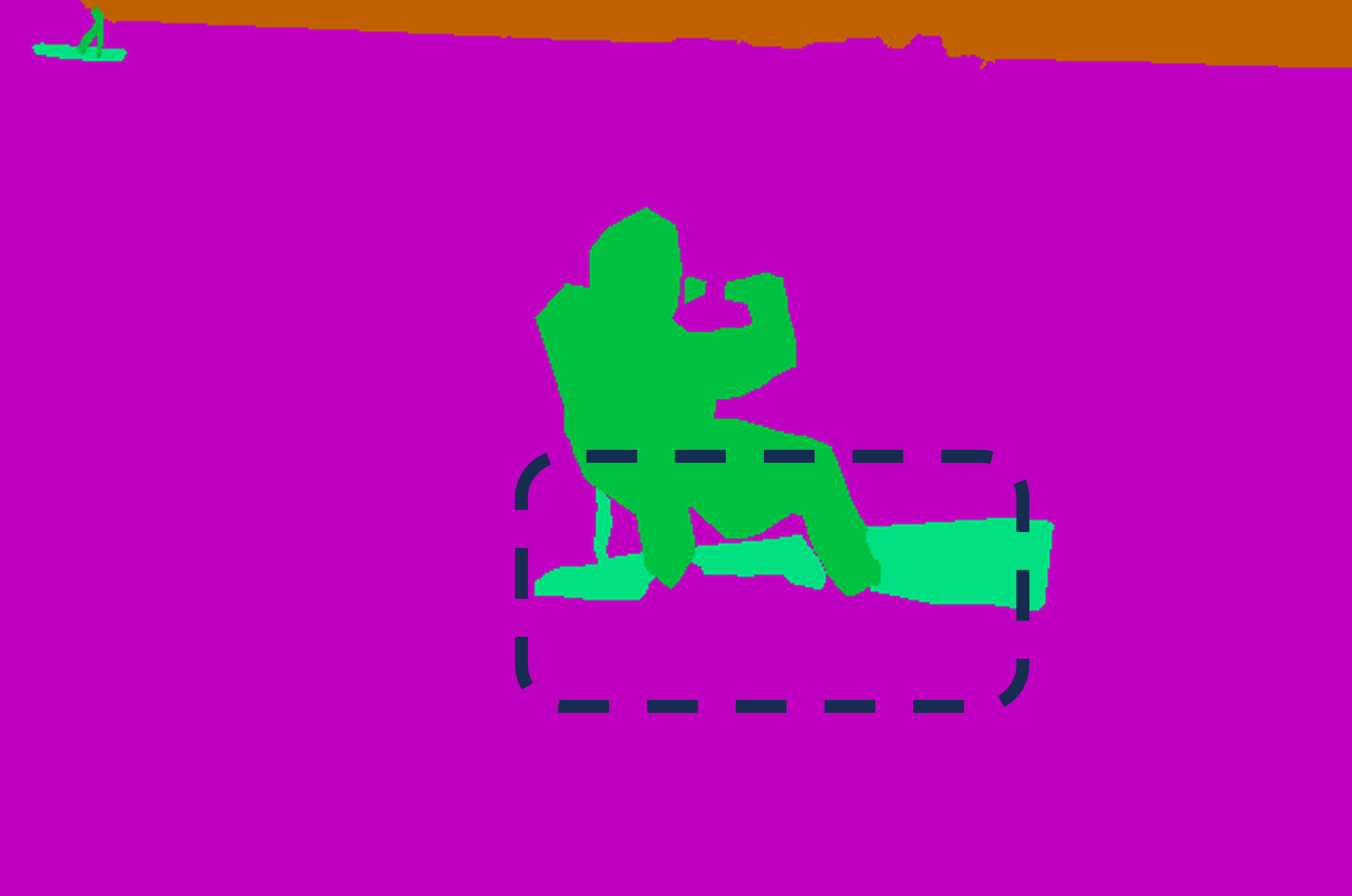}\vspace{1pt}
        \includegraphics[width=4.45cm,height=1.9cm]{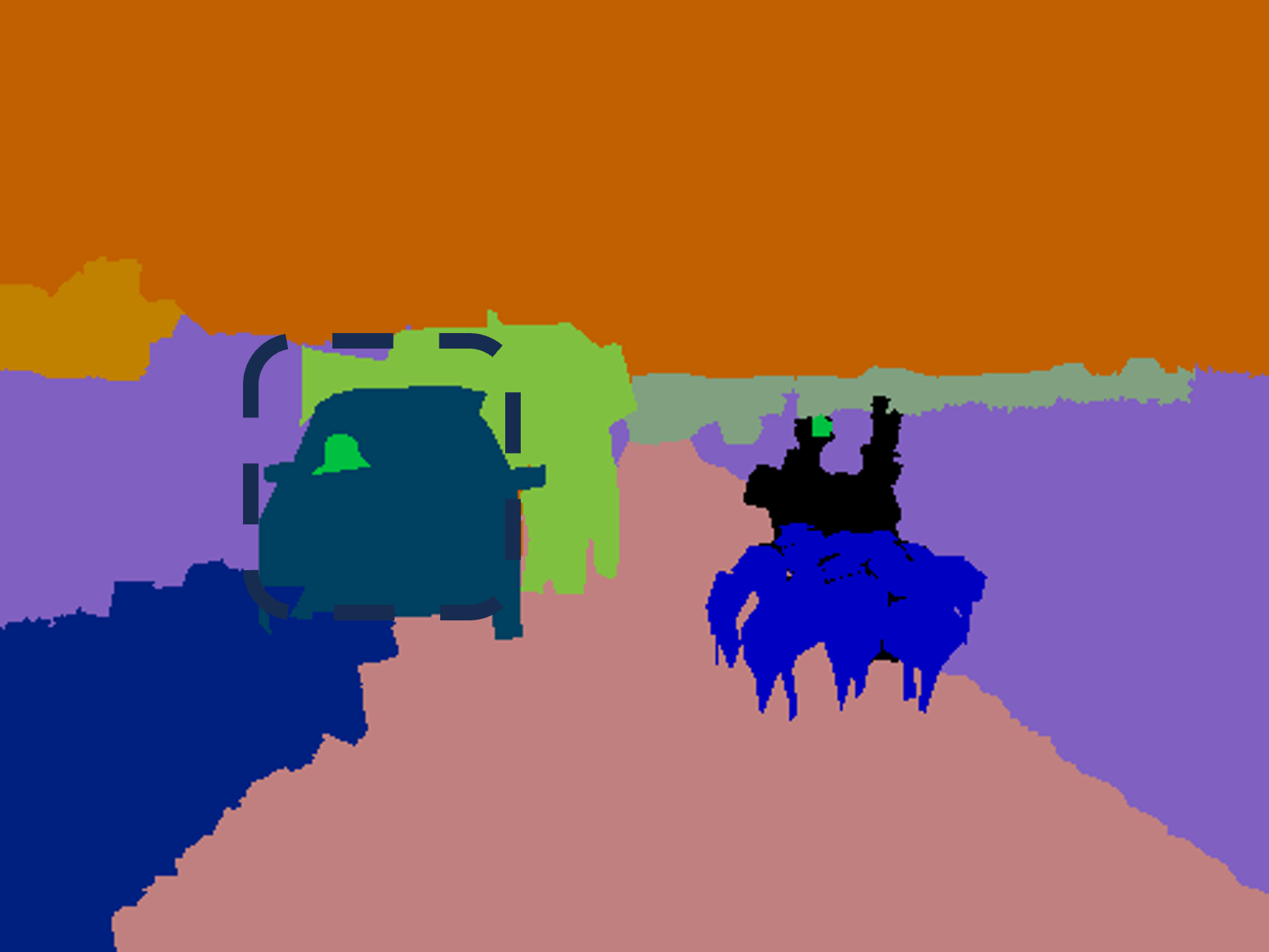}\vspace{1pt}
        \includegraphics[width=4.45cm,height=1.9cm]{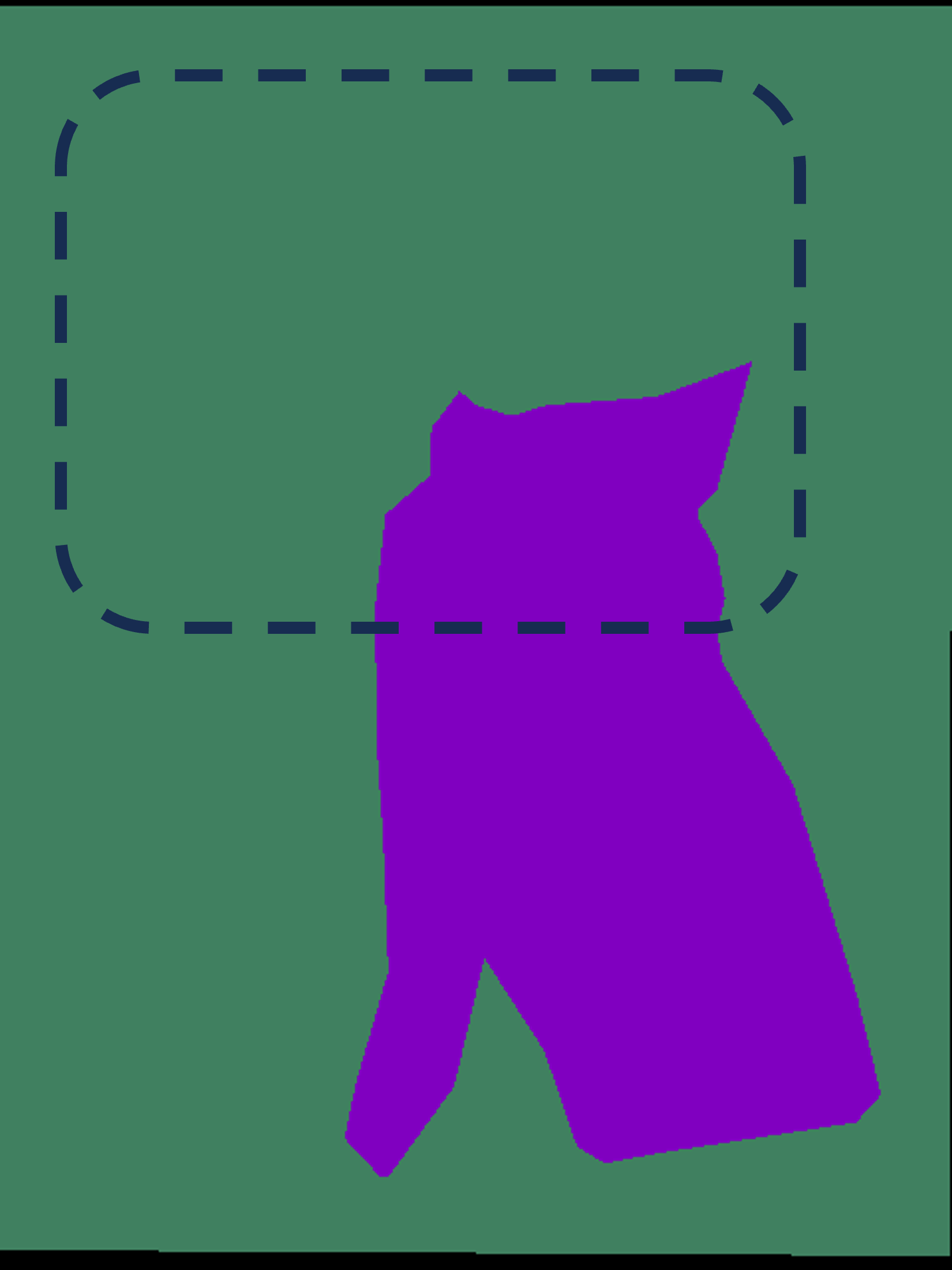}\vspace{1pt}
        \includegraphics[width=4.45cm,height=1.9cm]{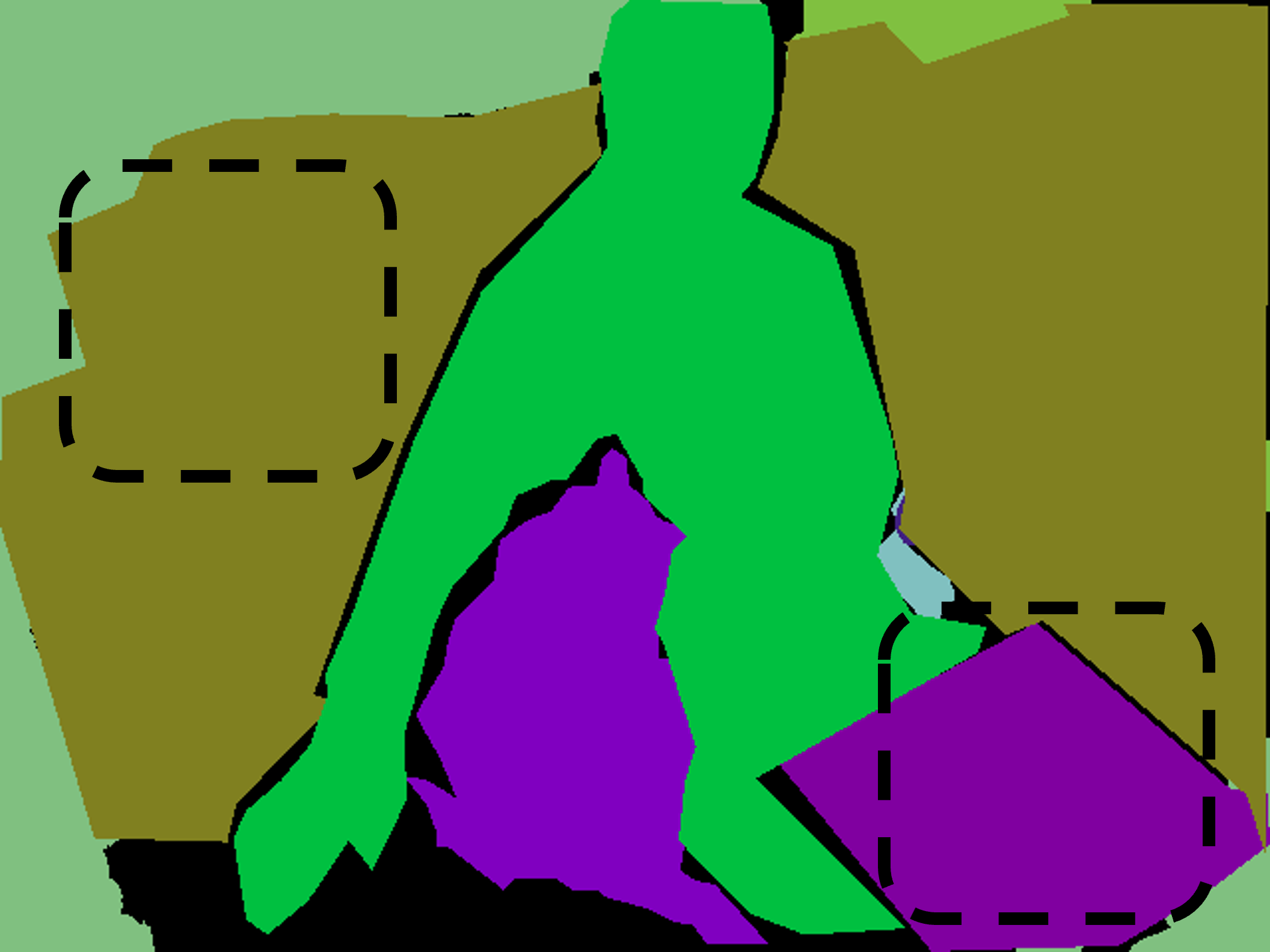}\vspace{1pt}
        \includegraphics[width=4.45cm,height=1.9cm]{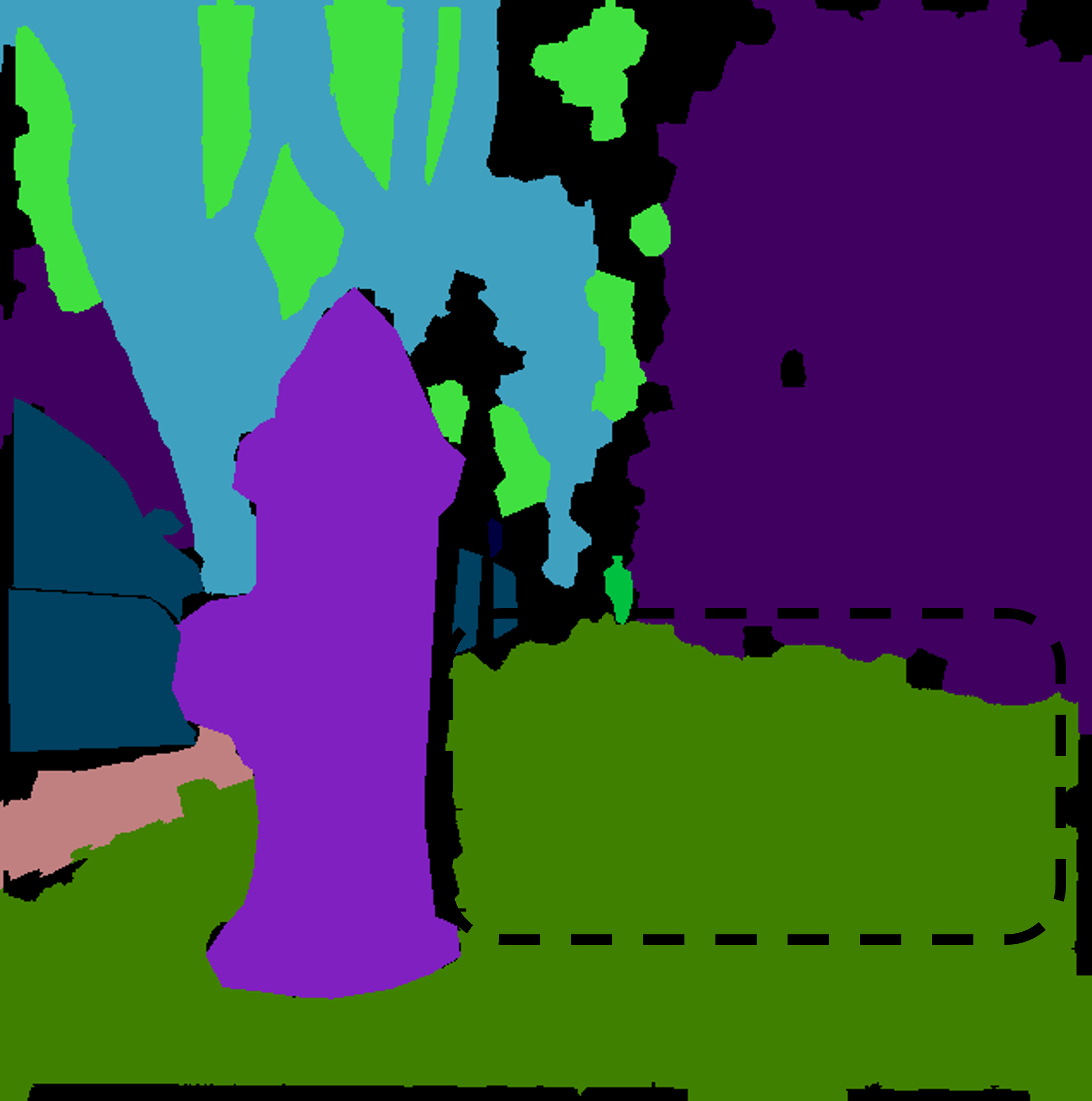}\vspace{1pt}
        \includegraphics[width=4.45cm,height=1.9cm]{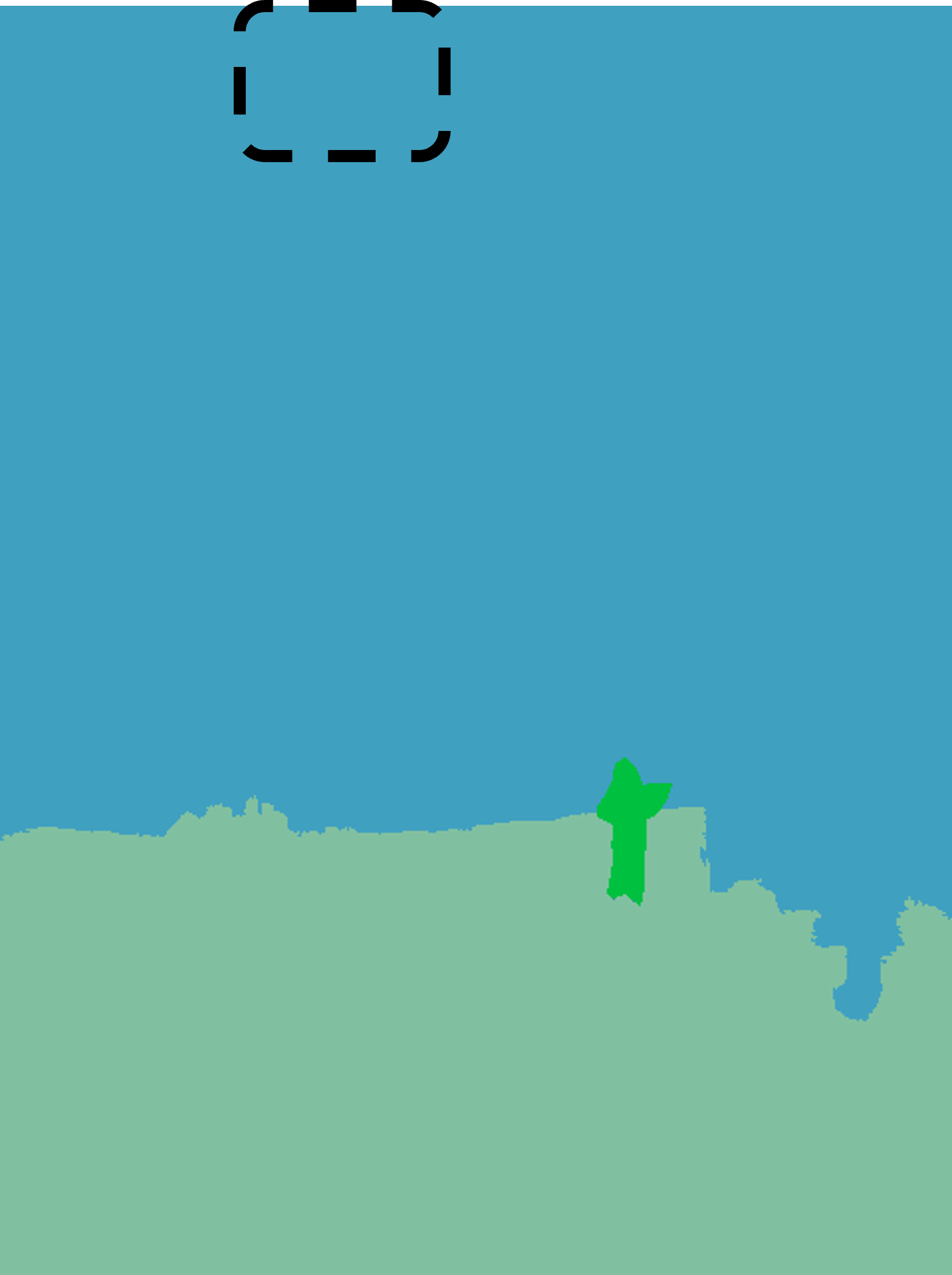}
        \caption*{(d) Ground truth}
    \end{minipage}
    \vspace{-5mm}
    \caption{Qualitative comparisons on COCO-Stuff10K test~\cite{coco}. The dotted boxes emphasize areas with significant improvements achieved through the application of the proposed ECAC. DLV3 is the abbreviation for DeeplabV3plus~\cite{deeplabv3+}.}
    \label{fig:coco}
\end{figure*}
\begin{figure*}[]
    \centering
    \begin{minipage}[b]{0.245\textwidth}
        \centering
        \includegraphics[width=4.45cm,height=1.9cm]{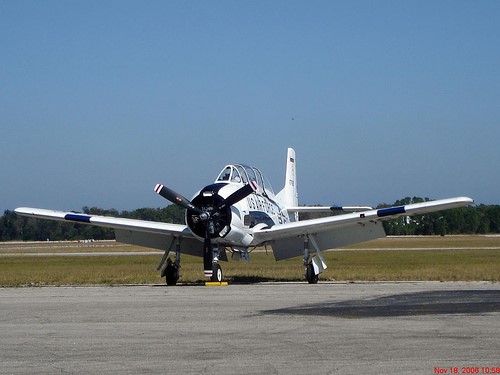}\vspace{1pt}
        \includegraphics[width=4.45cm,height=1.9cm]{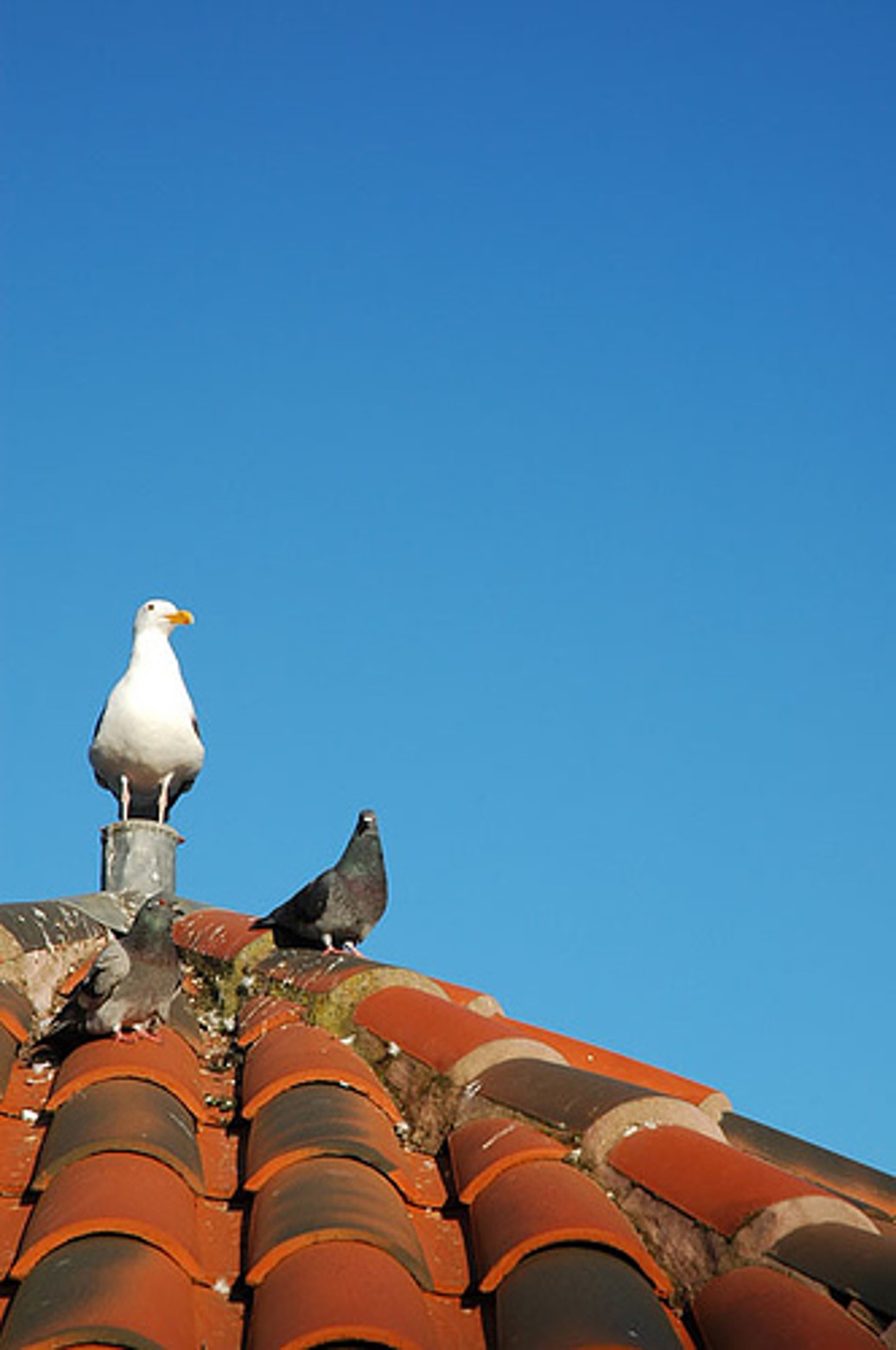}\vspace{1pt}
        \includegraphics[width=4.45cm,height=1.9cm]{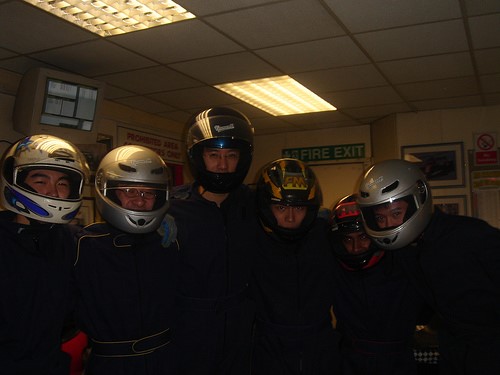}\vspace{1pt}
        \includegraphics[width=4.45cm,height=1.9cm]{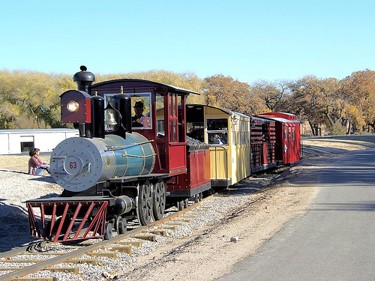}\vspace{1pt}
        \includegraphics[width=4.45cm,height=1.9cm]{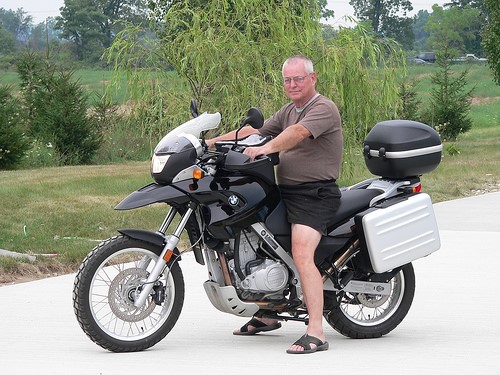}\vspace{1pt}
        \includegraphics[width=4.45cm,height=1.9cm]{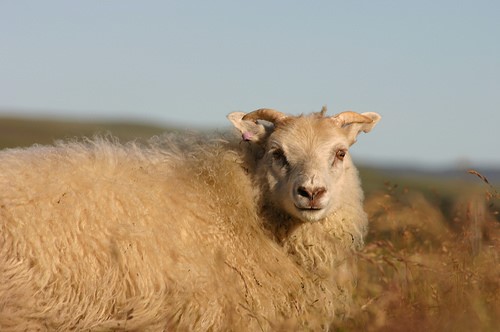}
        \caption*{(a) Image}
    \end{minipage}
    \begin{minipage}[b]{0.245\textwidth}
        \centering
        \includegraphics[width=4.45cm,height=1.9cm]{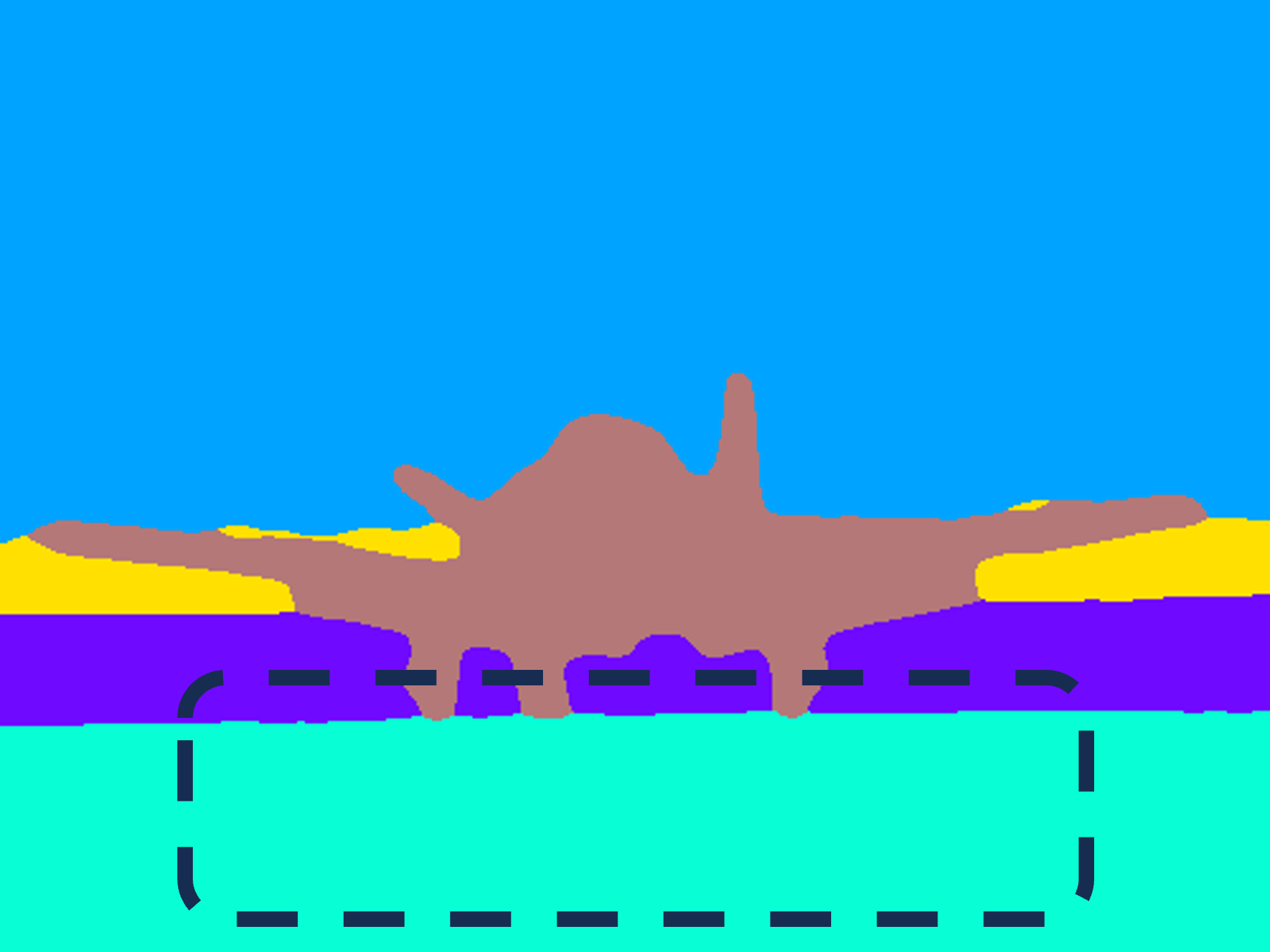}\vspace{1pt}
        \includegraphics[width=4.45cm,height=1.9cm]{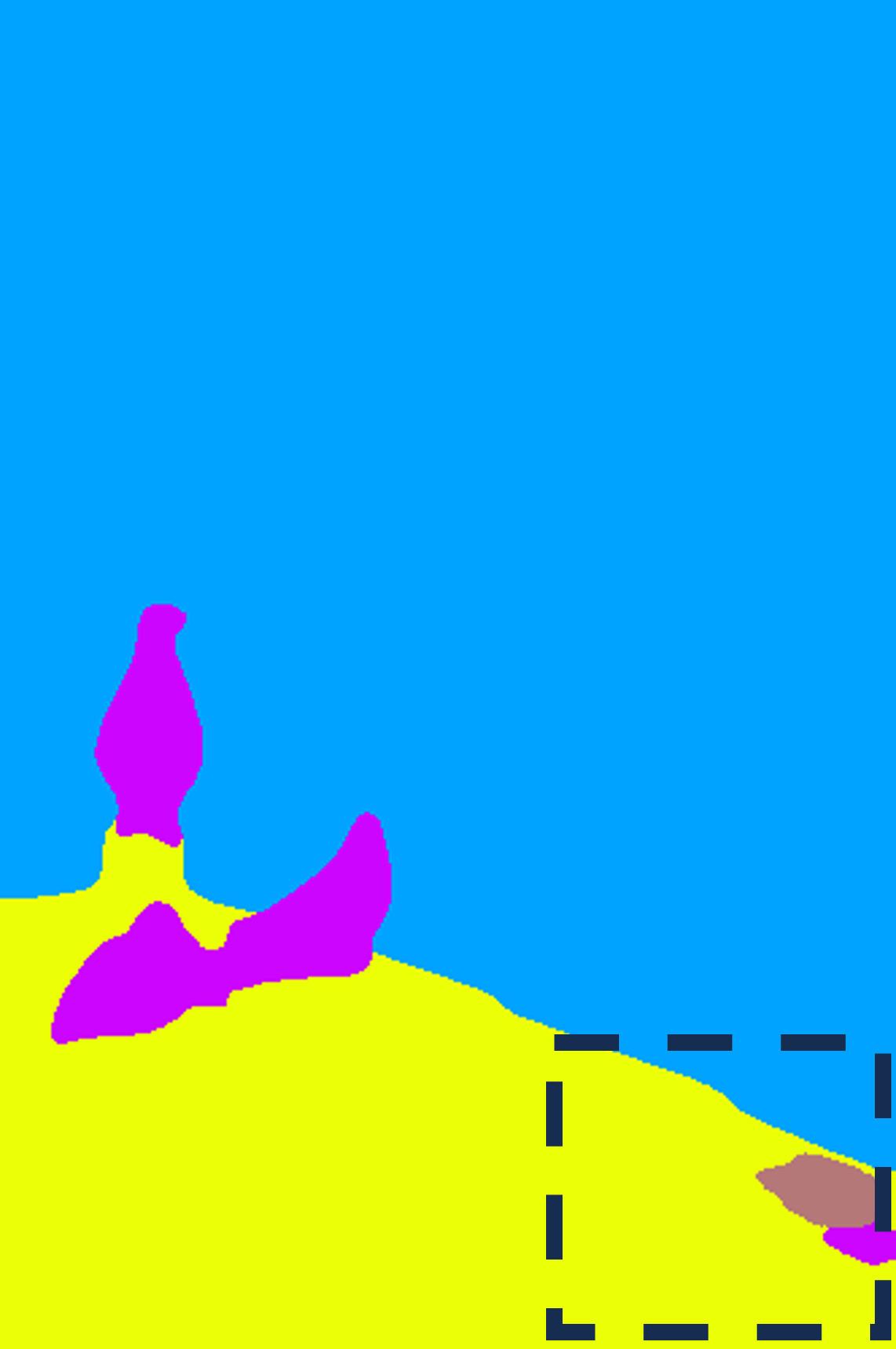}\vspace{1pt}
        \includegraphics[width=4.45cm,height=1.9cm]{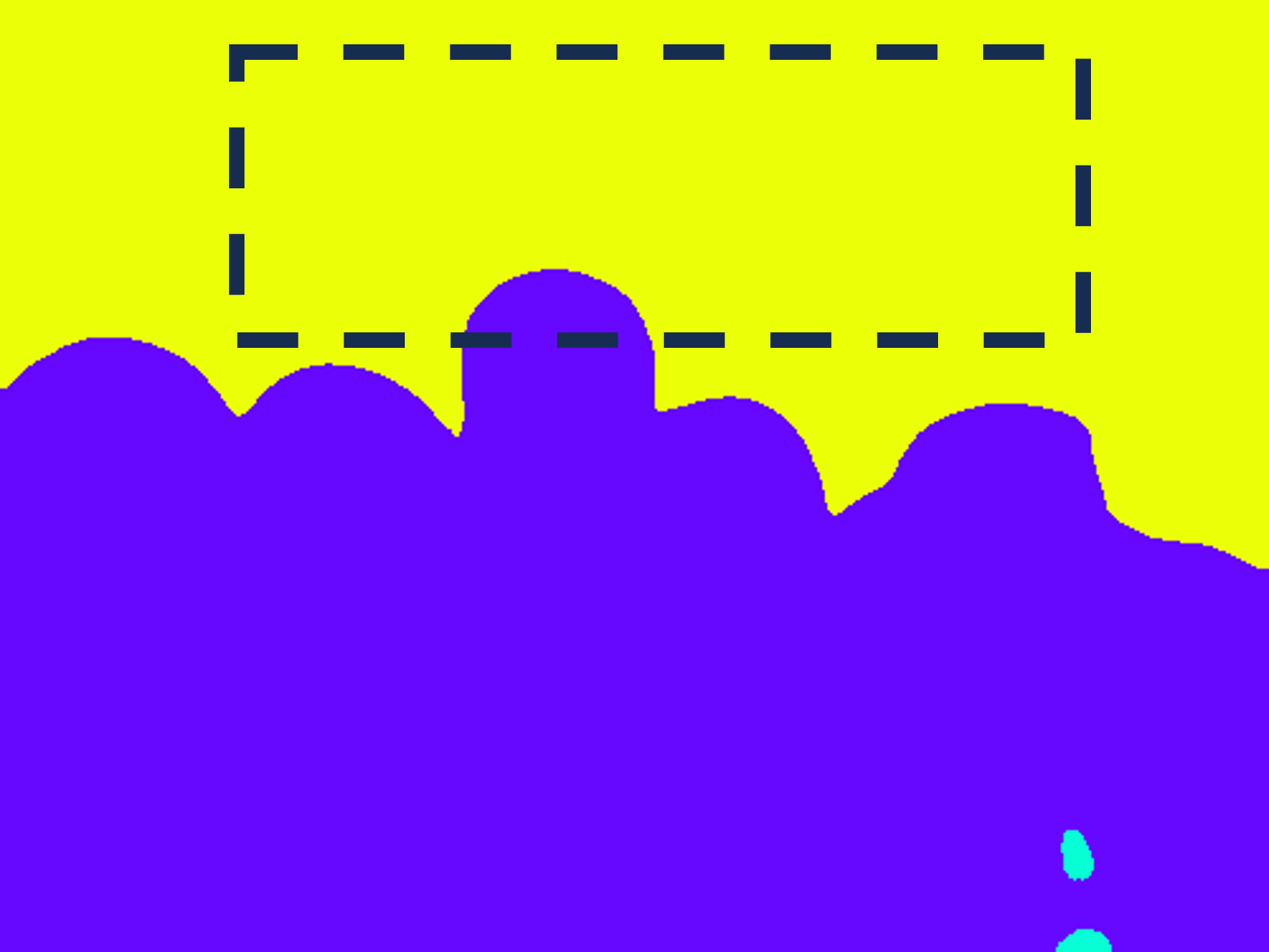}\vspace{1pt}
        \includegraphics[width=4.45cm,height=1.9cm]{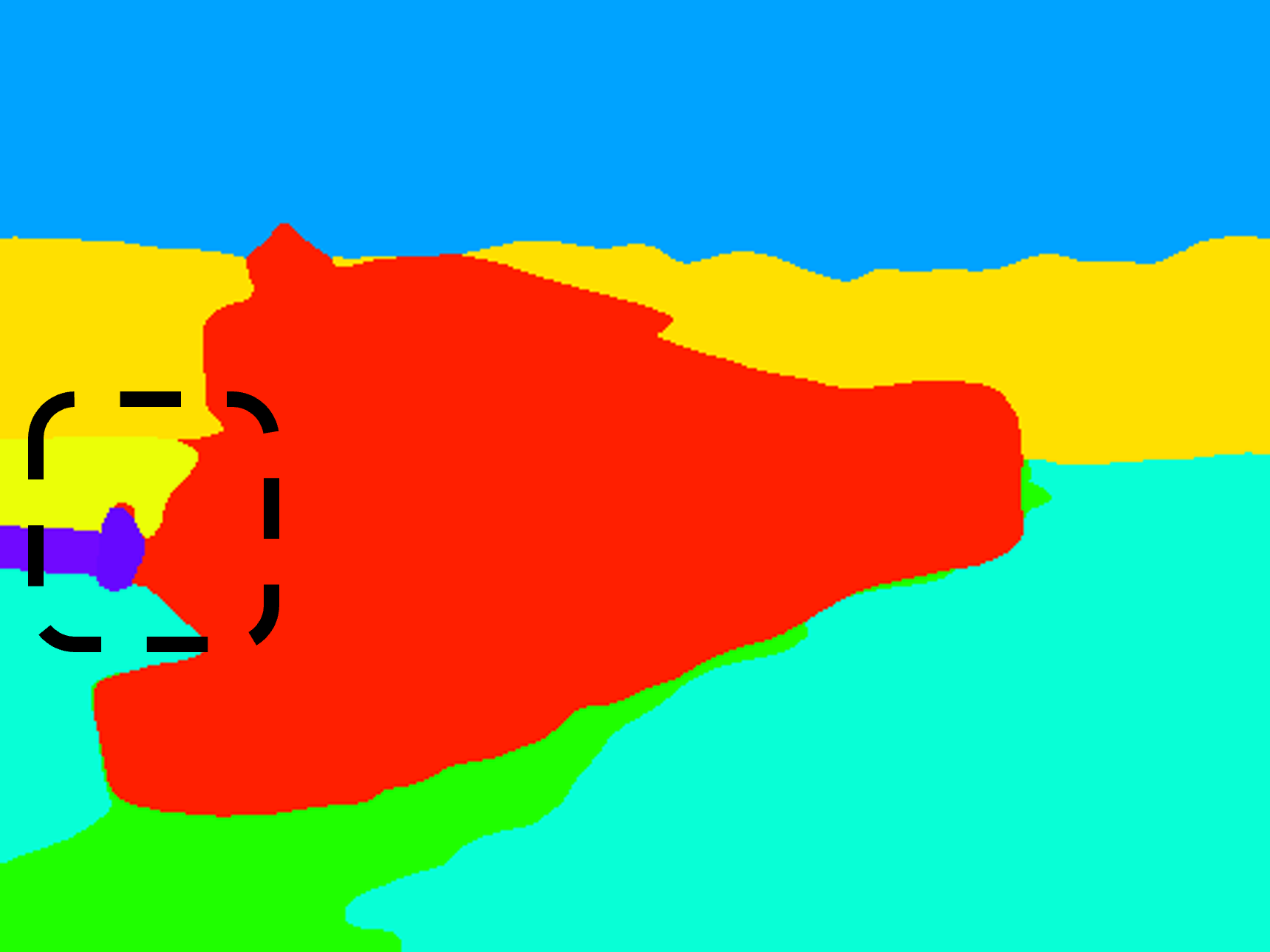}\vspace{1pt}
        \includegraphics[width=4.45cm,height=1.9cm]{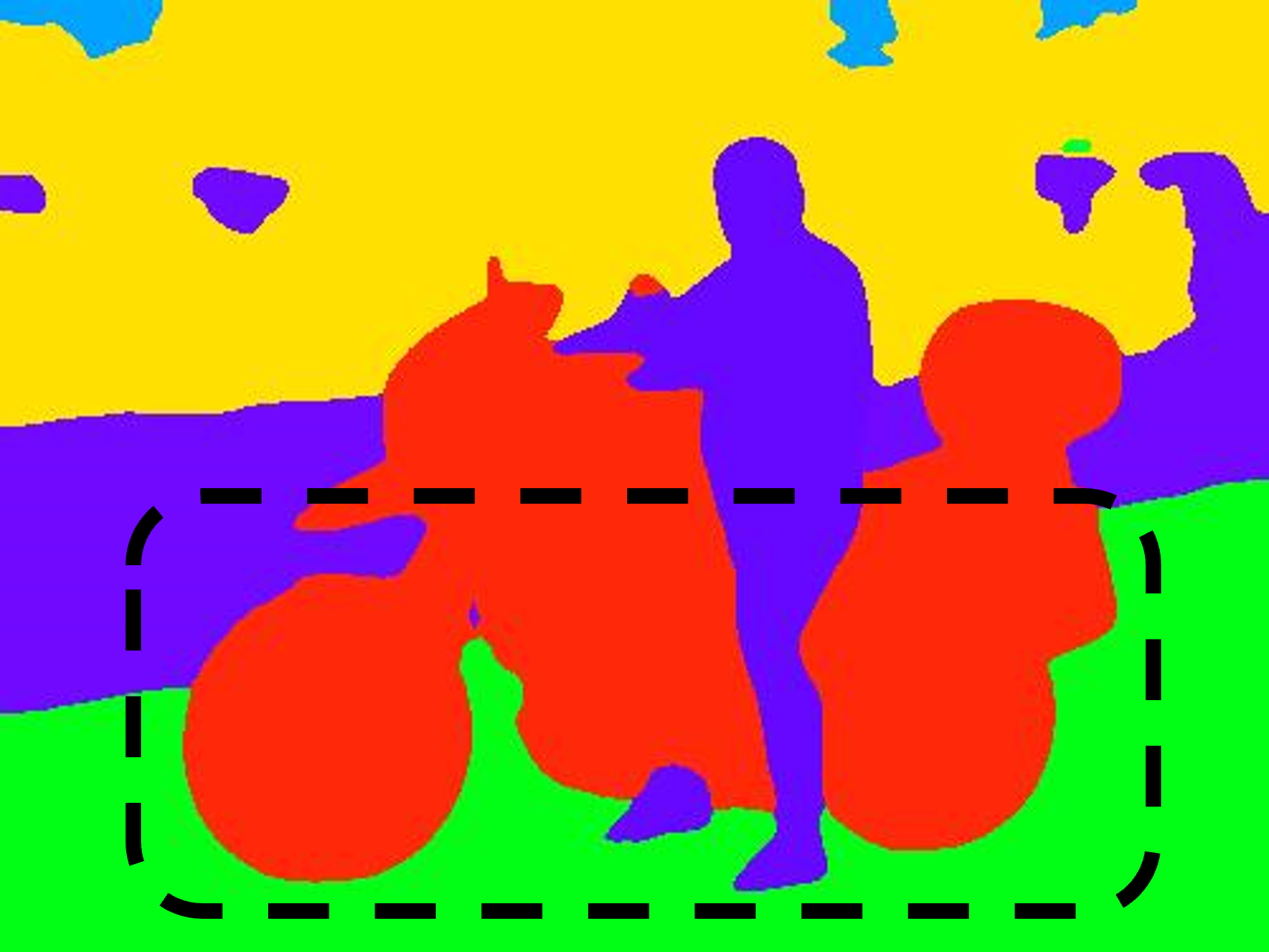}\vspace{1pt}
        \includegraphics[width=4.45cm,height=1.9cm]{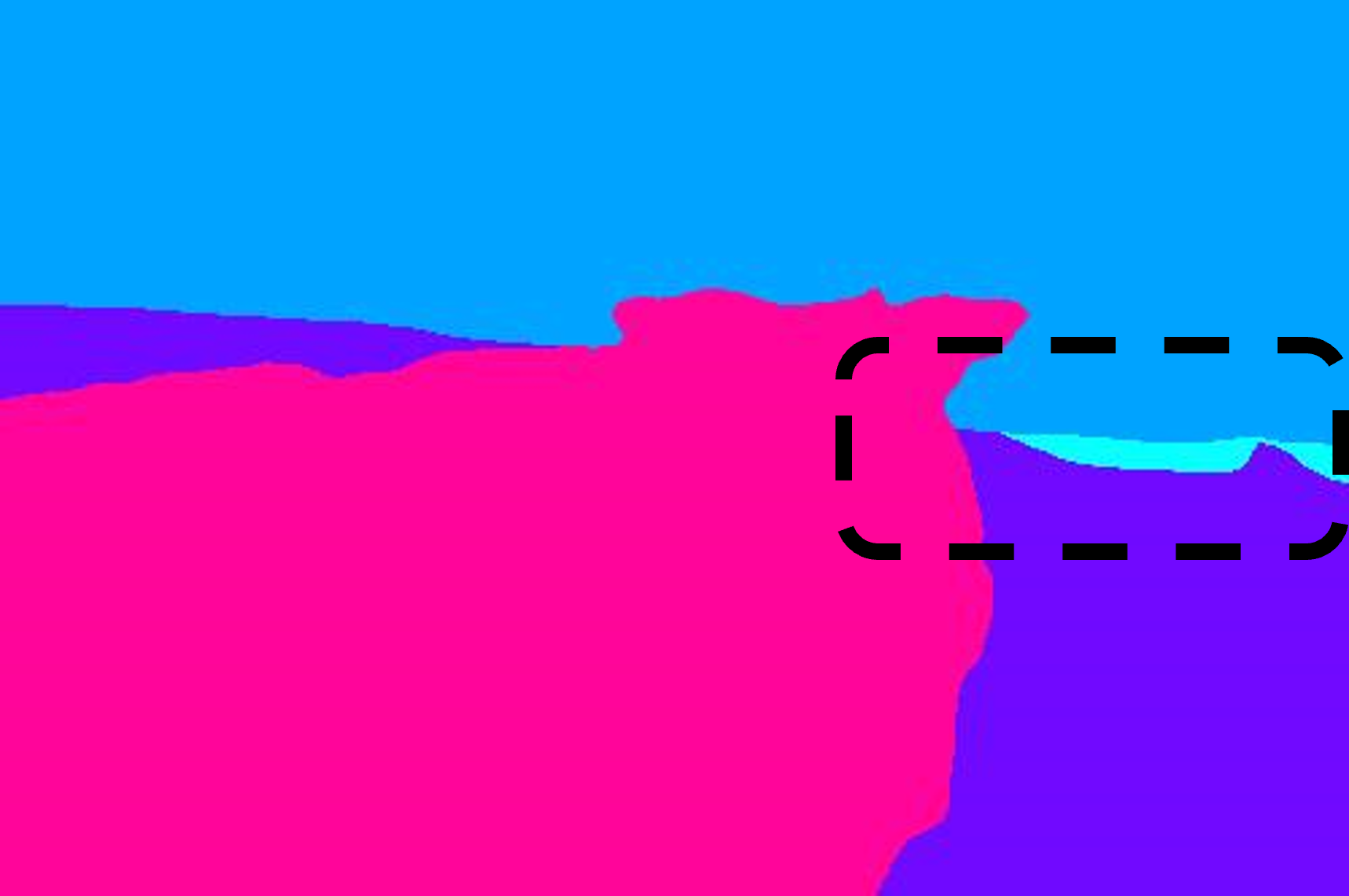}
        \caption*{(b) DLV3plus}
    \end{minipage}
    \begin{minipage}[b]{0.245\textwidth}
        \centering
        \includegraphics[width=4.45cm,height=1.9cm]{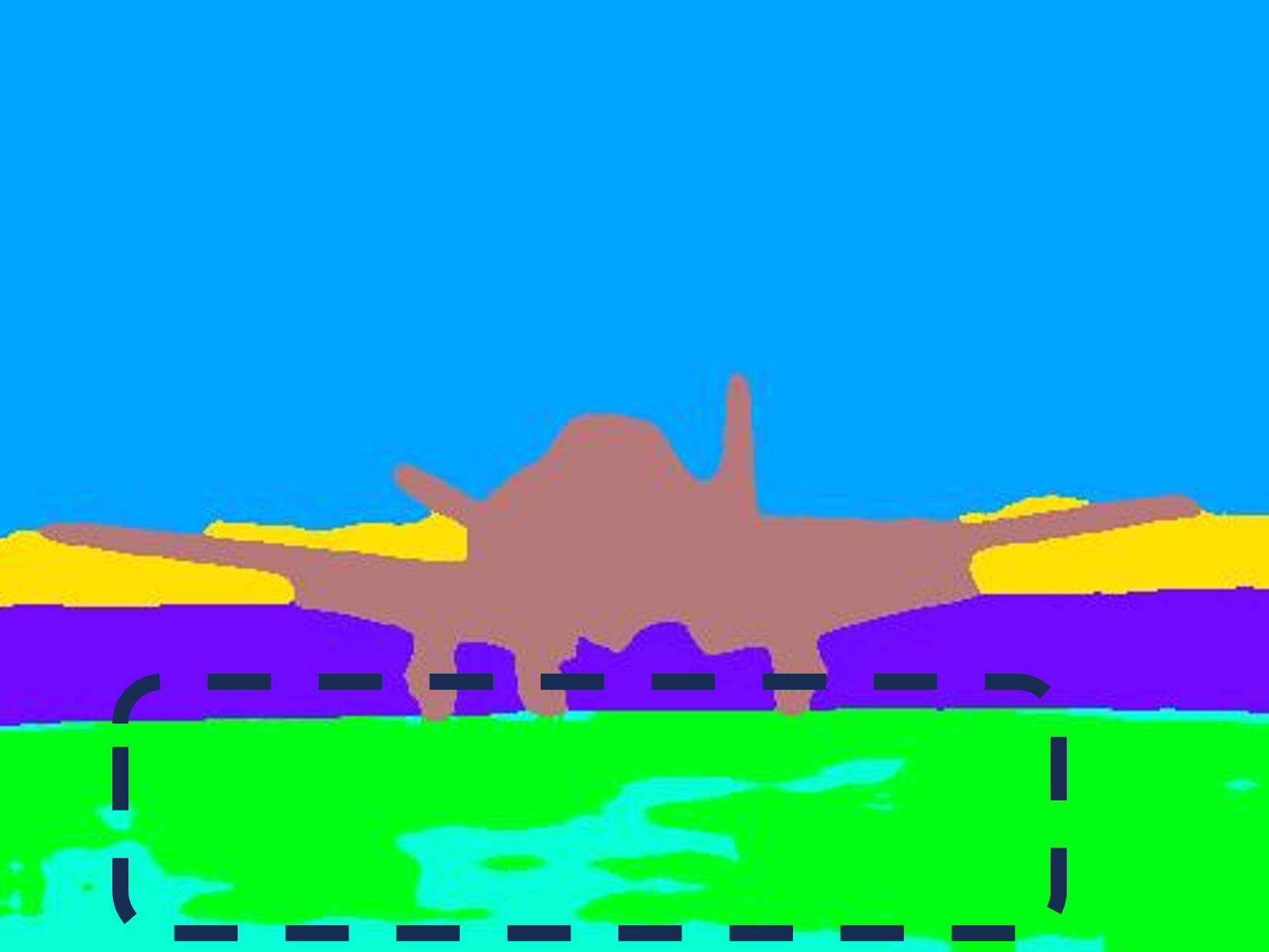}\vspace{1pt}
        \includegraphics[width=4.45cm,height=1.9cm]{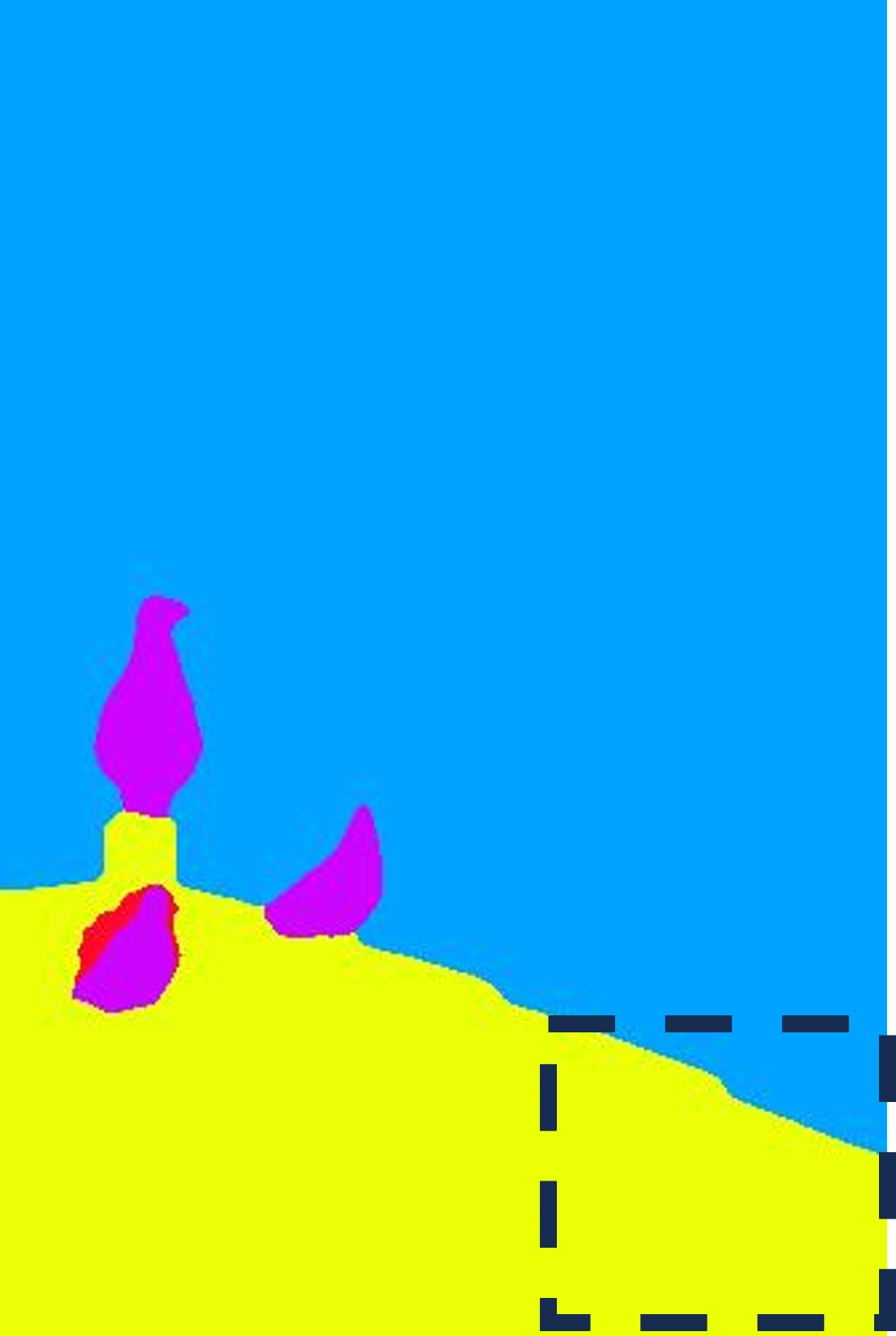}\vspace{1pt}
        \includegraphics[width=4.45cm,height=1.9cm]{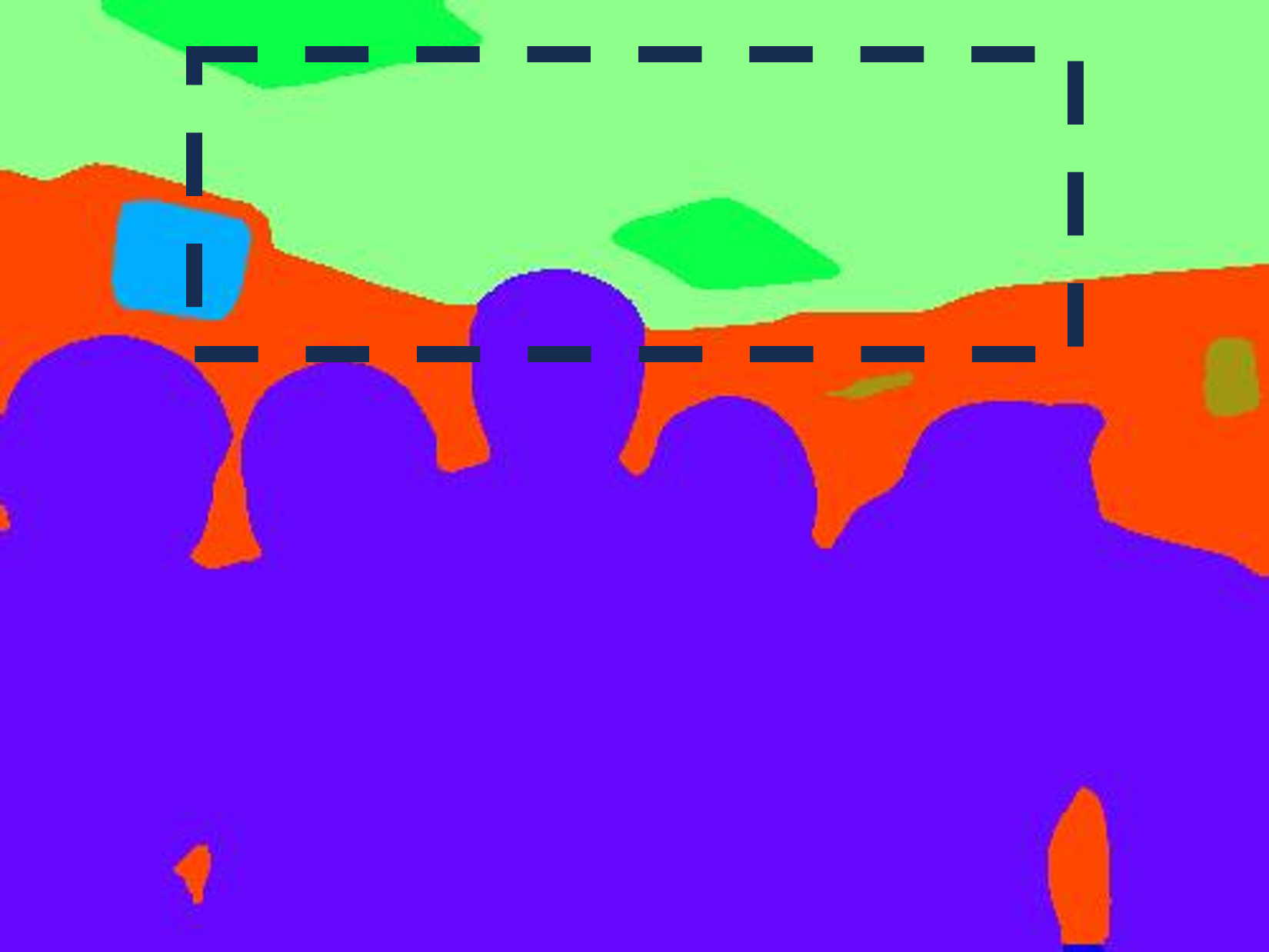}\vspace{1pt}
        \includegraphics[width=4.45cm,height=1.9cm]{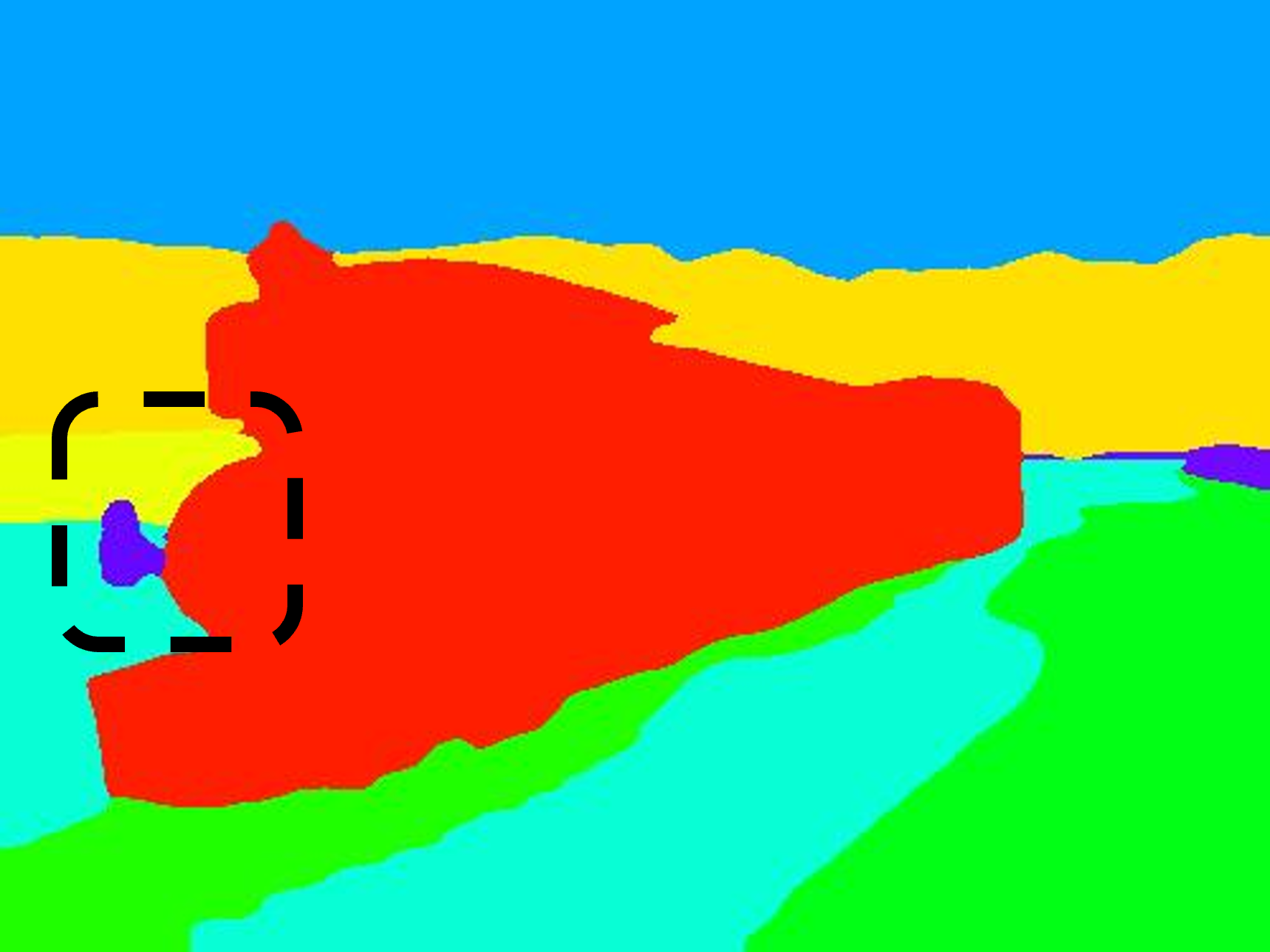}\vspace{1pt}
        \includegraphics[width=4.45cm,height=1.9cm]{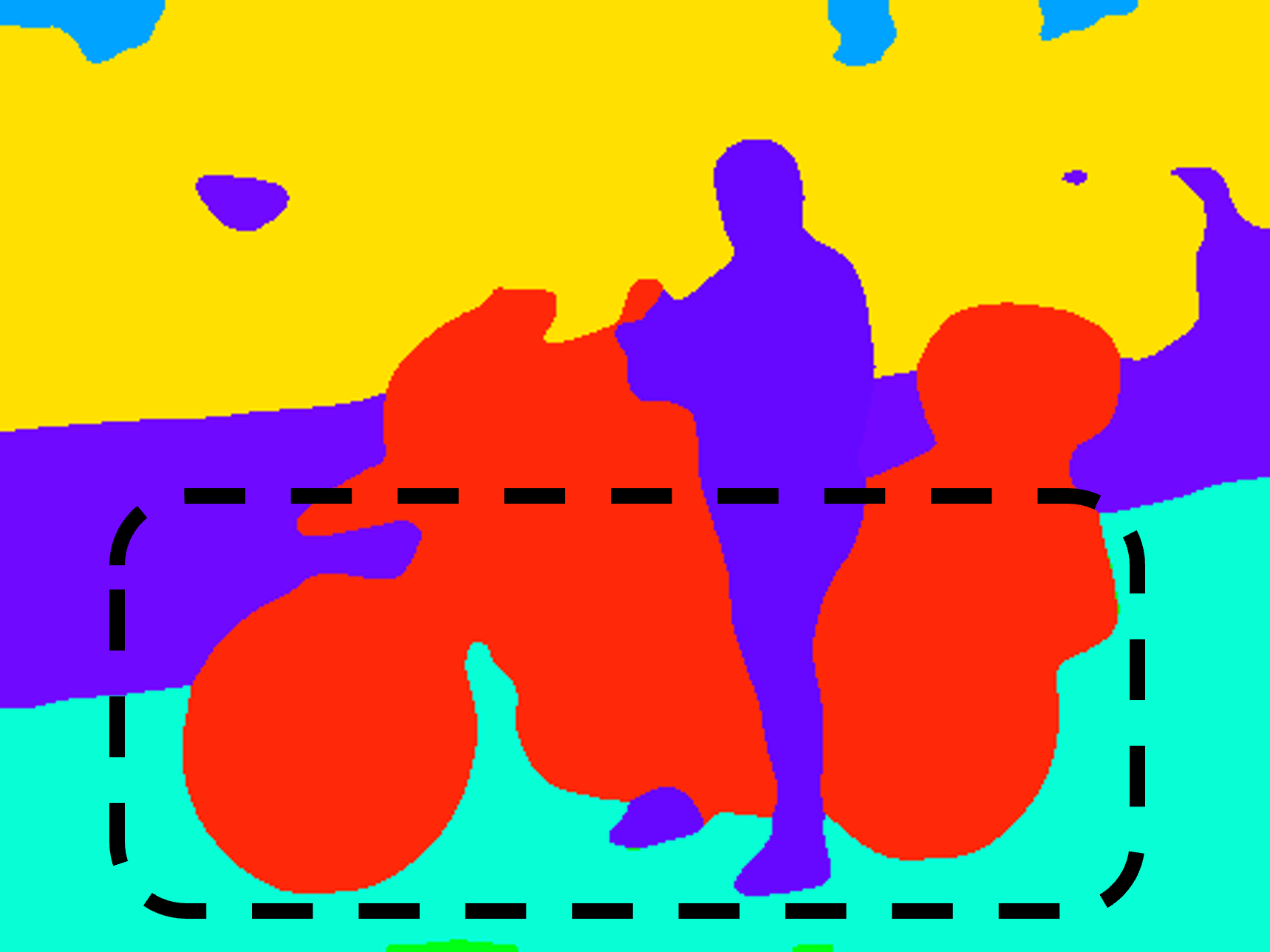}\vspace{1pt}
        \includegraphics[width=4.45cm,height=1.9cm]{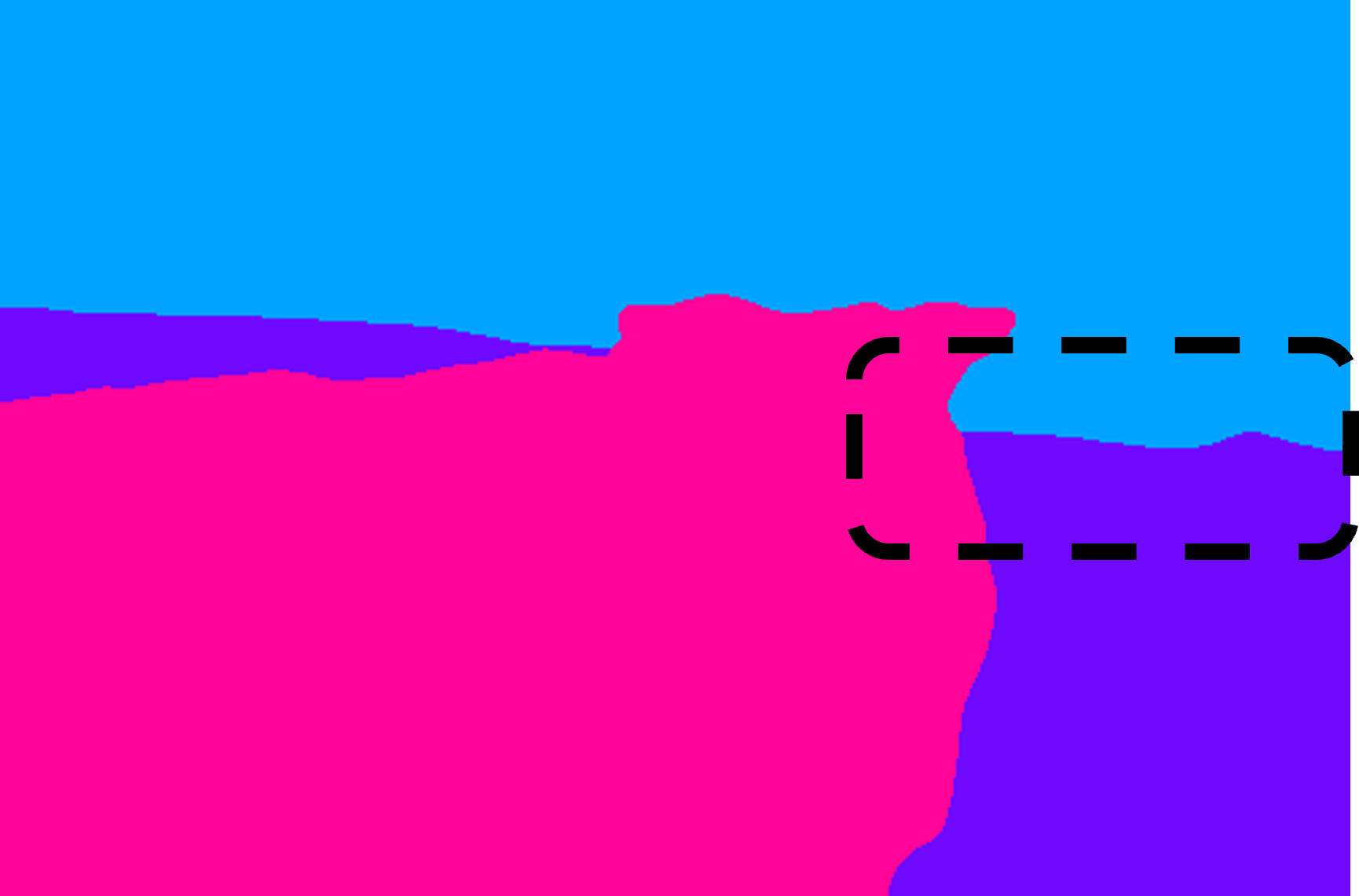}
        \caption*{(c) DLV3plus+ECAC(ours)}
    \end{minipage}
    \begin{minipage}[b]{0.245\textwidth}
        \centering
        \includegraphics[width=4.45cm,height=1.9cm]{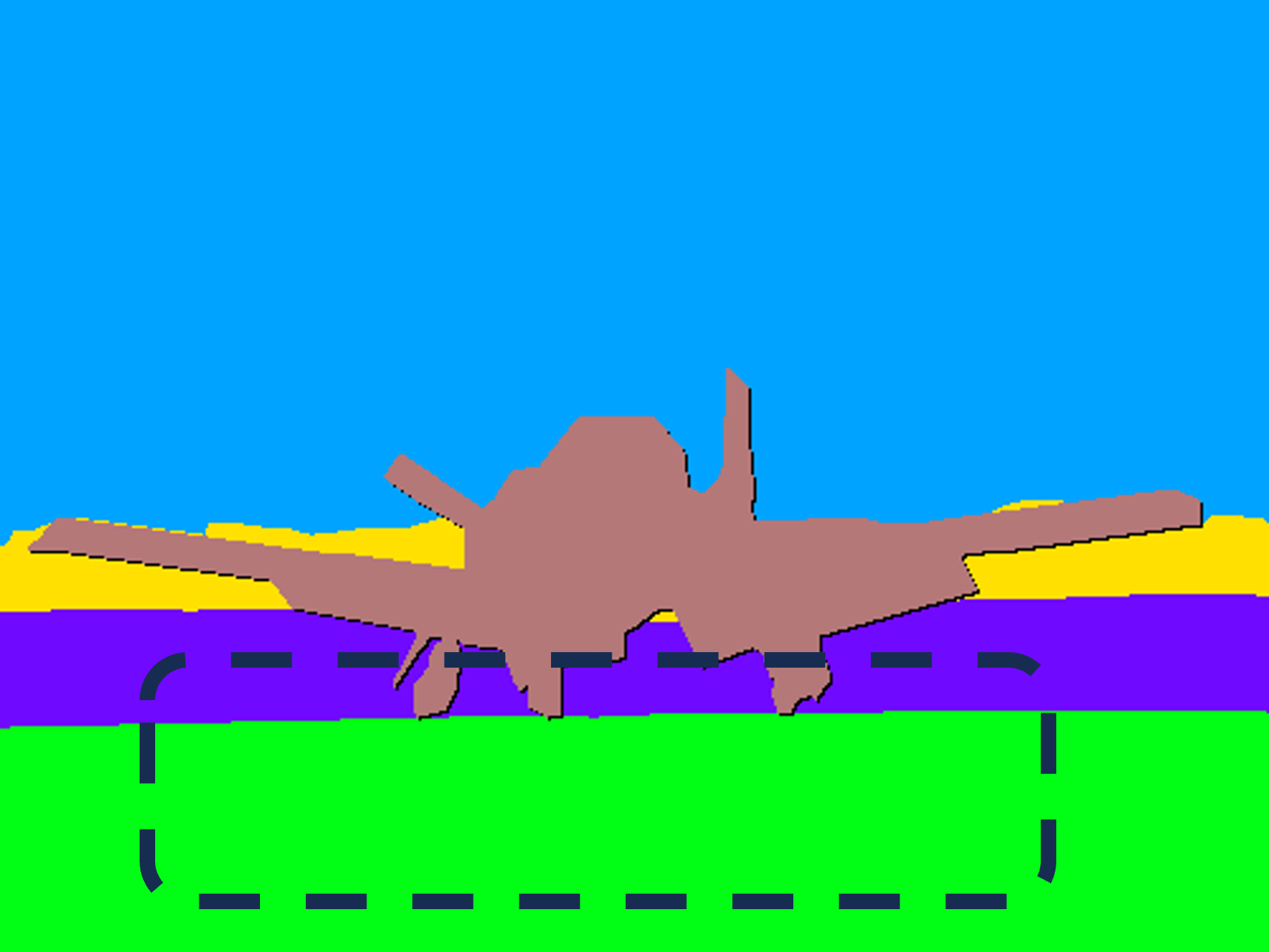}\vspace{1pt}
        \includegraphics[width=4.45cm,height=1.9cm]{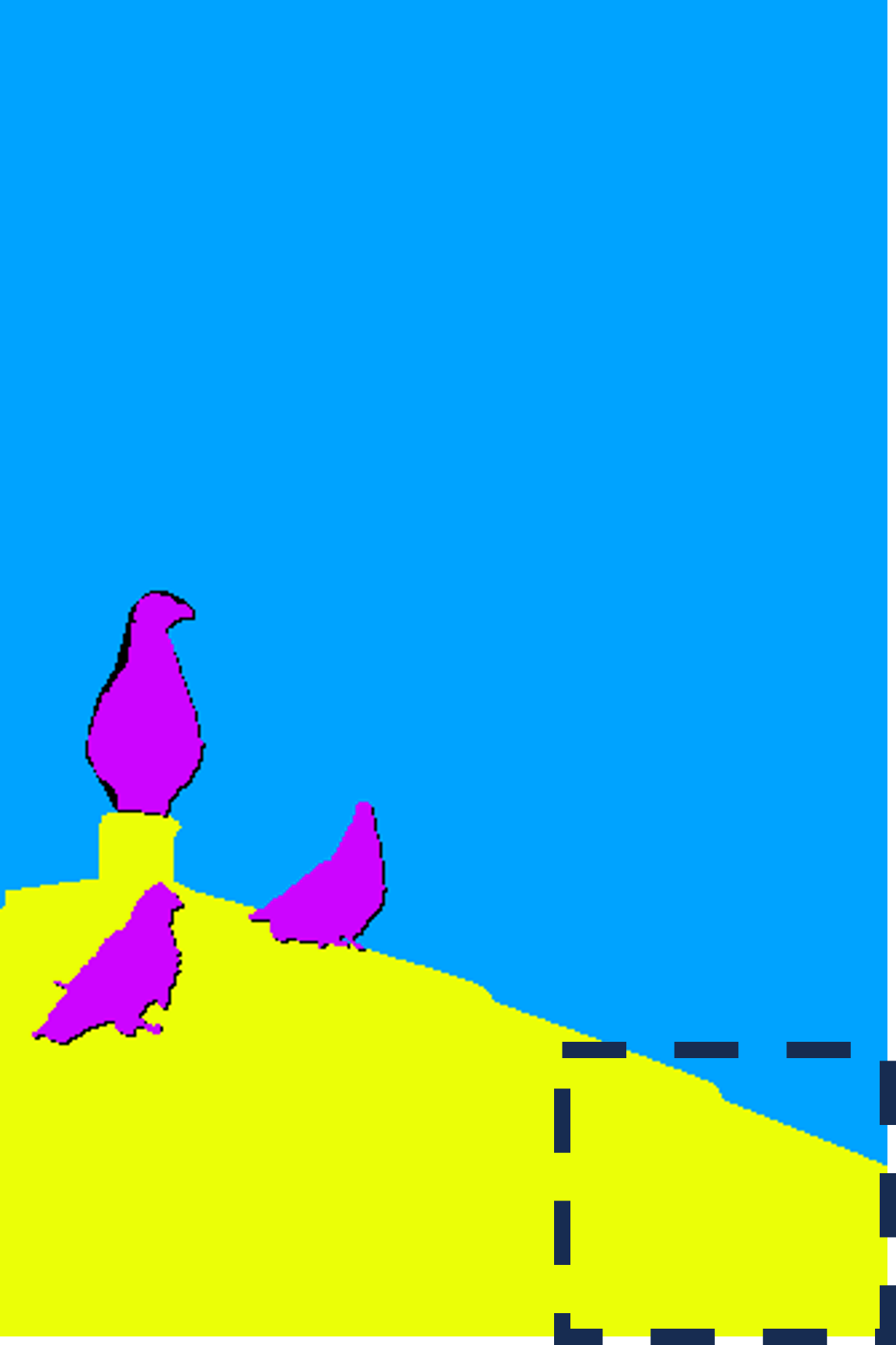}\vspace{1pt}
        \includegraphics[width=4.45cm,height=1.9cm]{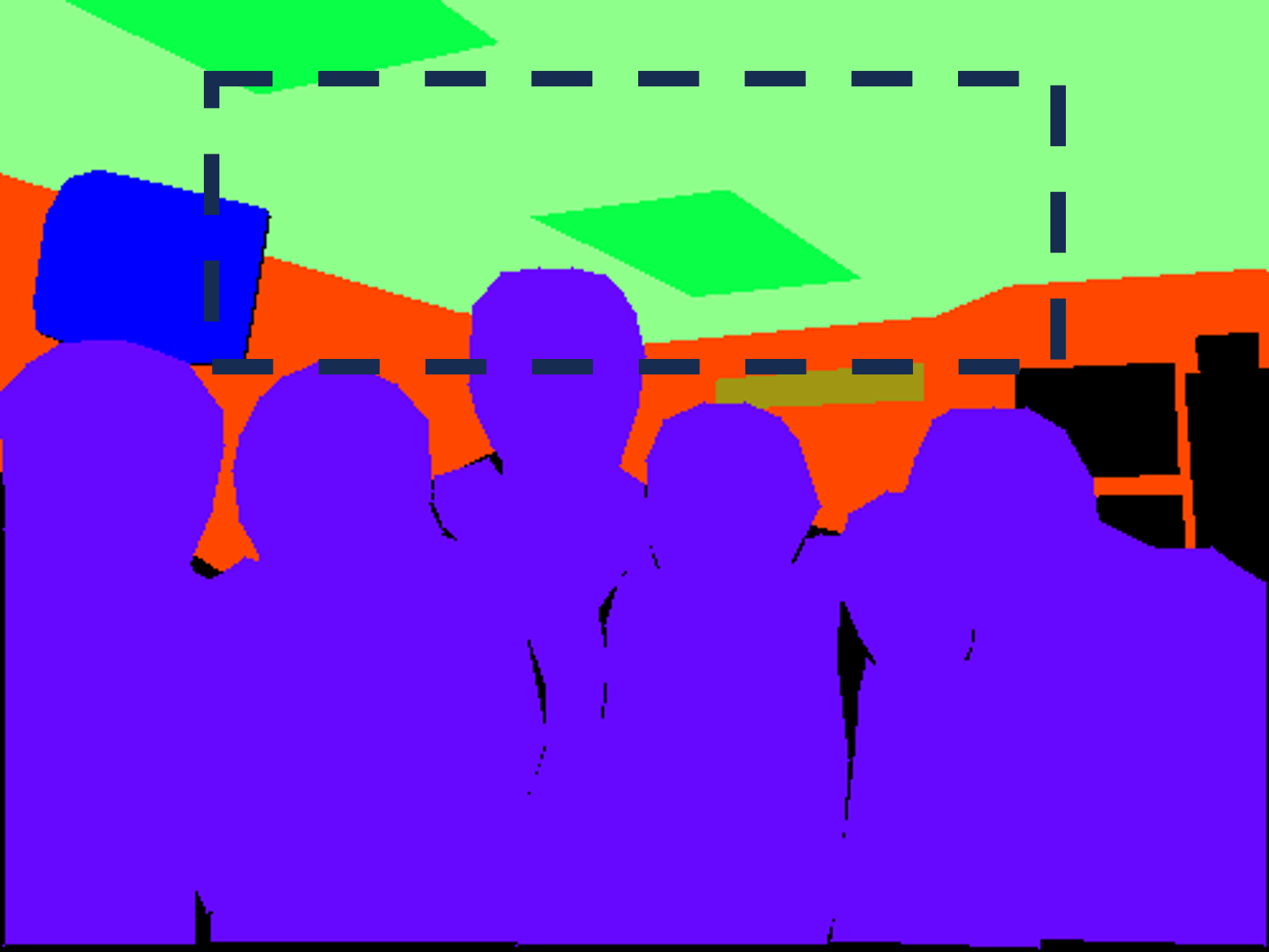}\vspace{1pt}
        \includegraphics[width=4.45cm,height=1.9cm]{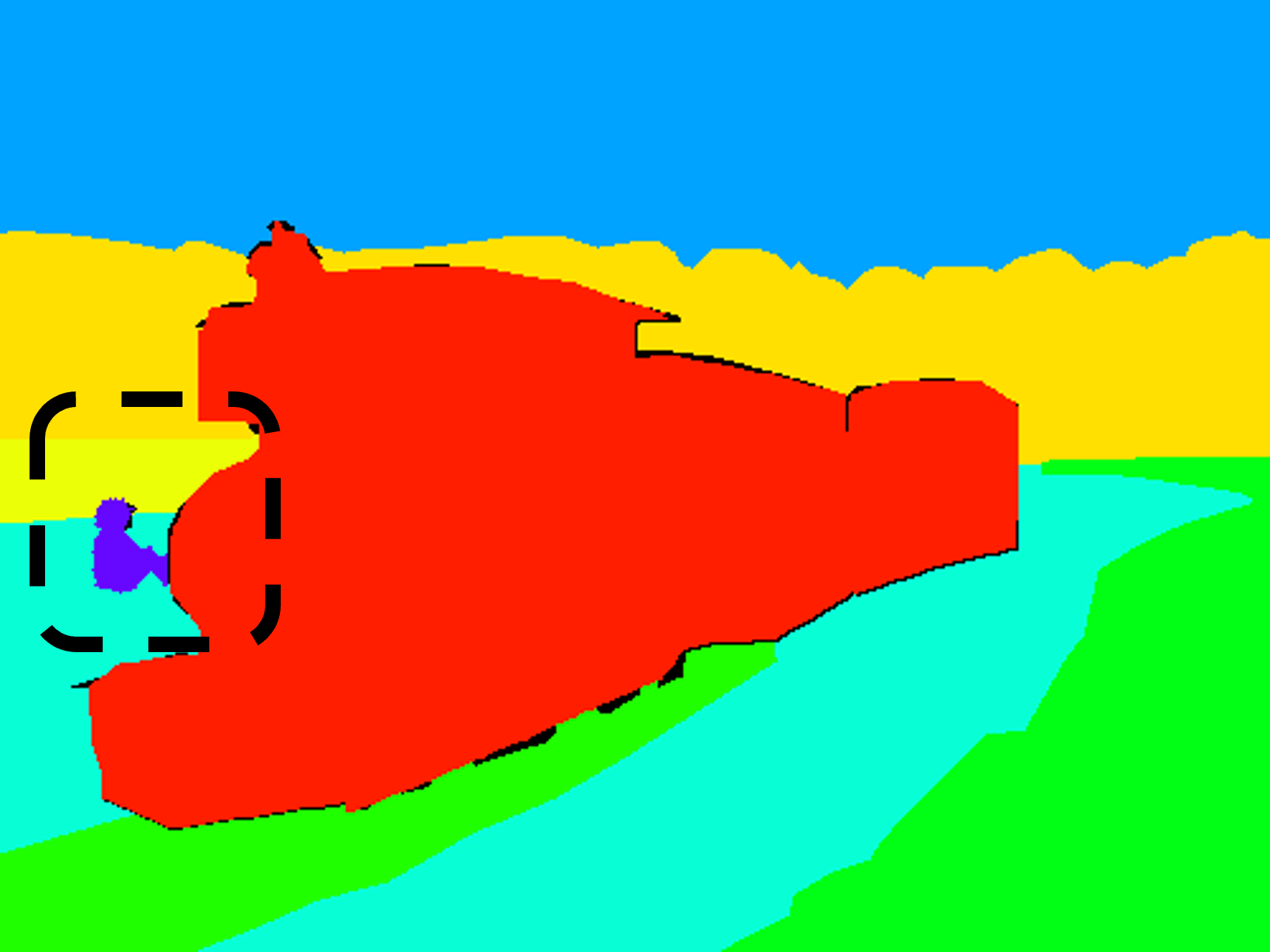}\vspace{1pt}
        \includegraphics[width=4.45cm,height=1.9cm]{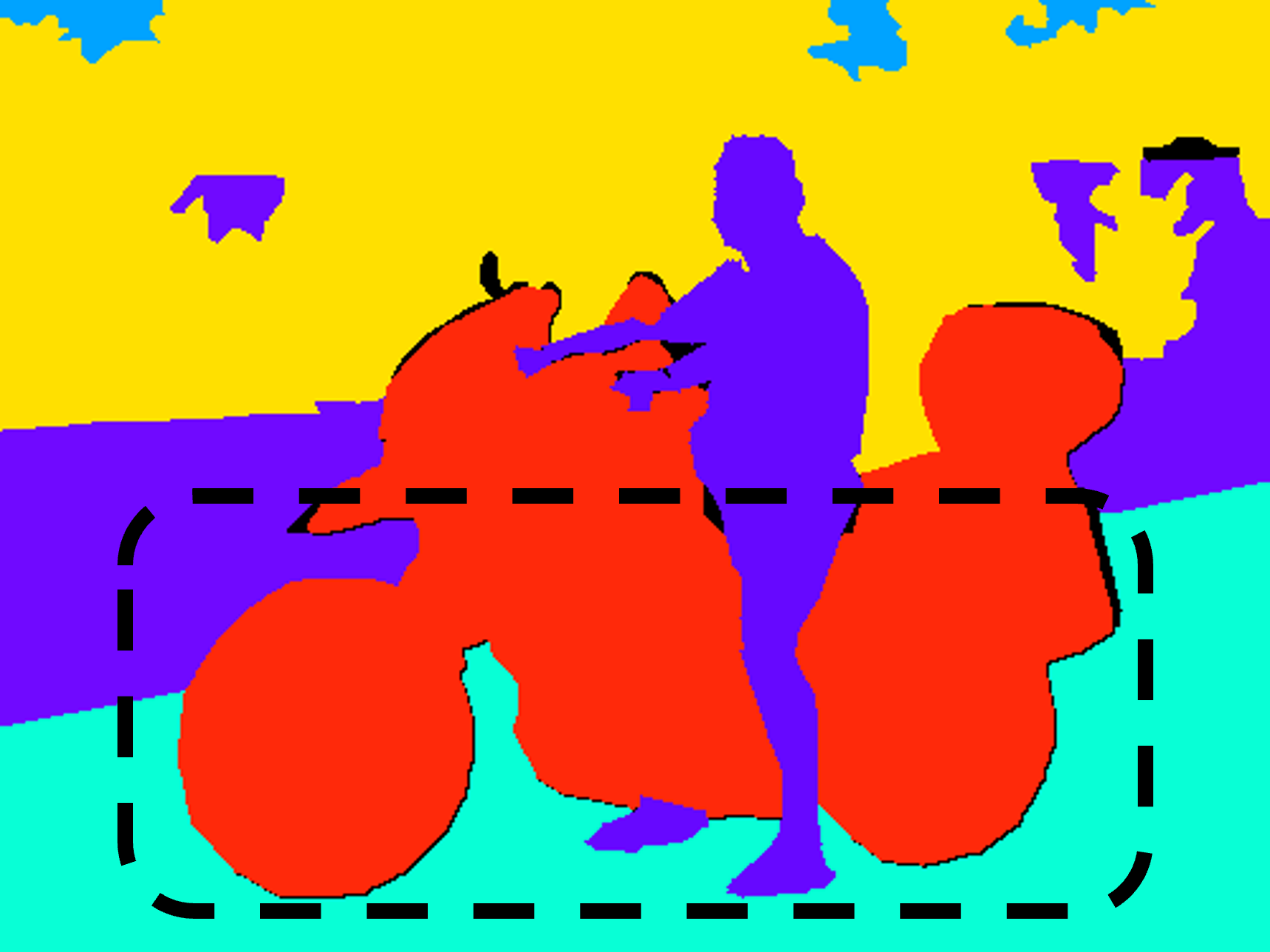}\vspace{1pt}
        \includegraphics[width=4.45cm,height=1.9cm]{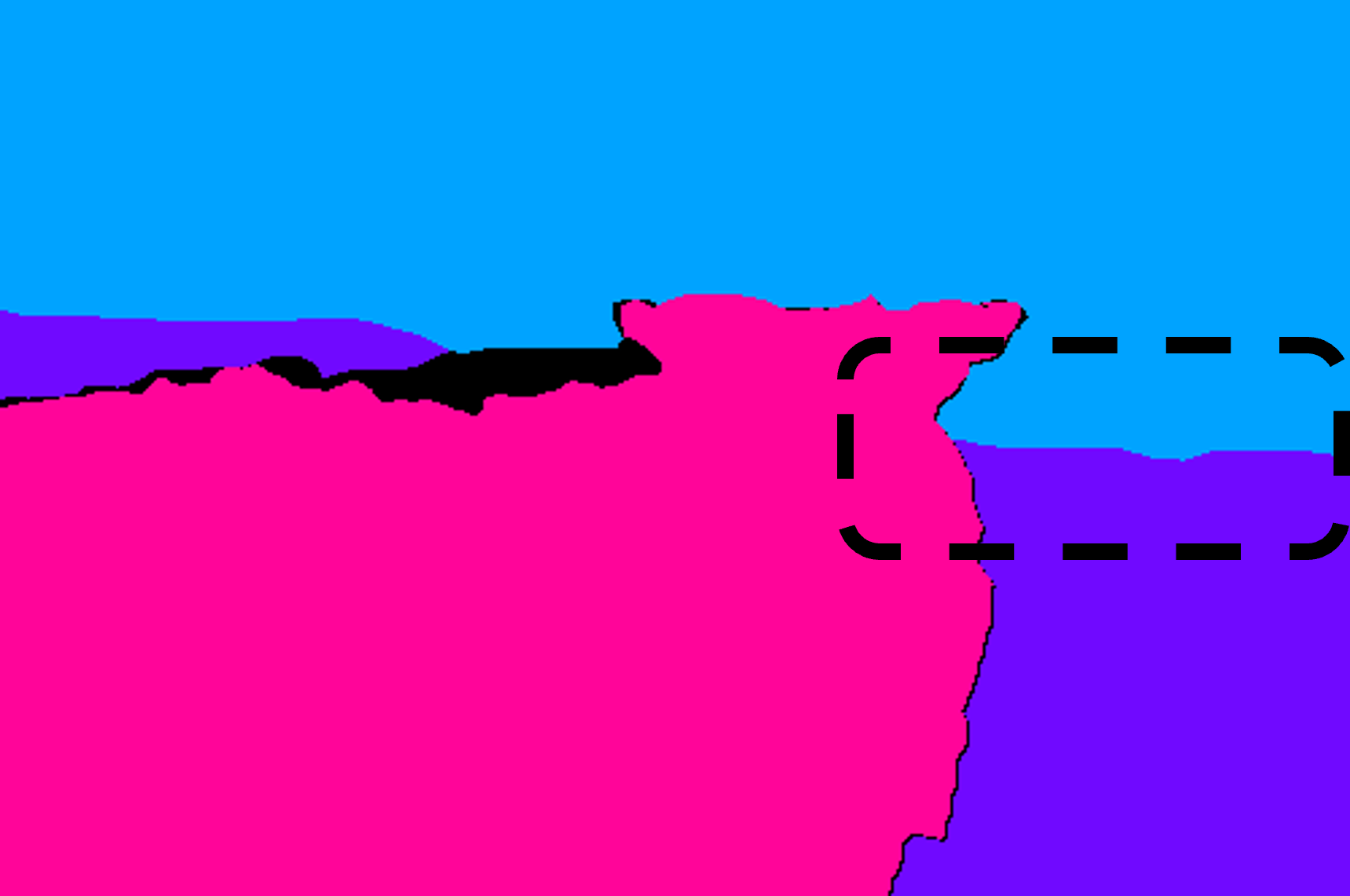}
        \caption*{(d) Ground truth}
    \end{minipage}
    \vspace{-5mm}
    \caption{Qualitative comparisons on Pascal Context test~\cite{pascalcontext}. The dotted boxes emphasize areas with significant improvements achieved through the application of the proposed ECAC. DLV3 is the abbreviation for DeeplabV3plus~\cite{deeplabv3+}.}
    \label{fig:pascal}
\end{figure*}
\subsection{Results}
Table~\ref{tab} summarizes the performance of state-of-the-art image segmentation methods on the ADE20K validation set, COCO-Stuff10K test set, and Pascal-Context test set, with mean Intersection over Union (mIoU) as the evaluation metric.
The table is divided into two main sections based on backbone architecture: CNN-based and Transformer-based methods. Each section lists methods with their respective backbones (~\emph{e.g.}, ResNet-101~\cite{resnet}, Swin-Large~\cite{swin}) and mIoU scores across the three datasets. 
We selected DeepLabV3plus~\cite{deeplabv3+} and UperNet~\cite{upernet} as the base architectures for integrating our proposed ECAC module due to their established effectiveness and complementary strengths in image segmentation tasks.
The CNN-based section starts with PSPNet and progresses to our DeepLabV3plus + ECAC. In contrast, the Transformer-based section begins with StructToken~\cite{StructToken} and culminates in our UperNet + ECAC, showcasing the impact of our approach across different architectures. 
Within the Transformer section, methods using the Swin-Large backbone are grouped to highlight performance trends, including GSS-FT~\cite{gss}, TSG~\cite{tsg}, CoT~\cite{cot}, and our UperNet + ECAC. Notably, our UperNet + ECAC with Swin-Large achieves the highest scores among Transformer-based methods, with mIoU values of 54.70\% on ADE20K, 50.49\% on COCO-Stuff10K, and 64.50\% on Pascal-Context.

Overall, the integration of our proposed ECAC module consistently enhanced the performance of both CNN and Transformer backbones, with the most notable gains observed in the Pascal-Context dataset, where our methods achieved the highest mIoU scores. Qualitative comparisons are presented in Figure \ref{fig:ade}, \ref{fig:coco} and \ref{fig:pascal}.

\begin{table}[]
\renewcommand\arraystretch{1.2}
{\footnotesize
\begin{center}
\begin{tabular}{cccccc|c} \hline
Baseline & ECAC &  $L_{rce}$ & $L_{mem}$ & $L_{KL}$ & $L_{mse}$ & mIoU(\%) \\ \hline
\checkmark & & &  &  &  & 45.76\\ 
\checkmark & &  \checkmark &  & &  & 46.56\\ 
\checkmark & \checkmark &\checkmark &  & &  & 49.10\\ 
\checkmark & \checkmark & \checkmark& \checkmark &  & & 50.21\\ 
\checkmark & \checkmark & \checkmark& \checkmark & \checkmark &  & 50.43\\  
\checkmark & \checkmark & \checkmark& \checkmark & \checkmark & \checkmark & \textbf{50.68}
\\  
\bottomrule
\end{tabular}
\end{center}}
\vspace{-1mm}
\caption{Ablation study about various loss combinations on Pascal-Context \textit{val}. $L_{rce}$, $L_{KL}$, $L_{mem}$ and $L_{mse}$ denote re-weighting cross-entropy loss, Kullback-Leibler (KL) divergence, memory loss and Mean Squared (L2) loss.} \label{abl_1}
\end{table}

\begin{table}
\renewcommand\arraystretch{1.2}
{\footnotesize
\begin{center}
\begin{tabular}{c|c} \hline
Loss  & mIoU(\%) \\ \hline
$L_{rceS} + L_{rceT} + L_M $ & 50.21\\ 
$L_{rceS} + L_{rceT} + L_M + 0.1\cdot L_{KL}$ & 50.22\\ 
$L_{rceS} + L_{rceT} + L_M + 1\cdot L_{KL}$ & \textbf{50.43}\\ 
$L_{rceS} + L_{rceT} + L_M + 10\cdot L_{KL}$ & 50.30\\\hline
$L_{rceS} + L_{rceT} + L_M + L_{KL}$ & 50.43\\ 
$L_{rceS} + L_{rceT} + L_M + L_{KL} + 50\cdot L_{iv}$ & \textbf{50.68}\\ 
$L_{rceS} + L_{rceT} + L_M + L_{KL} + 100\cdot L_{iv}$ & 50.58\\ 
$L_{rceS} + L_{rceT} + L_M + L_{KL} + 200\cdot L_{iv}$ & 50.45\\ 
\bottomrule
\end{tabular}
\end{center}}
\vspace{-1mm}
\caption{Ablation study about KL loss weight $\alpha$ and Mean Squared loss weight $\beta$ on Pascal-Context \textit{val}.} \label{abl_kdweight}
\end{table}

\begin{table}
\renewcommand\arraystretch{1.5}
{
\footnotesize
\begin{center}
\begin{tabular}{c|c|c} \hline
Method & Backbone  & mIoU(\%) \\ 
\hline
\textit{Baseline} & ResNet-50  & 45.76\\ 
\textit{Baseline} + ECAC(\emph{cosine})  & ResNet-50  & 50.10\\ 
\textit{Baseline} + ECAC(\emph{ours}) & ResNet-50 & \textbf{50.68}\\ \hline
\end{tabular}
\end{center}}
\vspace{-1mm}
\caption{Ablation study on the test set of Pascal-Context \textit{val} about the update strategy of the memory bank.} \label{abl_2}
\end{table}

\begin{table}
\renewcommand\arraystretch{1.5}
{
\footnotesize
\begin{center}
\begin{tabular}{c|c|c} \hline
Method & Backbone  & mIoU(\%) \\ 
\hline
\textit{DeeplabV3plus} & ResNet-50  & 50.73\\ 
\textit{DeeplabV3plus} + ECAC & ResNet-50  & 53.06\\  
\textit{DeeplabV3plus} + ECAC + Calibration & ResNet-50 & \textbf{53.56}\\ \hline
\end{tabular}
\end{center}}
\vspace{-1mm}
\caption{Ablation study on the test set of Pascal-Context \textit{val} about the output calibration.} \label{abl_cali}
\end{table}
\begin{table}[]
\centering
\renewcommand\arraystretch{1.1}
{\footnotesize 
  \resizebox{\columnwidth}{!}{
  \begin{tabular}{c|c|c|c} \hline
  Method & Params(M) & FLOPs(G) & mIoU (\%) \\ \hline
  FCN & 11.87 & 47.72 & 36.10\\
  + MCIBI & 16.41($\uparrow4.54$) & 67.22($\uparrow19.50$) & 42.84($\uparrow6.74$)\\
  + MCIBI++~\cite{mcibi++} & 15.45($\uparrow3.58$) & 64.44($\uparrow16.72$) & 43.39($\uparrow7.29$)\\
  + ECAC & \textbf{12.90($\uparrow1.03$)} & \textbf{49.17($\uparrow1.45$)} & \textbf{43.12($\uparrow7.02$)}\\ \hline
  PSPNet & 23.14 & 77.71 & 42.48\\
  + MCIBI & 36.33($\uparrow13.19$) & 131.75($\uparrow54.04$) & 43.77($\uparrow1.29$)\\
  + MCIBI++~\cite{mcibi++} & 33.99($\uparrow10.85$) & 122.16($\uparrow44.45$) & 43.89($\uparrow1.41$)\\
  + ECAC & \textbf{24.19($\uparrow1.05$)} & \textbf{78.84($\uparrow1.13$)} & \textbf{45.47($\uparrow2.99$)}\\ \hline
  DeeplabV3plus & 17.74 & 65.97 & 43.95\\
  + MCIBI & 29.68($\uparrow11.94$) & 117.34($\uparrow51.37$) & 44.34($\uparrow0.39$)\\
  + MCIBI++~\cite{mcibi++} & 28.59($\uparrow10.85$) & 110.42($\uparrow44.45$) & 44.85($\uparrow0.9$)\\
  + ECAC & \textbf{18.76($\uparrow1.02$)} & \textbf{67.22($\uparrow1.25$)} & \textbf{46.32($\uparrow2.37$)}\\ \hline
  \end{tabular}
  }
}
\vspace{-1mm}
\caption{Comparison of inference computational complexity and accuracy on ADE20K \textit{val}.}
\label{complex_table}
\end{table}

\begin{table}
\renewcommand\arraystretch{1.5}
{\footnotesize
\resizebox{\columnwidth}{!}{
\begin{tabular}{c|c|c|c|c} \hline
Method & Backbone & ADE20K & COCO-Stuff & Pascal-Context \\ \hline
FCN~\cite{fcn}  & ResNet-50 & 36.10 & 34.10 & 45.76\\ 
FCN + ECAC& ResNet-50 & 43.12 $(\uparrow{7.02})$ & 36.68 $(\uparrow{2.58})$ & 50.68 $(\uparrow{4.92})$\\ \hline
UperNet~\cite{upernet}  & ResNet-50 & 42.05 & 35.80 & 48.90\\ 
UperNet + ECAC  & ResNet-50 & 45.29 $(\uparrow{1.76})$& 37.61 $(\uparrow{1.81})$ & 53.13 $(\uparrow{4.23})$\\ \hline
PSPNet~\cite{pspnet}  & ResNet-50 & 42.48 & 36.33 & 50.21\\ 
PSPNet + ECAC& ResNet-50 & 45.47 $(\uparrow{2.99})$ & 38.32 $(\uparrow{1.99})$ & 53.47 $(\uparrow{3.26})$\\ \hline
DeeplabV3plus~\cite{deeplabv3+}  & ResNet-50 & 43.95 & 36.92 & 50.73\\ 
DeeplabV3plus + ECAC  & ResNet-50 & 46.32 $(\uparrow{2.37})$& $40.19 (\uparrow{3.27})$ & 53.56 $(\uparrow{2.83})$\\ \hline
\end{tabular}}}
\vspace{-1mm}
\caption{The performance of ECAC's integration into mainstream frameworks on various benchmarks. We adopt single-scale testing for the sake of convenience and simplicity, which is different from Table~\ref{tab}.} \label{tabnew}
\end{table}

\subsection{Ablation Study}
Here, we present ablation studies on the Pascal-Context to investigate the effectiveness of our proposed method. 
ResNet-50 is selected as the backbone for the FCNs~\cite{fcn} baseline due to its lower computational cost compared to ResNet-101, ViT, and Swin-Transformer.
Unless otherwise specified, all experiments in this section are performed using \textbf{single-scale testing} for simplicity and fair comparison, which differs from the Table~\ref{tab}.

\subsubsection{Loss function}
Our ECAC framework incorporates a combination of re-weighting cross-entropy loss, memory loss, KL divergence loss, and MSE loss to enhance segmentation performance. To investigate the contribution of each loss component, we conduct an ablation study, and the results are summarized in Table~\ref{abl_1}.

We achieve 0.8\% mIoU improvement using the re-weighting cross-entropy loss. Utilizing ECAC without memory loss, we achieve 49.10\% mIoU. When memory loss is included, the performance improves significantly, reaching a mIoU of 50.21\%. 
Further, adding KL loss alongside memory loss leads to a mIoU of 50.43\%. The KL divergence loss helps align the distributions of the teacher and student models, enabling more efficient knowledge transfer and improving the segmentation accuracy. This result highlights the complementary nature of memory loss and KL loss in refining the learned representations.
Finally, combining memory loss, KL loss, and MSE loss achieves the highest mIoU of 50.68\%. The combination of these three losses allows ECAC to leverage both inter-class relationships and intra-class consistency effectively.

We additionally present experiments on Pascal-Context to verify the appropriate loss weight for the KL loss $\alpha$ and MSE loss $\beta$ in Table \ref{abl_kdweight}. The results show that the best performance, with a mIoU of 50.68\%, is achieved when  $\alpha=1$ and $\beta=50$.

\subsubsection{Update strategy of memory bank}
Table~\ref{abl_2} shows the results of different update strategies on the memory bank. 
Our memory bank is updated by employing ground-truth masks calculated class center. 
Compared with the memory bank update strategy in MCIBI~\cite{mcibi}, which uses a simple cosine similarity strategy, our method improves the mIoU from 50.10\% to 50.68\%. 
Therefore, we adopt the masks calculated class center for updating the memory bank.

The improvement demonstrates that using ground-truth mask-calculated class centers allows for a more precise and consistent representation of each class within the memory bank. Unlike the cosine similarity strategy, which relies solely on feature alignment, our approach benefits from directly incorporating spatial and semantic information from the ground-truth, leading to enhanced model performance.

\subsubsection{Ablation about output calibration}
We also conduct an ablation study on output calibration in Table~\ref{abl_cali}. As we mentioned in section~\ref{3.2}, the calibration strategy aligns the output with a reference class distribution, ensuring more balanced and accurate predictions. We gain 0.5\% mIoU improvement compared with our ECAC.

\subsubsection{Computational Complexity}
Table \ref{complex_table} presents the efficiency of ECAC integrated with existing segmentation frameworks on ADE20K using a ResNet-50 backbone.
The results are derived from an input size $512\times64\times64$ (an 8$\times$ down-sampling of $512\times512$).
Compared with MCIBI~\cite{mcibi}, we achieve a significant improvement in accuracy with only a marginal increase in the number of parameters. For instance, incorporating ECAC with PSPNet, we achieve a mIoU of 45.47\%, with 1.05M extra parameters and 1.13G more FLOPs. MICBI~\cite{mcibi} achieves 43.77\% mIoU but increases 13.19M parameters and 54.04 GFLOPs. 
\subsubsection{Integrated with Various Semantic Segmentation Frameworks}
ECAC fits effortlessly into current segmentation frameworks. 
To validate the performance and stability of our strategy, we integrate ECAC into four popular segmentation frameworks: FCN~\cite{fcn}, PSPNet~\cite{pspnet}, UperNet~\cite{upernet} and DeepLabVplus~\cite{deeplabv3+}. 
Table \ref{tabnew} illustrates the comparative performance analysis on three segmentation datasets: ADE20k, COCO-Stuff10K, and Pascal-Context. To ensure fairness, we perform single-scale testing. Integrating ECAC into the vanilla segmentation framework (e.g., FCN) significantly improves mIoU by 7.02\% on ADE20K, 2.58\% on COCO-Stuff, and 4.92\% on Pascal-Context. With the stronger baseline DeeplabV3plus, ECAC boosts mIoU from 43.95\% to 46.32\% on ADE20K, 36.92\% to 40.19\% on COCO-Stuff, and 50.73\% to 53.56\% on Pascal-Context. These results highlight the flexibility and efficacy of ECAC.

\begin{figure*}[]
    \centering
    \begin{minipage}{0.9\linewidth}  
        \centering
        \begin{minipage}{0.68\linewidth}
            \centering
            \subfigure[DeeplabV3plus]{
                \includegraphics[width=0.47\linewidth]{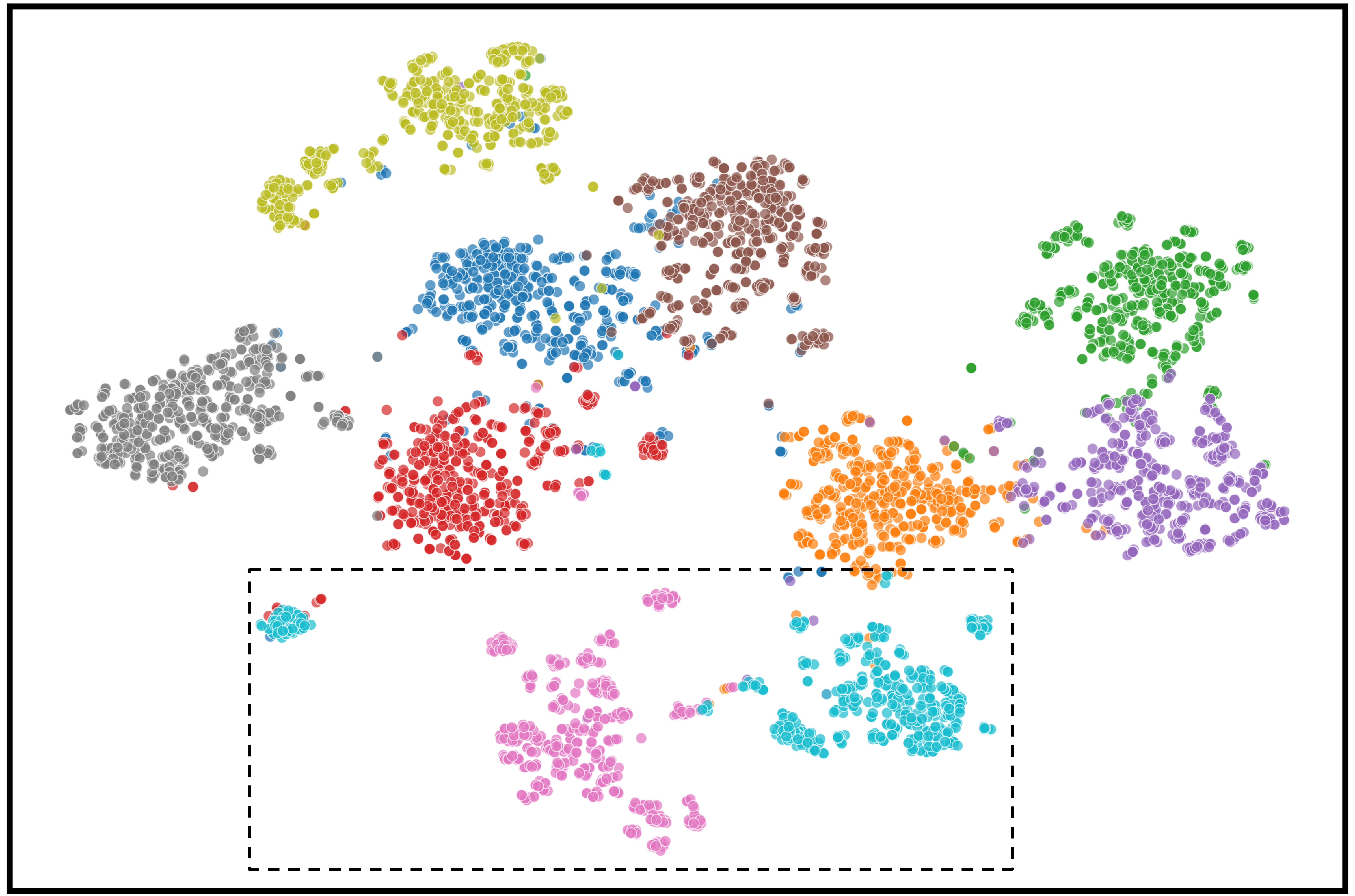}
            }
            \hfill
            \subfigure[DeeplabV3plus+ECAC]{
                \includegraphics[width=0.47\linewidth]{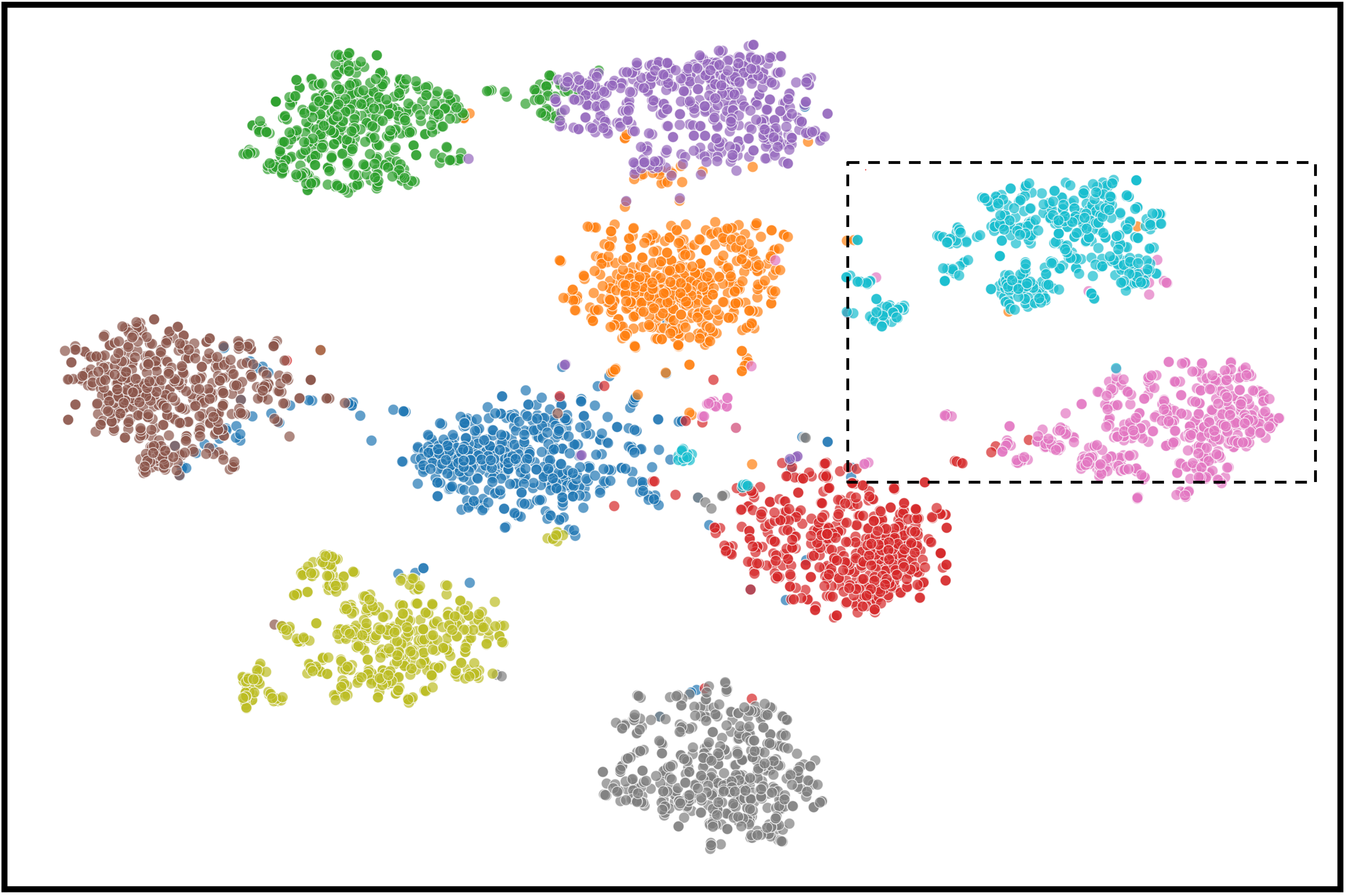}
            }
        \end{minipage}
        \hspace{4mm}  
        \begin{minipage}{0.1\linewidth}
        \centering
        \begin{tikzpicture}[every node/.style={font=\footnotesize}]
            \fill[tabblue] (-0.5, -0.5) circle (2pt);
            \node[anchor=west] at (0, -0.5) {bed};
            \fill[taborange] (-0.5, -0.9) circle (2pt);
            \node[anchor=west] at (0, -0.9) {building};
            \fill[tabgreen] (-0.5, -1.3) circle (2pt);
            \node[anchor=west] at (0, -1.3) {car};
            \fill[tabred] (-0.5, -1.7) circle (2pt);
            \node[anchor=west] at (0, -1.7) {ceiling};
            \fill[tabpurple] (-0.5, -2.1) circle (2pt);
            \node[anchor=west] at (0, -2.1) {curtain};
            \fill[tabbrown] (-0.5, -2.5) circle (2pt);
            \node[anchor=west] at (0, -2.5) {floor};
            \fill[tabpink] (-0.5, -2.9) circle (2pt);
            \node[anchor=west] at (0, -2.9) {road};
            \fill[tabgray] (-0.5, -3.3) circle (2pt);
            \node[anchor=west] at (0, -3.3) {sky};
            \fill[tabyellowgreen] (-0.5, -3.7) circle (2pt);
            \node[anchor=west] at (0, -3.7) {tree};
            \fill[tabcyan] (-0.5, -4.1) circle (2pt);
            \node[anchor=west] at (0, -4.1) {wall};
        \end{tikzpicture}
    \end{minipage}
    \end{minipage}
    \vspace{-2mm}
    \caption{Visualization of features distribution learned with DeeplabV3plus~\cite{deeplabv3} (left) and our ECAC (right).}
    \label{tsne}
\end{figure*}

\begin{table*}[]
\renewcommand\arraystretch{1.2}
\setlength{\tabcolsep}{1.2mm}
\centering
\begin{tabular}{c|cccccccccc|c} \hline
\toprule
\rowcolor{gray!20}
 Method (Major Class) & wall & building  & sky & floor & tree & ceiling & road & bed & windowpane & grass & mIoU(\%)\\ \hline
DeeplabV3plus~\cite{deeplabv3+}& 75.53 & 82.18 & 94.09 & 79.11 & 73.61 & 82.63 & 82.0 & 87.52 & 60.06 & 65.05 & 78.18\\  
DeeplabV3plus + ECAC & 76.02 & 82.21 & 93.99 & 79.43 & 73.81 & 82.79 & 82.68 & 87.79 & 61.73 & 69.79 & 79.02\\ \hline 
\rowcolor{gray!20}
 Method (Moderate Class)& bench & countertop  &  stove & palm & kitchen island & computer & swivel chair & boat & bar & arcade machine & mIoU(\%)\\ \hline
 DeeplabV3plus~\cite{deeplabv3+} & 38.59 & 54.02 & 74.64 & 50.87 & 43.74 & 55.5 & 46.83 & 42.83 & 24.12 & 29.48 & 46.06\\
 DeeplabV3plus + ECAC & 40.44 & 59.91 & 75.84 & 49.31 & 42.88 & 55.46 & 41.73 & 76.45 & 26.49 & 47.38 & 51.59\\\hline
 \rowcolor{gray!20}
 Method (Minor Class)& pier & crt screen  &  plate & monitor & bulletin board & shower & radiator & glass & clock & flag & mIoU(\%)\\ \hline
 DeeplabV3plus~\cite{deeplabv3+} & 25.11 & 0.33 & 42.83 & 55.34 & 35.98 & 0.0 & 41.73 & 9.42 & 22.7 & 34.1 & 26.75\\
 DeeplabV3plus + ECAC & 32.58 & 0.37 & 47.32 & 73.60 & 27.58 & 12.20 & 46.00 & 16.74 & 29.81 & 32.90 & 31.91\\
\bottomrule
\end{tabular}
\caption{\textbf{Comparison of each category in ADE20k}. We show the major 10  categories, the moderate 10 categories, and the 10 minor categories.
} \label{freq}
\end{table*}
\subsubsection{Per-Class Analysis}
We analyze the performance of DeepLabV3plus + ECAC (ResNet-101 as the backbone) on the ADE20k dataset to assess its impact on class imbalance, as shown in Table \ref{freq}. 
For the major categories, DeepLabV3plus~\cite{deeplabv3+} achieves an mIoU of 78.18\%, which improves to 79.02\% with ECAC. 
Per-class results reveal significant improvements in categories like ``windowpane" (60.06\% to 61.73\%) and ``grass" (65.05\% to 69.79\%), though dominant classes such as ``sky" show a slight decline (94.09\% to 93.99\%). 
For moderate categories, the mIoU increases from 46.06\% to 51.59\% with ECAC, with notable gains in ``arcade machine" (29.48\% to 47.38\%) and ``palm" (50.87\% to 53.74\%). 
For minor categories, the mIoU improves substantially from 26.75\% to 31.91\%, with significant gains in ``shower" (0.0\% to 12.20\%) and ``pier" (25.11\% to 32.58\%). However, ``bulletin board" declines from 35.98\% to 27.58\%, likely due to its limited samples and high visual variability, suggesting that certain category-specific factors may require further refinement.
Per-class analysis indicates that ECAC enhances feature representation for minor classes by leveraging contextual cues, as seen in the improved performance for rare categories. 
However, inconsistencies in certain minor classes, such as ``bulletin board," suggest potential overfitting or sensitivity to specific class characteristics. 
Overall, ECAC improves segmentation performance for both moderate and minor classes.

\subsubsection{Visualization of Learned Features}
To investigate the feature representations learned by our model, we visualize the feature distributions using t-SNE for a subset of 10 randomly selected classes from the ADE20K dataset, as shown in Fig~\ref{tsne}.
The left plot (a) depicts the feature space of DeepLabV3plus~\cite{deeplabv3+}, where clusters show noticeable overlap, indicating limited class discriminability.
This overlap suggests that DeepLabV3plus struggles to learn distinct feature representations, potentially leading to suboptimal segmentation performance.
In contrast, the right plot (b) illustrates the feature space of our proposed DeepLabV3plus+ECAC, which exhibits tighter and better-separated clusters across the selected classes.
This improvement highlights ECAC's ability to enhance feature discriminability, enabling better differentiation between classes.
However, some overlap remains in the ECAC feature space, suggesting that while our approach significantly improves feature representations, there is still room for further refinement to achieve optimal class separation across the entire dataset.

\vspace{-1mm}
\section{Limitations and Future Works}
Our Extended Context-Aware Classifier (ECAC) effectively addresses class imbalance and enhances segmentation by leveraging global and local contextual information, achieving superior performance on ADE20K, COCO-Stuff10K, and Pascal-Context benchmarks. However, challenges remain in handling extreme class imbalances and noisy datasets. Specifically, the memory bank struggles with rare classes (e.g., ``shower” and ``pier” in Table VIII) due to limited samples, and some minor classes (e.g., ``bulletin board,” 35.98\% to 27.58\% mIoU) show performance declines, indicating potential overfitting. In the future, we will explore adaptive memory bank update strategies to better capture rare class representations. Additionally, we will incorporate dynamic weighting mechanisms or contrastive learning to enhance robustness across diverse and noisy datasets.
\vspace{-1mm}
\section{Conclusion}
This paper presents an extended context-aware classifier by leveraging global (dataset-level) and local (image-level) class information. 
We first introduce the memory bank to store dataset-level category information. Then, we concatenate the memory bank with the class center of the current image to obtain an enhanced classifier. This will not cause too much computational complexity and will significantly improve performance. To further enhance the classifier, a teacher-student network is adopted to transfer comprehensive context from teacher to student. 
Lastly, ECAC serves as an add-on module for mainstream segmentation frameworks.
Comprehensive experiments validate the efficacy of our approach.




\ifCLASSOPTIONcaptionsoff
  \newpage
\fi

\bibliographystyle{IEEEtran}
\bibliography{egbib}
\vspace{-1.2cm}
\begin{IEEEbiography}[{\includegraphics[width=0.95in,height=1.2in,clip]{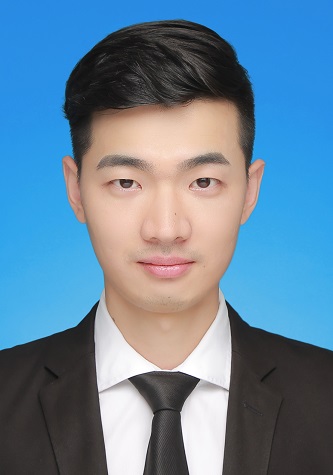}}]{Huadong Tang}
received the B.S. degree from Nanjing Agricultural University, Nanjing, China, in 2016 and the M.S. degree from Beijing Jiaotong University, Beijing, China, in 2018. He is currently pursuing a Ph.D. degree with the School of Electrical and Data Engineering, University of Technology Sydney, Sydney, Australia. 
His research interests include semantic image segmentation, object detection and multi-modal learning.
\end{IEEEbiography}
\vspace{-1.4cm}
\begin{IEEEbiography}[{\includegraphics[width=1in,height=1.2in,clip,keepaspectratio]{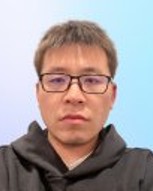}}]{Youpeng Zhao}
received the B.S. degree from Wuhan University, Wuhan, China, in 2018 and the M.S. degree from Georgia Institute of Technology, Atlanta, USA, in 2020. He is now a Ph.D. student at the University of Central Florida, Orlando, USA.
His research interests include deep learning and machine learning systems.
\end{IEEEbiography}
\vspace{-1.4cm}
\begin{IEEEbiography}
[{\includegraphics[width=1in,height=1.25in,clip,keepaspectratio]{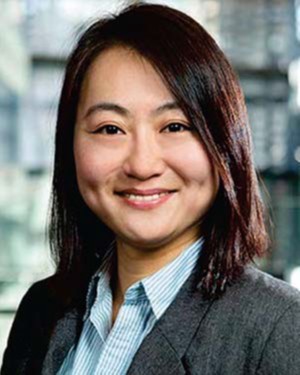}}]{Min Xu} received the BE degree
from the University of Science and Technology of
China, the MS degree from the National University
of Singapore, and the PhD degree from the University
of Newcastle, Australia. She is currently a professor
with the School of Electrical and Data Engineering,
Faculty of Engineering and Information Technology,
University of Technology Sydney and the Leader of
Visual and Aural Intelligence Lab with the Global
Big Data Technologies Centre. Her research interests
include multimedia, computer vision, and machine
learning.
\end{IEEEbiography}
\vspace{-1.3cm}
\begin{IEEEbiography}
[{\includegraphics[width=1in,height=1.5in,clip,keepaspectratio]{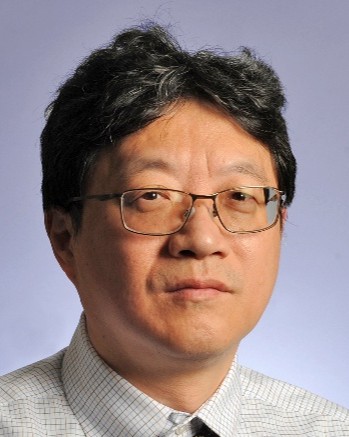}}]{Jun Wang} received the B.E. degree in computer engineering from Wuhan University, Wuhan, China, in 1993, the M.E. degree in computer engineering from the Huazhong University of Science and Technology, Wuhan, in 1996, and the Ph.D. degree in computer science and engineering from the University of Cincinnati, Cincinnati, OH, USA, in 2002.,e is currently a Full Professor of computer science and engineering and the Director of the Computer Architecture and Storage Systems Laboratory, University of Central Florida, Orlando, FL, USA. His research include big data and big compute system, data-intensive high-performance computing and deep neural network and transformers.
\end{IEEEbiography}
\vspace{-1.3cm}
\begin{IEEEbiography}[{\includegraphics[width=1in,height=1.2in,clip,keepaspectratio]{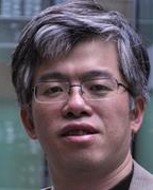}}]{Qiang Wu}
received the B.S. degree and M.S. degree from Harbin Institute of Technology, Harbin, China, in 1996 and 1998 respectively  and the P.h.D. degree from the University of Technology Sydney, Australia in 2004. He is currently an Associate Professor and core member of Global Big Data Technologies Centre.
His research interests include computer vision, image processing, pattern recognition, machine learning, and multimedia processing.
\end{IEEEbiography}




\end{document}